\documentclass[journal]{IEEEtran}

\usepackage{amsmath}
\usepackage{amsfonts}
\usepackage{hyperref}
\usepackage{graphicx}
\usepackage{subcaption}
\usepackage{makecell}
\usepackage{multirow}
\usepackage{color,soul}
\DeclareMathOperator*{\argmin}{arg\,min}

\begin{document}

\title{Deep Feature Space: A Geometrical Perspective}

\author{Ioannis~Kansizoglou,
        Loukas~Bampis,
        and~Antonios~Gasteratos,~\IEEEmembership{Senior~Member,~IEEE}
\thanks{The authors are with the Department of Production and Management Engineering, Democritus University of Thrace, Xanthi 67100, Greece e-mail: (see https://robotics.pme.duth.gr).}
}

\markboth{Journal of \LaTeX\ Class Files,~Vol.~XX, No.~XX, July~2021}%
{Shell \MakeLowercase{\textit{et al.}}: Bare Demo of IEEEtran.cls for IEEE Journals}

\maketitle

\begin{abstract}
One of the most prominent attributes of Neural Networks (NNs) constitutes their capability of learning to extract robust and descriptive features from high dimensional data, like images.
Hence, such an ability renders their exploitation as feature extractors particularly frequent in an abundance of modern reasoning systems.
Their application scope mainly includes complex cascade tasks, like multi-modal recognition and deep Reinforcement Learning (RL).
However, NNs induce implicit biases that are difficult to avoid or to deal with and are not met in traditional image descriptors.
Moreover, the lack of knowledge for describing the intra-layer properties -and thus their general behavior- restricts the further applicability of the extracted features.
With the paper at hand, a novel way of visualizing and understanding the vector space before the NNs' output layer is presented, aiming to enlighten the deep feature vectors' properties under classification tasks.
Main attention is paid to the nature of overfitting in the feature space and its adverse effect on further exploitation.
We present the findings that can be derived from our model's formulation and we evaluate them on realistic recognition scenarios, proving its prominence by improving the obtained results.
\end{abstract}


\IEEEpeerreviewmaketitle

\section{Introduction}

\IEEEPARstart{T}{he} contribution of Neural Networks (NNs) in computer science comprises an indisputable fact.
A simple search through contemporary works suffices to convince us that Deep Learning (DL) has significantly benefitted the performance of machines on competitive problems, in fields like computer vision~\cite{badrinarayanan2017segnet}, signal~\cite{purwins2019deep} and natural-language processing~\cite{young2018recent}.
Consequently, the use of NNs is expanded rapidly in more and more tasks, conventional~{\cite{rawat2017deep}} or innovative~{\cite{zhang2020deep}}, while their manipulation is simplified by the development of advanced hardware processing units~\cite{jouppi2017datacenter} and DL frameworks~\cite{abadi2016tensorflow,paszke2017automatic}.

However, one should never forget that NNs, like the rest of machine learning algorithms, constitute the tool and not the solution to a problem.
The more we comprehend about them as tools, by discovering useful properties along with the advantages and disadvantages that they display, the better we can exploit them in plenty of problems.
Frequently, such an understanding is based on their behavior on various tasks since they appear to be particularly complex as a whole.
Hence, our knowledge originates from experimental results and empirical rules, which however describe how they behave rather than how they operate.
The above distinction denotes how far we find ourselves from their optimal usage~\cite{hawkins2007intelligence}.

In that event, the paper at hand focuses on the behavior of the feature vectors before the output layer of a Deep Neural Network (DNN).
The motivation behind this analysis mainly lies in the tendency to exploit DNNs as feature extractors in cascade and/or multi-modal tasks.
Several novel approaches have already been proposed, which employ suitable loss functions, aiming to improve the classification performance of DNNs~\cite{ranjan2017l2}.
The main benefit of the proposed functions constitutes their capability of learning angularly discriminative features~\cite{liu2017sphereface},~{\cite{wang2018additive}}, \cite{deng2019arcface}. 
The work at hand takes a step further and analyzes practical properties of DNNs as feature extractors without adopting any convention.
The provided contributions are as follows:
\begin{itemize}

\item the loci of the target classes in the deep feature space of the last hidden layer of a DNN are defined and visually illustrated for the first time;

\item the supremacy of the feature vectors' orientation against their norm in the classification outcome is proved, while a descriptive demonstration about their distribution within the target class locus is provided;

\item an efficient method for assessing the nature of overfitting and the distribution of classifiers in the feature space is proposed and empirically studied, by introducing two metrics, \textit{viz.}, \textit{centrality} and \textit{separability};

\item a data handling strategy is suggested to cope with overfitting in neural-based extractors, as it is captured by our analysis, in cases of further training and fusion.

\end{itemize}
Note that in contrast to the aforementioned works, the following analysis applies to both shallow networks, \textit{i.e.}, networks with one hidden layer and deep neural networks, without any further convention, like custom softmax loss functions or weight and feature vector normalization~{\cite{liu2016large}}.
The only premise of our analysis constitutes the usage of the Softmax loss or one of its variants~{\cite{liu2017sphereface}},~{\cite{wang2018additive}},~{\cite{deng2019arcface}},~{\cite{liu2016large}}.

The remainder of this paper is structured as follows.
In Section II, we discuss representative works that highlight the exploitation of DNNs as feature extractors in unimodal and multi-modal tasks, as well as approaches that deal with the creation of discriminative feature vectors.
Consequently, Section III contains our detailed theoretical approach, while in Section IV experiments are conducted to display its beneficial results on practical problems.
In the last section, we draw conclusions and present suggestions for future work.

\section{Related Work}

\begin{figure*}
    \centering
    \begin{subfigure}[b]{0.245\textwidth}
        \centering
        \includegraphics[width=\textwidth]{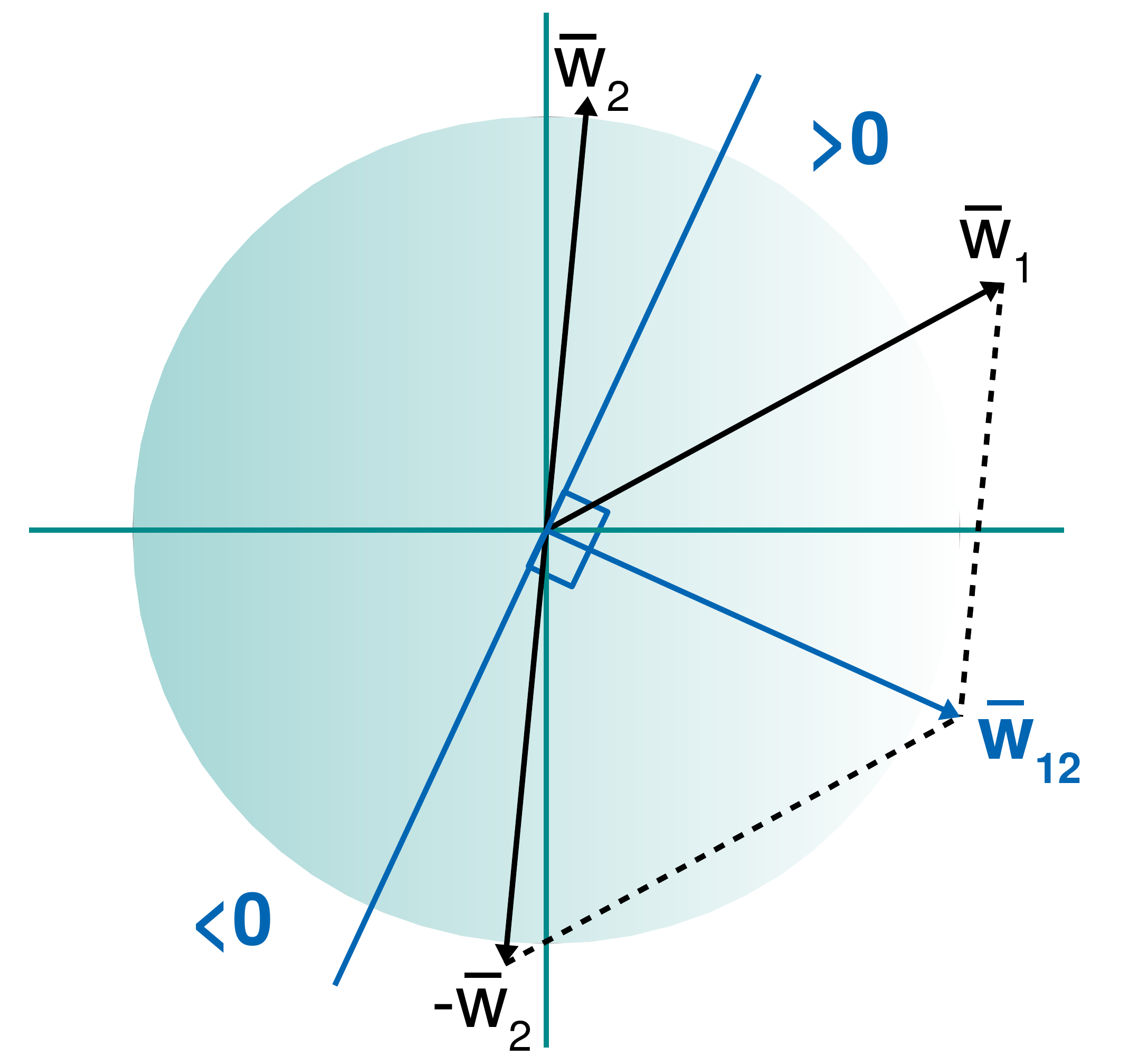}
        \caption[]{}   
        \label{fig:sub1a}
    \end{subfigure}
    \hfill
    \begin{subfigure}[b]{0.245\textwidth}
        \centering 
        \includegraphics[width=\textwidth]{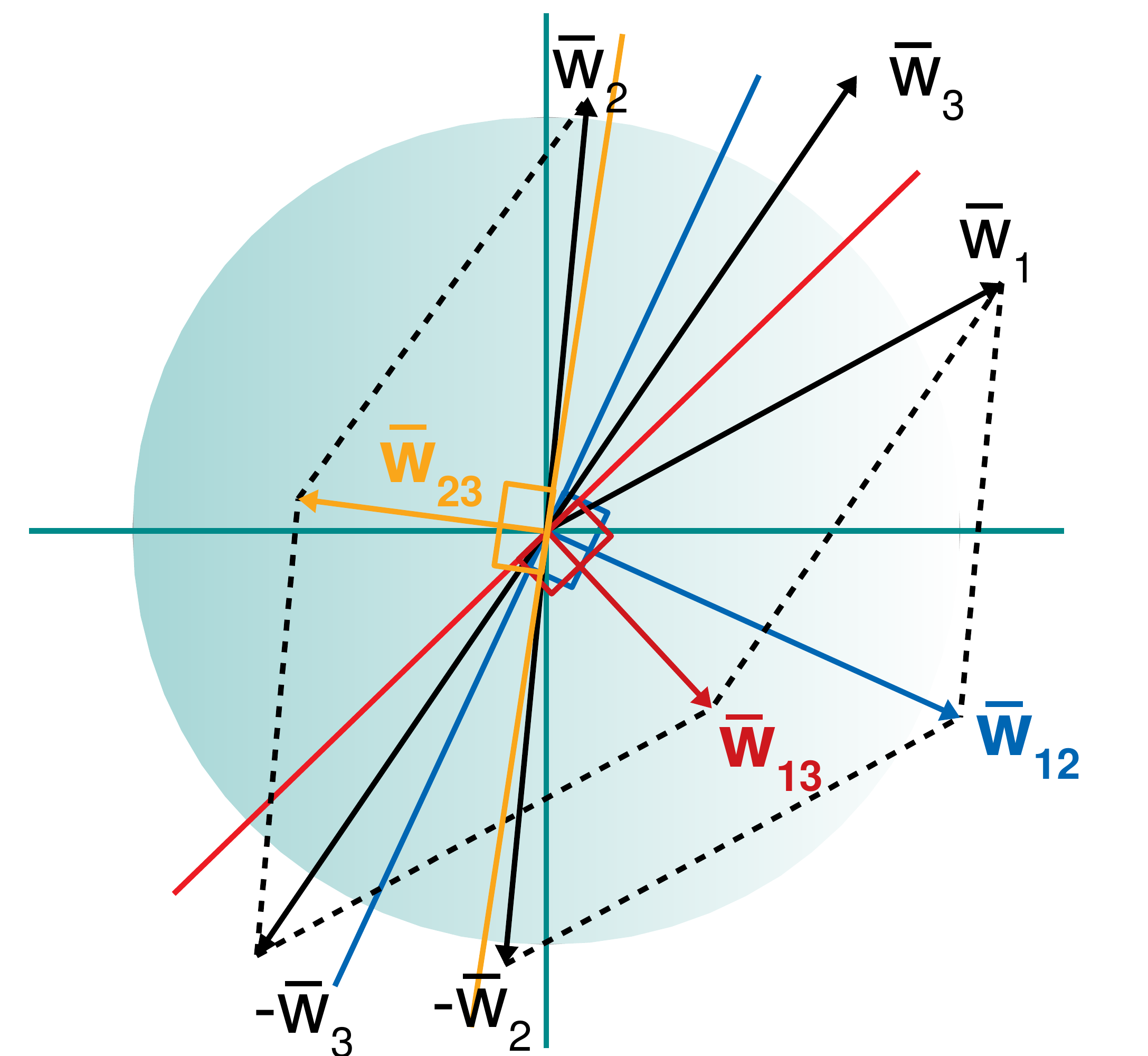}
        \caption[]{} 
        \label{fig:sub1b}
    \end{subfigure}
    \hfill
    \begin{subfigure}[b]{0.245\textwidth}   
        \centering 
        \includegraphics[width=\textwidth]{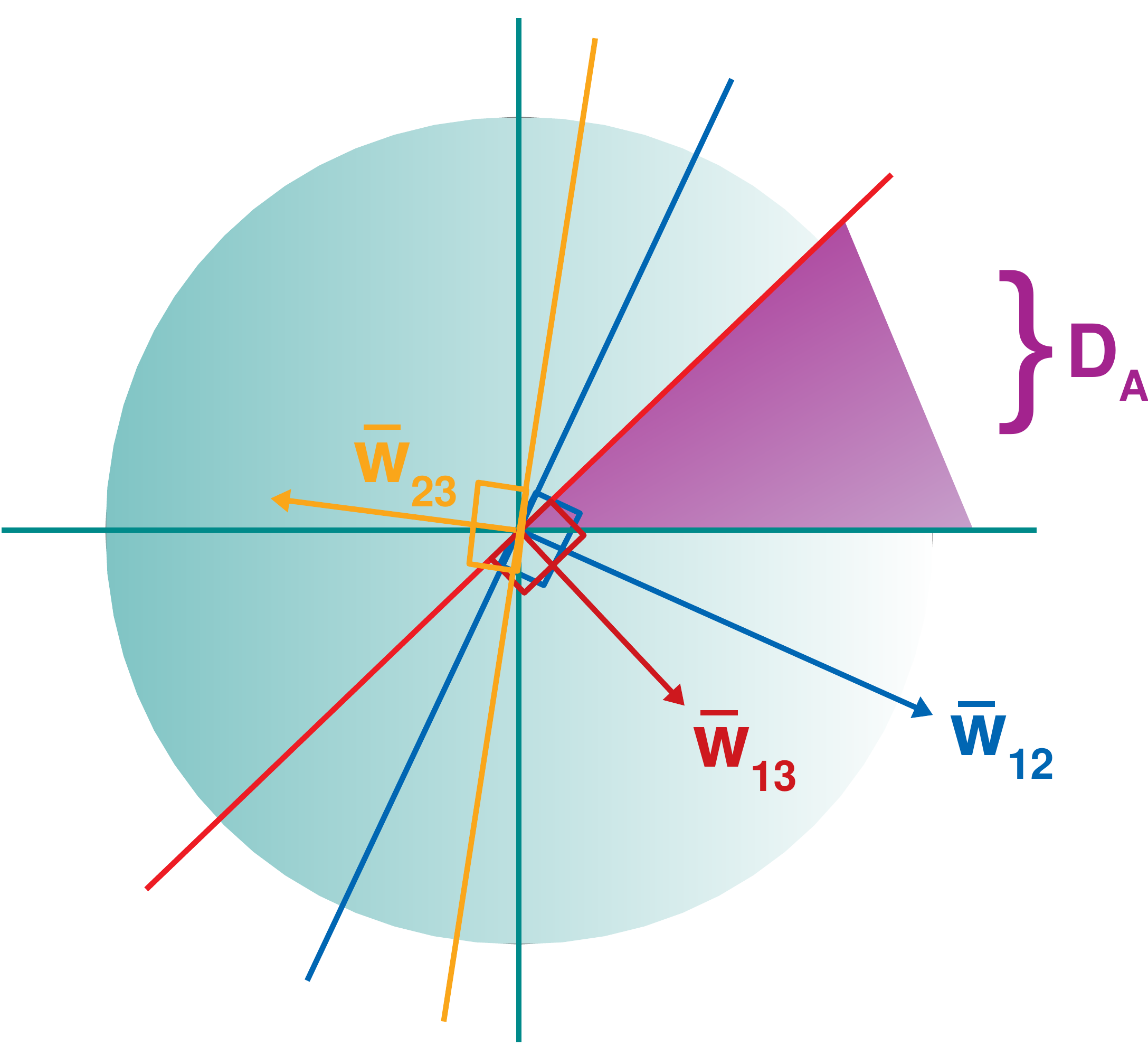}
        \caption[]{}   
        \label{fig:sub1c}
    \end{subfigure}
    \hfill
    \begin{subfigure}[b]{0.245\textwidth}   
        \centering 
        \includegraphics[width=\textwidth]{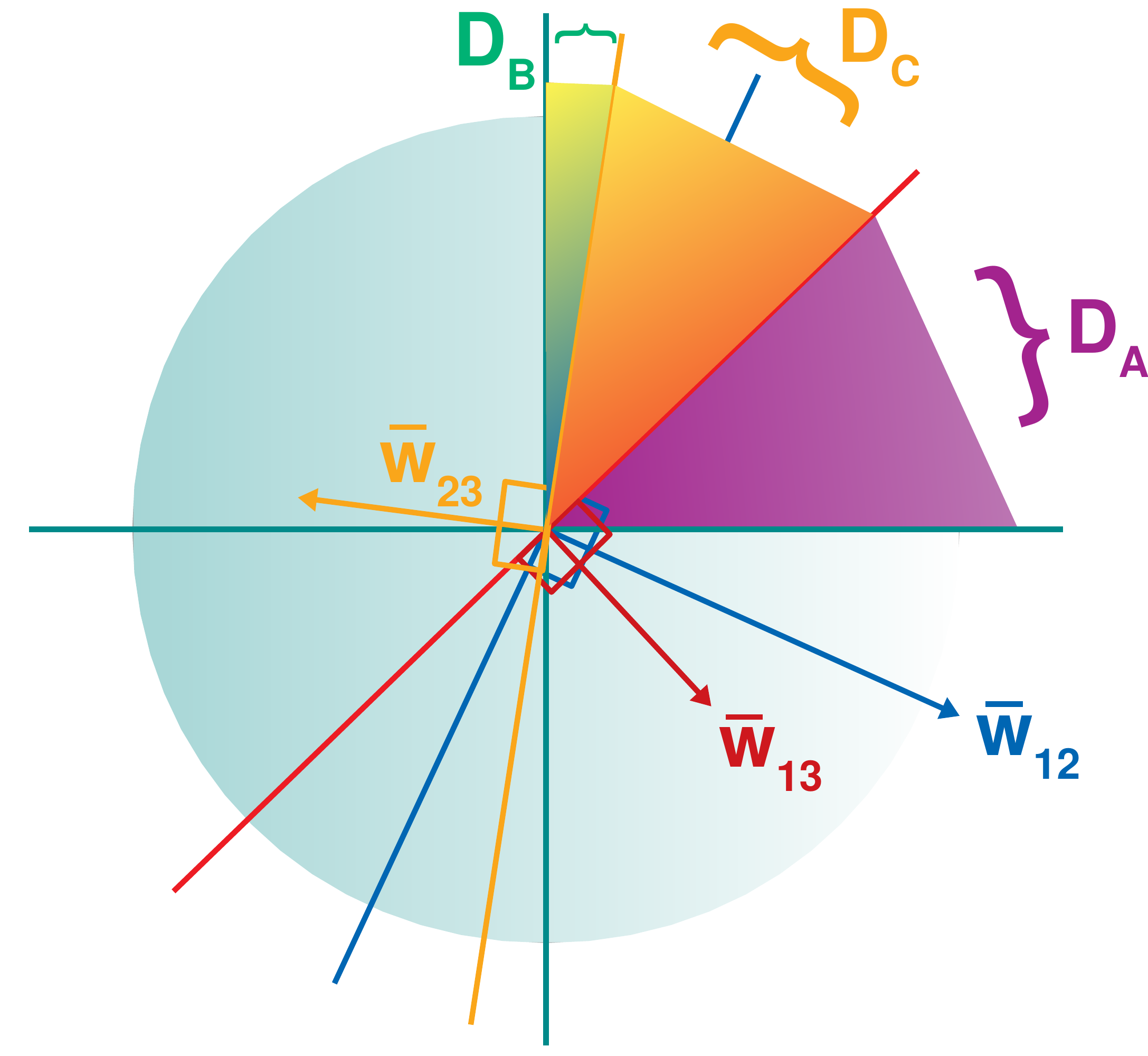}
        \caption[]{}   
    \label{fig:sub1d}
    \end{subfigure}
    \caption[ The average and standard deviation of critical parameters ]
    {\small Feature space division in $\mathbb{R}^2$ with the \textit{ReLU} activation function: (a) The differential vector $\bar{w}_{12}=\bar{w}_{1}-\bar{w}_{2}$  divides the space into two halves, with the positive one denoting the subspace with $z_1>z_2$ and vise versa. (b) The same applies for every differential vector. (c) The positive subspaces' intersection of the differential vectors that include $\bar{w}_1$, \textit{i.e.}, $\bar{w}_{12}$ and $\bar{w}_{13}$, define the locus ($D_1$) of $A_1$ class, while (d) the same applies for the rest of the classes' loci ($D_2$ and $D_3$).}
    \label{fig:1}
\end{figure*}

The current section presents a comprehensive description regarding the most representative works that deal with the extraction of features exploiting large inter-class distance from DNNs.
Subsequently, their wide usage on several cascade and multi-modal tasks is discussed.

\subsection{Discriminative Features}
\label{subsec:DisFeat}
The demand on keeping the compactness of the intra-class learned features relatively larger than their corresponding inter-class separability is a salient subject in several learning approaches~\cite{sun2014deep,wen2016discriminative,adeli2018semi}.
A contemporary trend includes the reinforcement of the DNN's loss with angular margins between the different classes, so as to enhance the feature vectors' discriminability.
Towards that end, the Large-Margin Softmax Loss (L-Softmax)~\cite{liu2016large} combines the cross-entropy loss with the last fully-connected layer and a softmax function to create more rigorous angular boundaries.
Several theoretical insights along with empirical studies showed that the introduction of hyperspherical or decoupled convolution operations in CNNs can improve performance~{\cite{liu2017deep}},~{\cite{liu2018decoupled}}.
Especially, in Face Recognition (FR) numerous approaches  for discriminative features have been proposed since the aforementioned necessity is highly desired both for face identification and verification.
As an instance, an $L_2$-normalization on the feature vectors can restrict them to lie on a hypersphere manifold with adjustable radius, achieving improved results on face verification~\cite{ranjan2017l2}.
Moreover, SphereFace~\cite{liu2017sphereface} succeeded advanced discriminability by applying adjustable multiplicative angular margins in the decision boundaries.
On the contrary, a corresponding additive cosine margin was investigated in CosFace~\cite{wang2018cosface}.
Eventually, the combination of the multiplicative and the additive cosine, together with an additive angular margin adopted in ArcFace, proved to outperform any other method~\cite{deng2019arcface}.

\subsection{DNNs for Feature Extraction}

The advantageous behavior of DNNs and more specifically of Convolutional Neural Networks (CNNs) as image descriptors has been particulartly investigated in FR~\cite{lu2019experimental}, as already stated in Section~\ref{subsec:DisFeat}.
However, their exploitation has been also proved beneficial in a wide variety of complex tasks.
Firstly, in the case of multi-modal learning~\cite{baltruvsaitis2018multimodal}, a priori training of the unimodal architectures is often performed~\cite{mroueh2015deep}, given the lack of available multi-modal data and the complexity of the developed architectures.
The above models can be also implemented by incorporating temporal information through Long Short-Term Memory (LSTM) cells~\cite{fan2016video}.
Then, a fusion algorithm can be trained from the output feature vectors of the unimodal architectures~\cite{zhang2017learning}.
The fusion model can be a DNN~\cite{kansizoglou2019active}, a Deep Belief Network (DBN)~\cite{ma2019audio}, or even an LSTM network~\cite{tzirakis2017end}.
Modality fusion has also become extremely widespread in Visual Question Answering (VQA), where the last hidden layer of an embedded CNN architecture, such as VGGNet~\cite{simonyan2014very}, is concatenated with the output vector of an LSTM network to combine visual and textual data~\cite{antol2015vqa}.
In addition, in cases that temporal quality needs to be included in the decision procedure, like in human action recognition, the use of deep feature extractors has been proved particularly beneficial~\cite{ji20123d}.
Moreover, deep Reinforcement Learning (RL) constitutes another challenging task, where a CNN can be accurately trained as an image descriptor to provide the RL agent with a feature vector, representing its current state~\cite{Caicedo_2015_ICCV}.

\subsection{Hyperspherical Learning} \label{subsec:HypLearn}

Hence, considering the applicability of deep feature vectors in a wide variety of scientific fields, it becomes essential to investigate their behavior and properties during and after training.
In this regard, several geometrical properties of DNNs have been discussed in ~{\cite{liu2018learning}} and ~{\cite{lin2020regularizing}}, where the deep feature space is trained through a hyperspherical energy minimization scheme inspired by the Thompson problem in physics~\cite{thomson1904xxiv}.
The above scheme replaces the common Softmax optimization goal and encourages specific symmetries regarding the distribution of a DNN's weights in the feature space.
This allows for the utilization of efficient regularization techniques, such as the widely known approaches of orthonormal~{\cite{liu2017deep}} and orthogonality~{\cite{rodriguez2016regularizing}}.
Moreover, hyperspherical learning has been proved beneficial in implementations like the one presented in~{\cite{chen2019angular}}.
This technique refers to the calculation of a sample's hardness based on its angular distance from the target class, which is empirically considered as the weight vector of the class.
Due to the distinctive optimization goal, it is crucial to clarify that, by forcing the weights of the last hidden layer to follow specific symmetries, the findings of hyperspherical learning approaches obey to different principles than the ones studied here.
Yet, an analysis regarding the usage of DNNs for feature extraction shall initially focus on the simple Softmax function since this is the commonly adopted technique in classification tasks.
Subsequently, within our study, we will also discover and understand useful effects of approaches, like the $L_2$-constrained Softmax~{\cite{liu2016large}}, on the feature vectors' distribution.

\section{Method}
\label{sec:Method}

\subsection{Feature Space Division}\label{sub:SpDiv}

As stated above, the main scope of our work is to study the behavior in one of the most common uses of DNNs, \textit{viz.}, the classification task. 
Specifically, we examine the way that the input space of the output layer is divided by the target classes and, as a result, how the classification goal affects the distribution of its input vectors $\bar{a}\in\mathbb{R}^n$. 
Note that $n$ corresponds to the number of neurons in the previous layer and in turn, the feature vectors' size. 
The term deep feature vector will be employed to describe the whole output of the penultimate layer, thus referring to a vector that captures the input's properties as quantified by the entire DNN and forming a global descriptor of the input data.

\subsubsection{Preface}\label{subsub:pref1} We denote the feature space $\mathcal{F}\subseteq\mathbb{R}^{n+1}$ according to the neuron's equation:
\begin{equation}\label{eq:NeuEq}
z = \sum_{j}^{n}{a_j w_j}+b,
\end{equation}
with $a_j$ being the feature vector's coefficients, $w_j$ the trainable weights of the neuron and $b$ its trainable bias.
Then, we can define the expanded feature vector $\bar{a}_e\in\mathcal{F}$ and the $i$-th neuron's trainable parameters $\bar{w}_i\in\mathcal{F}$:
\begin{equation}
\begin{split}
\bar{a}_e = [a_1, a_2, ..., a_n, 1]\text{ and } 
\bar{w}_i = [w_1, w_2, ..., w_n, b],
\end{split}
\end{equation}
in order to simplify Eq.~\ref{eq:NeuEq} to the dot product:
\begin{equation}
z_i = \bar{a}_e\cdot\bar{w}_i.
\end{equation}
Moreover, without loss of generality, we will consider $\mathcal{F}$ as an $n$-sphere with an infinite radius $R\simeq \infty$, so that the $S^n(R)$ encompasses all the possible feature vectors.

\subsubsection{Simple Case}

Considering a simple classification problem between three discrete classes, namely class $A_1$, $A_2$ and $A_3$, we employ three output neurons in the output layer.
Then, the softmax activation function is applied to extract the classification result, known by the following equation:
\begin{equation}\label{eq:1}
y_i = \frac{e^{z_i}}{\sum_j^3{e^{z_j}}}=\frac{e^{\bar{a}_e\cdot\bar{w}_i}}{\sum_j^3{e^{\bar{a}_e\cdot\bar{w}_j}}},
\end{equation}
where $z_i$ represents the output of the $i$-th output neuron and $\bar{w}_{i}$ its trainable parameters.
Keeping the case simple enough for visualization purposes (Fig.~\ref{fig:1}), we consider 2-dimensional feature vectors by also ignoring the bias.
Since the $ReLU$ activation function is applied in most deep learning architectures, it is also adopted here, forcing the feature vector's coordinates to take non-negative values.
However, the following analysis applies even when $ReLU$ is not employed.

In order for the feature vector $\bar{a}_e$ to be classified in class $A_1$, both of the following criteria should be satisfied:
\begin{equation}\label{eq:21}
\begin{split}
\left.
\begin{array}{ll}
z_1 > z_2 \\
z_1 > z_3 \\
\end{array} 
\right\} \implies
\left.
\begin{array}{ll}
\bar{a}_e\cdot{\bar{w}_1} > \bar{a}_e\cdot{\bar{w}_2} \\
\bar{a}_e\cdot{\bar{w}_1} > \bar{a}_e\cdot{\bar{w}_3} \\
\end{array} 
\right\} \implies \\
\implies
\left.
\begin{array}{ll}
\bar{a}_e\cdot({\bar{w}_1}-{\bar{w}_2}) > 0 \\
\bar{a}_e\cdot({\bar{w}_1}-{\bar{w}_3}) > 0 \\
\end{array} 
\right\} \implies
\left.
\begin{array}{ll}
\bar{a}_e\cdot{\bar{w}_{12}} > 0 \\
\bar{a}_e\cdot{\bar{w}_{13}} > 0 \\
\end{array} 
\right\},
\end{split}
\end{equation}
where $\bar{w}_{12}=\bar{w}_1-\bar{w}_2$ and $\bar{w}_{13}=\bar{w}_1-\bar{w}_3$.
From now on, we keep this notation for vectors defined as $\bar{w}_{ij}=\bar{w}_i-\bar{w}_j$ and refer to them as differential vectors.
One can easily observe that $\bar{w}_{ij}=-\bar{w}_{ji}$.

Hence, according to Eq. \ref{eq:21}, the first step is to calculate the differential vectors $\bar{w}_{12}$ and $\bar{w}_{13}$.
Then, the sign of the dot product and consequently the angles of vector $\bar{a}_e$ with $\bar{w}_{12}$ and $\bar{w}_{13}$ specifies the locus of each criterion, respectively.
For instance, by solving the equation $\cos\widehat{(\bar{a}_e,\bar{w}_{12})} = 0 => \widehat{(\bar{a}_e,\bar{w}_{12})} = \pm \frac{\pi}{2}$, we define in $\mathbb{R}^2$ the perpendicular to $\bar{w}_{12}$ line, as the separation line between the positive and negative values of the above dot product.
The vector $\bar{w}_{12}$ indicates the subspace of the positive ones, specified by the condition $\widehat{(\bar{a}_e,\bar{w}_{12})}\in(-\frac{\pi}{2},\frac{\pi}{2})$.
The geometric interpretation of the above procedure is illustrated in Fig.~\ref{fig:sub1a}.
We can work similarly to define the separation lines of every differential vector, as shown in Fig.~\ref{fig:sub1b}.
All the possible feature vectors lie inside the $1$-sphere, $S^1(R)$ (see Section~\ref{subsub:pref1}) and each separation line divides the $S^1(R)$ into two semicircles (Fig.~\ref{fig:sub1a}).

The locus of the subspace of class $A_1$ is defined by the simultaneous satisfaction of the criteria in Eq. \ref{eq:21}, \textit{i.e.}, by the intersection of the following positive subspaces:
\begin{equation}
\begin{split}
D_{12} = \{\bar{a}_e\in S^1(R): \widehat{(\bar{a}_e,\bar{w}_{12})}\in(-\frac{\pi}{2},\frac{\pi}{2})\} \\
D_{13} = \{\bar{a}_e\in S^1(R): \widehat{(\bar{a}_e,\bar{w}_{13})}\in(-\frac{\pi}{2},\frac{\pi}{2})\}
\end{split}
\end{equation}
Moreover, the $ReLU$ function restricts the allowable space of the feature vectors only to the $\mathbb{R}^2_{\geq{0}}$ subspace.
This includes the first quadrant of the $S^1(R)$, where both coordinates take non-negative values.
Hence, the desired subspace results as:
\begin{equation}\label{eq:22}
D_1 = \mathbb{R}^2_{\geq{0}}\bigcap{D_{12}}\bigcap{D_{13}},
\end{equation}
which is depicted in Fig.~\ref{fig:sub1c}.

Similarly, we define the subspaces of classes $A_2$ and $A_3$ ($D_2$, $D_3$) as follows:
\begin{equation}\label{eq:23}
\begin{split}
D_2 = \mathbb{R}^2_{\geq{0}}\bigcap{D_{21}}\bigcap{D_{23}}, \\
D_3 = \mathbb{R}^2_{\geq{0}}\bigcap{D_{31}}\bigcap{D_{32}},
\end{split}
\end{equation}
resulting to the space division of Fig. \ref{fig:sub1d}.
Notice that all the available space of $R^2_{\geq 0}$ is exploited, while no intersection exists between the subspaces $D_1$, $D_2$ and $D_3$.
As mentioned above, each separation line divides the $S^1(R)$ into two semicircles, thus into two convex sets.
In addition, $R^2_{\geq 0}$ is also a convex set.
Since the intersection of any family of convex sets (finite or infinite) constitutes also a convex set~\cite{gallier2011basic}, we ensure that the subspaces $D_1$, $D_2$ and $D_3$ are convex, as well.
The above property is essential for our following analysis.

\subsubsection{Generalizing}

\begin{figure}
    \centering
    \begin{subfigure}{0.24\textwidth}
        \centering
        \includegraphics[width=0.8\linewidth]{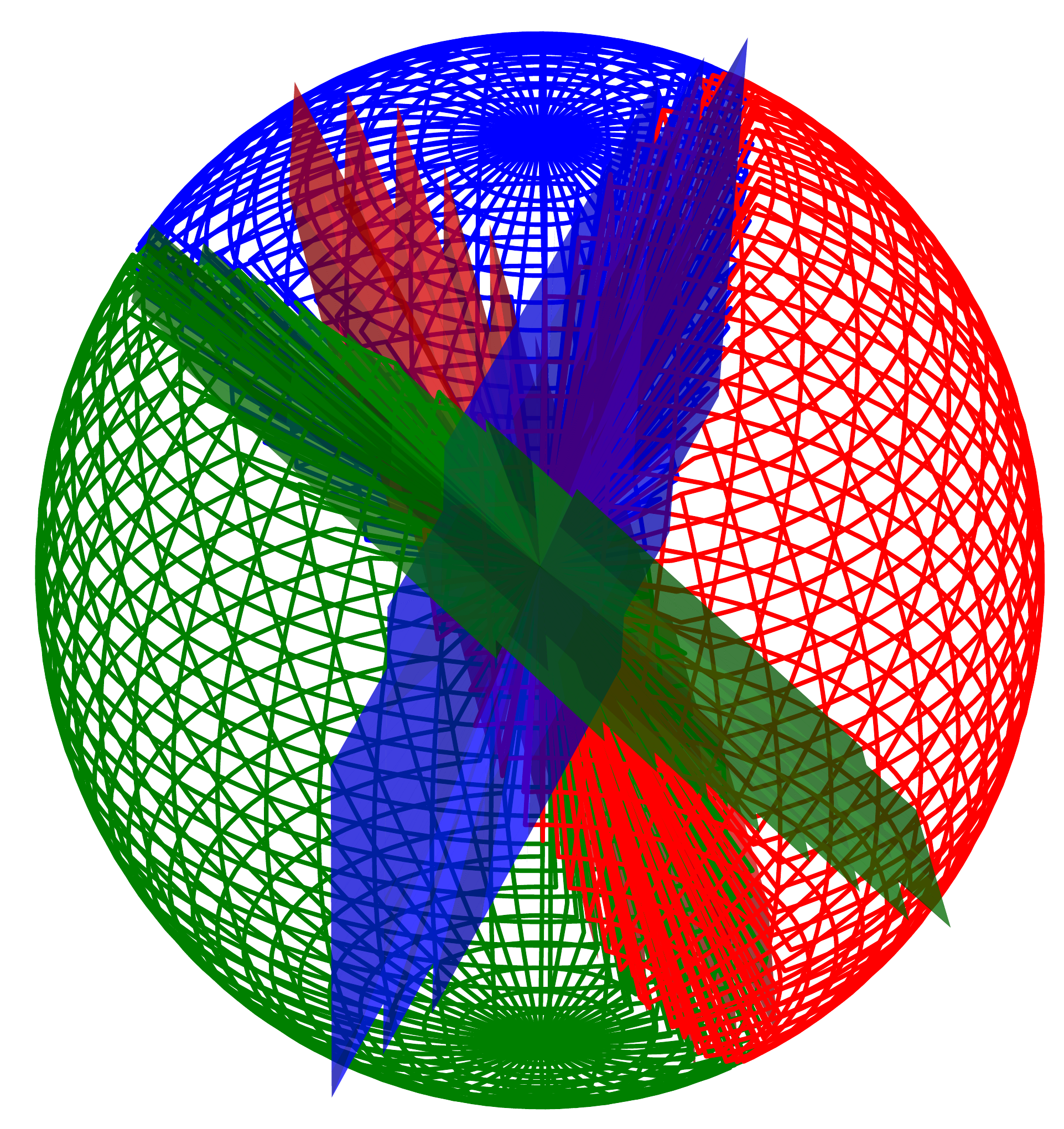}
        \caption{}
        \label{fig:sub2a}
    \end{subfigure}%
	~
    \begin{subfigure}{0.24\textwidth}
        \centering
        \includegraphics[width=0.8\linewidth]{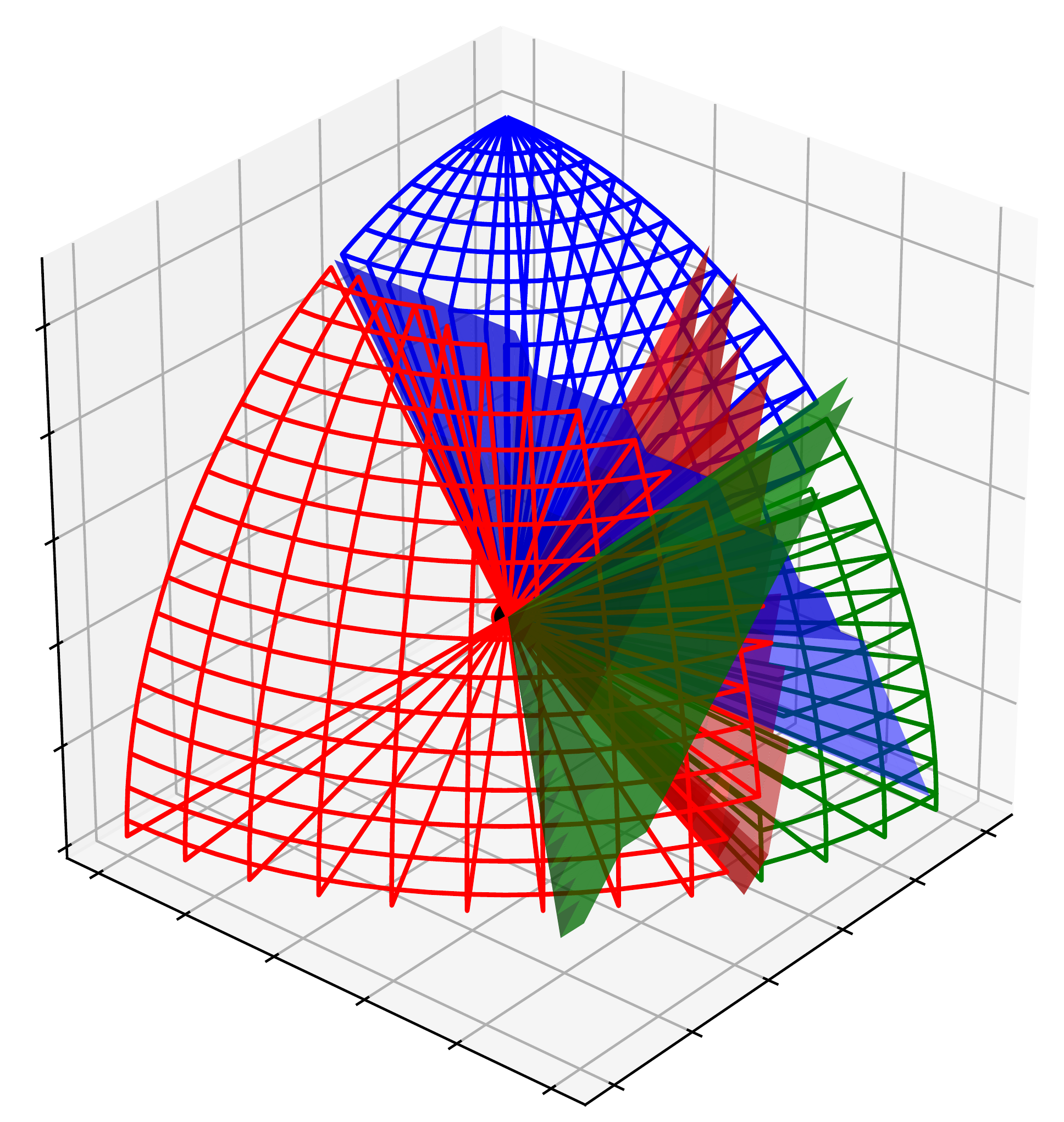}
        \caption{}
        \label{fig:sub2b}
    \end{subfigure}
    \caption{Feature space division in $\mathbb{R}^3$ (a) without and (b) with the exploitation of the \textit{ReLU} activation function.}
    \label{fig:2}
\end{figure}

\begin{table}
\centering
\caption{Number of differential vectors based on the number of the target classes.}
\label{table:NumSepHyp}
\resizebox{\linewidth}{!}{%
\renewcommand{\arraystretch}{1.2}
\begin{tabular}{|c|c|c|c|c|c|c|} 
 \hline
 Number of Classes & 2 & 3 & 4 & 5 & ... & $N$\\
 \hline
 Differential Vectors & 1 & 3 & 6 & 10 & ... & $\sum_{i=1}^{N-1}{i}$\\ 
 \hline
\end{tabular}}
\end{table}

Let us consider a classification task within the feature space $\mathcal{F}\subseteq\mathbb{R}^{n+1}$ with \mbox{$N\in\mathbb{N}-\{0,1\}$} possible classes for the feature vectors \mbox{$\bar{a}_e\in\mathcal{F}$}.
Similarly, all the possible feature vectors will be located inside the \mbox{$n$-sphere}, $S^{n}(R)$.
Then, each differential vector $\bar{w}_{ij}$, with $i,j = \{1,2,...,N\}$ and $i\neq{j}$, defines a separation hyperplane $H_{ij}\in\mathbb{R}^n$ that constitutes the solution of the linear equation $\bar{a}_e\bar{w}_{ij}=0$.
Such a hyperplane is defined in geometry as a vector hyperplane and passes through the origin.
Thus, every separation hyperplane divides $S^{n}(R)$ into two half-spaces, one including the locus of the positive results of the aforementioned linear equation and the other including the negative ones.
The differential vector indicates the subspace with the positive results defined as:

\begin{equation}\label{eq:24}
D_{ij} =\{\bar{a}_e\in S^{n}(R): \widehat{(\bar{a}_e,\bar{w}_{ij})}\in(-\frac{\pi}{2},\frac{\pi}{2})\}.
\end{equation}
The number of the separation hyperplanes adhere to Table. \ref{table:NumSepHyp}.

In accordance to our simple case, the subspace of the $i$-th class in $S^{n}(R)$ is specified by the fulfillment of all the criteria that include the differential vectors of the specific class $i$.
As a consequence, the positive subspaces that those differential vectors define, intersect to the locus of the desired class.
Considering the above and Eq. \ref{eq:24}, we can write for the subspace of the $i$-th class:
\begin{equation}\label{eq:25}
D_i{=}
\begin{cases}
\begin{aligned}
\,\,\bigcap_{j=2}^{N}{D_{ij}},& \quad\text{for}\ i{=}1,\\
\,\,\bigg(\bigcap_{j=1}^{i-1}{D_{ij}}\bigg)\bigcap{\bigg(\bigcap_{j=i+1}^{N}{D_{ij}}\bigg)},& \quad\text{for}\ i{=}\{2,...,N{-}1\},\\
\,\,\bigcap_{j=1}^{N-1}{D_{ij}},& \quad\text{for}\ i{=}N.
\end{aligned}
\end{cases}
\end{equation}
Eq. \ref{eq:25} applies for classification problems working on any dimension, without applying the $ReLU$ activation function on the feature vectors.
Otherwise the space for each one of the $N$ classes arises from the intersection of the Eq. \ref{eq:25} with the $R^{n+1}_{\geq 0}$ subspace.
In Fig. \ref{fig:2}, a division of the $\mathbb{R}^3$ space is illustrated both with or without the exploitation of $ReLU$.

Finally, since both $R^{n+1}_{\geq 0}$ and $D_{ij}$ constitute convex sets $\forall i,j=\{1,2,...,N\}$ with $i\neq{j}$, their intersections are also convex sets.
Consequently, the subspace of any class in the $n$-dimensional space is always a convex set, securing that no discontinuity exists inside the class's subspace. 

\subsection{Feature vectors' sensitivity analysis }
\label{sub:SenAnal}

As described in Section \ref{sub:SpDiv}, the angles between the feature vector $\bar{a}_e$ and the differential vectors are decisive for the definition of the prevalent neuron.
However, a high classification confidence is not exclusively met by the criteria of Eq.~\ref{eq:21}.
On the contrary, a specific distance from the decision boundaries is required, which is dependent both on the angular distances and the feature vector's norm.
Hence, the impact of both angle's and norm's variations of a feature vector ($\bar{a}_e$) on the final result is investigated. 

\subsubsection{Preface}
\begin{figure}
    \centering
    \includegraphics[width=0.8\linewidth]{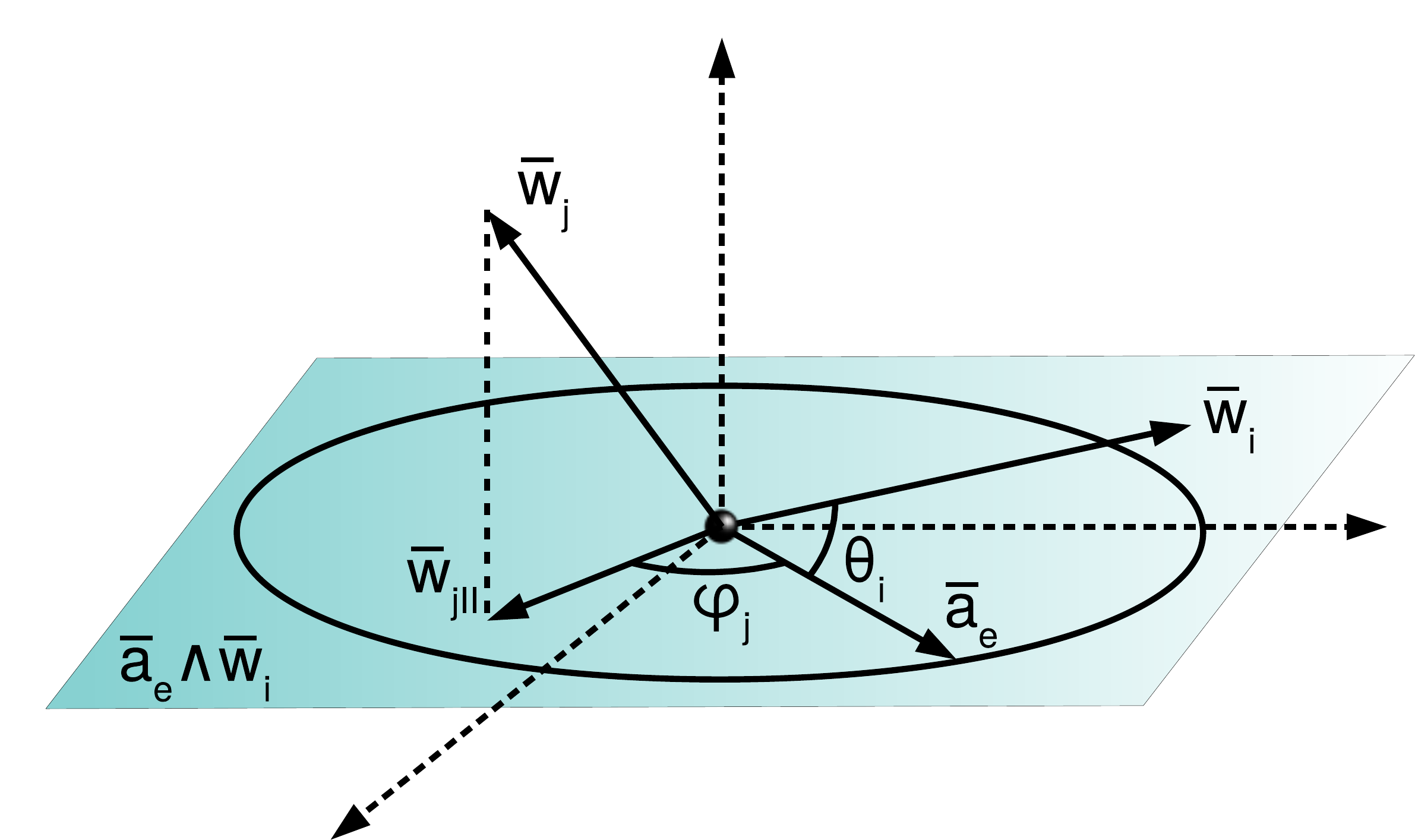}
    \caption{\small Plane of variations defined by the feature vector $\bar{a}_e$ and the weight $\bar{w}_1$ of the prevailing class $A_1$. Each variation of $\bar{a}_e$, including its norm and its orientation, takes place onto this plane.}
        \label{fig:RP}
\end{figure}
We focus our interest on feature vectors that belong to the subspace of their correct class, but they are not suitably placed in order for their classification cost to approximate zero.
In any other case, either the angle should initially be fixed to minimize the calculated loss, or the loss is too small and no further variation is required.

\begin{figure*}
    \centering
    \begin{subfigure}[b]{0.325\textwidth}
        \centering
        \includegraphics[width=\textwidth]{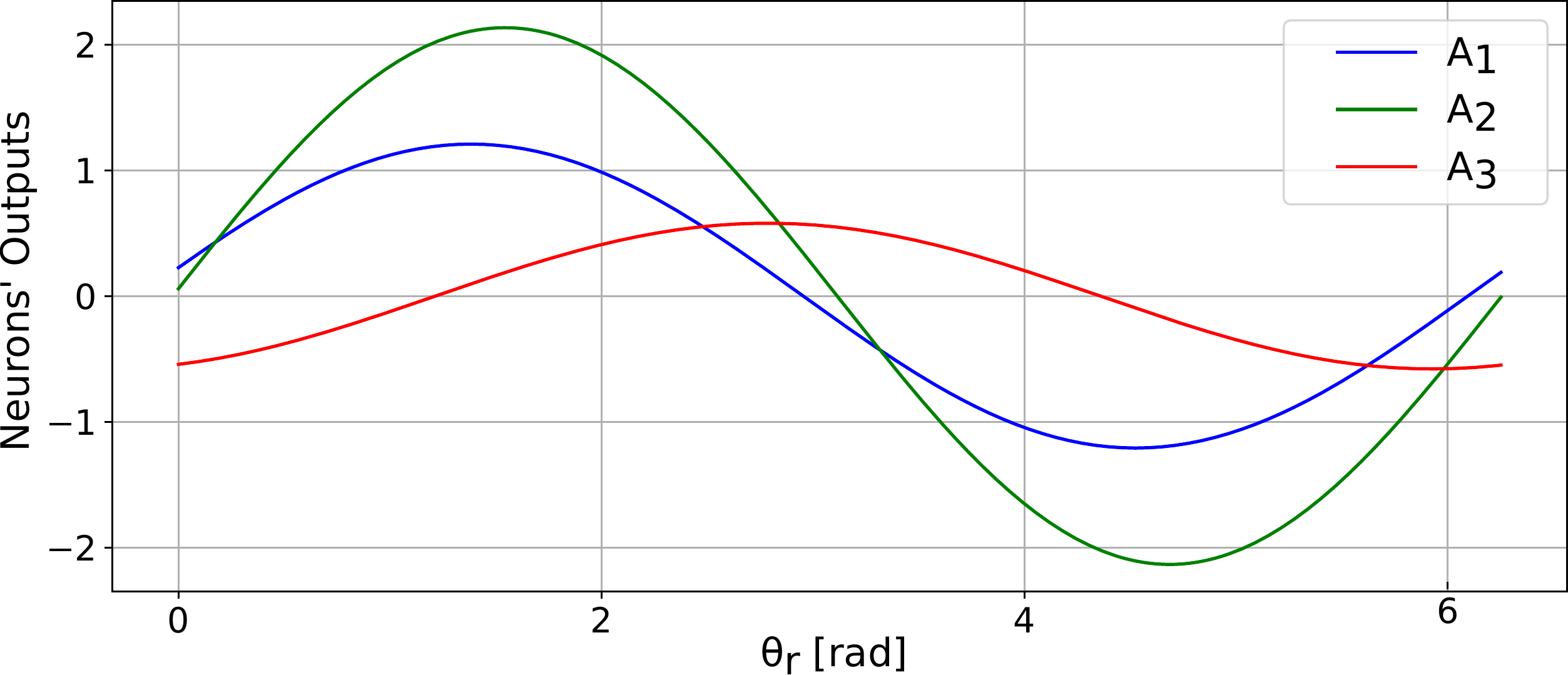}
		\hfill        
        \caption{}
        \label{fig:sub5a}
    \end{subfigure}
    \begin{subfigure}[b]{0.325\textwidth}
        \centering
        \includegraphics[width=\textwidth]{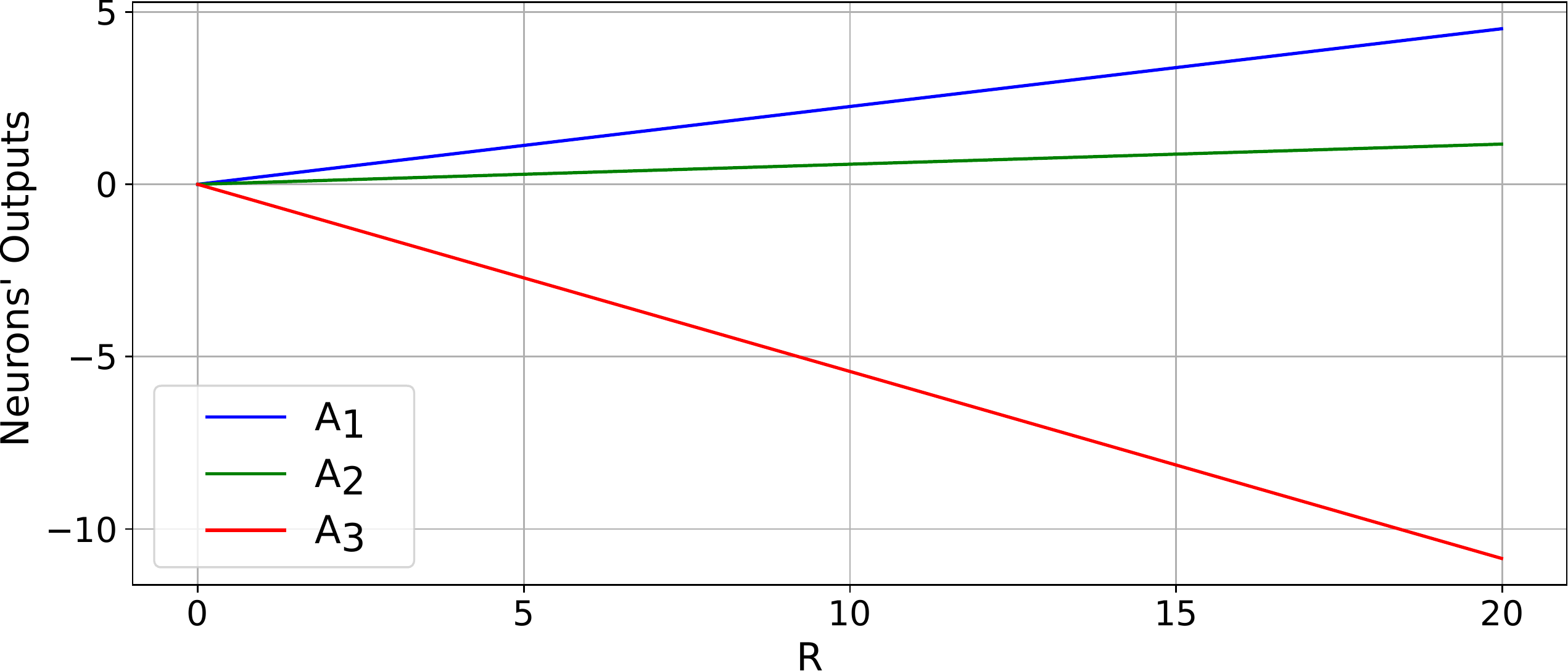}
		\hfill           
        \caption[]{}
        \label{fig:sub5b}
    \end{subfigure}
    \begin{subfigure}[b]{0.325\textwidth}
        \centering 
        \includegraphics[width=0.7\textwidth]{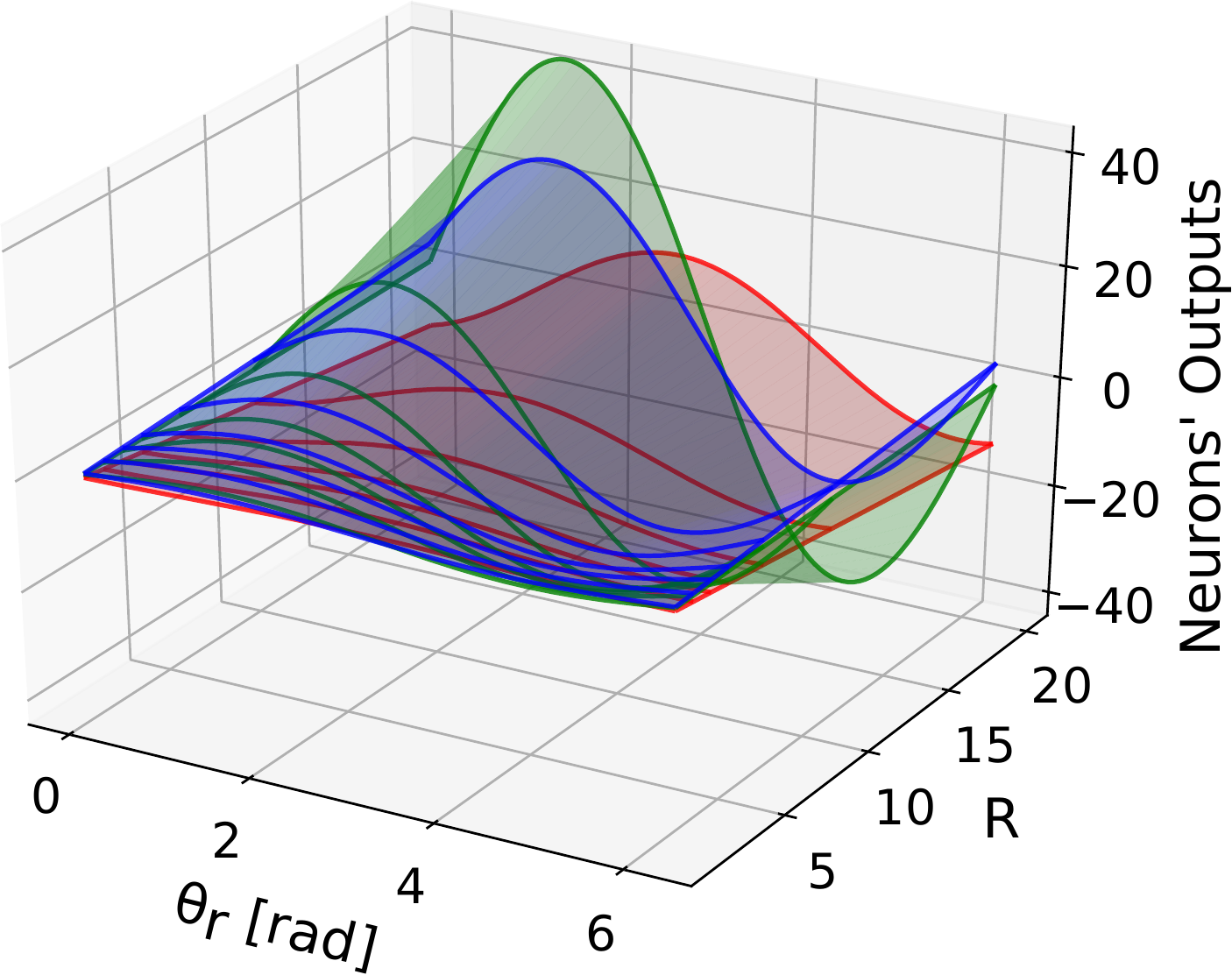}      
		\hfill        
        \caption[]{}    
        \label{fig:sub5c}
    \end{subfigure}
    \vskip\baselineskip
    \centering
    \begin{subfigure}[b]{0.325\textwidth}   
        \centering 
        \includegraphics[width=\textwidth]{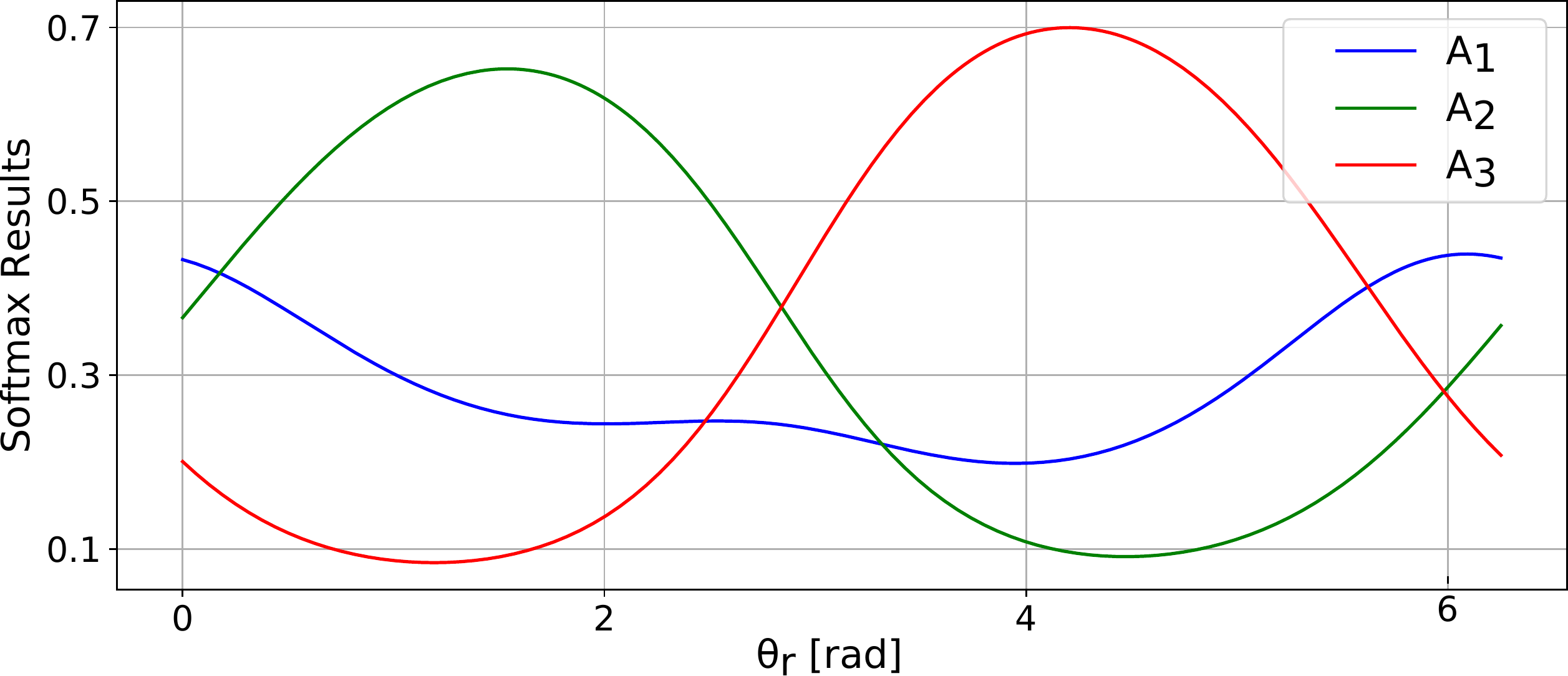}
        \hfill 
        \caption[]{}    
        \label{fig:sub5d}
    \end{subfigure}
    \begin{subfigure}[b]{0.325\textwidth}   
        \centering 
        \includegraphics[width=\textwidth]{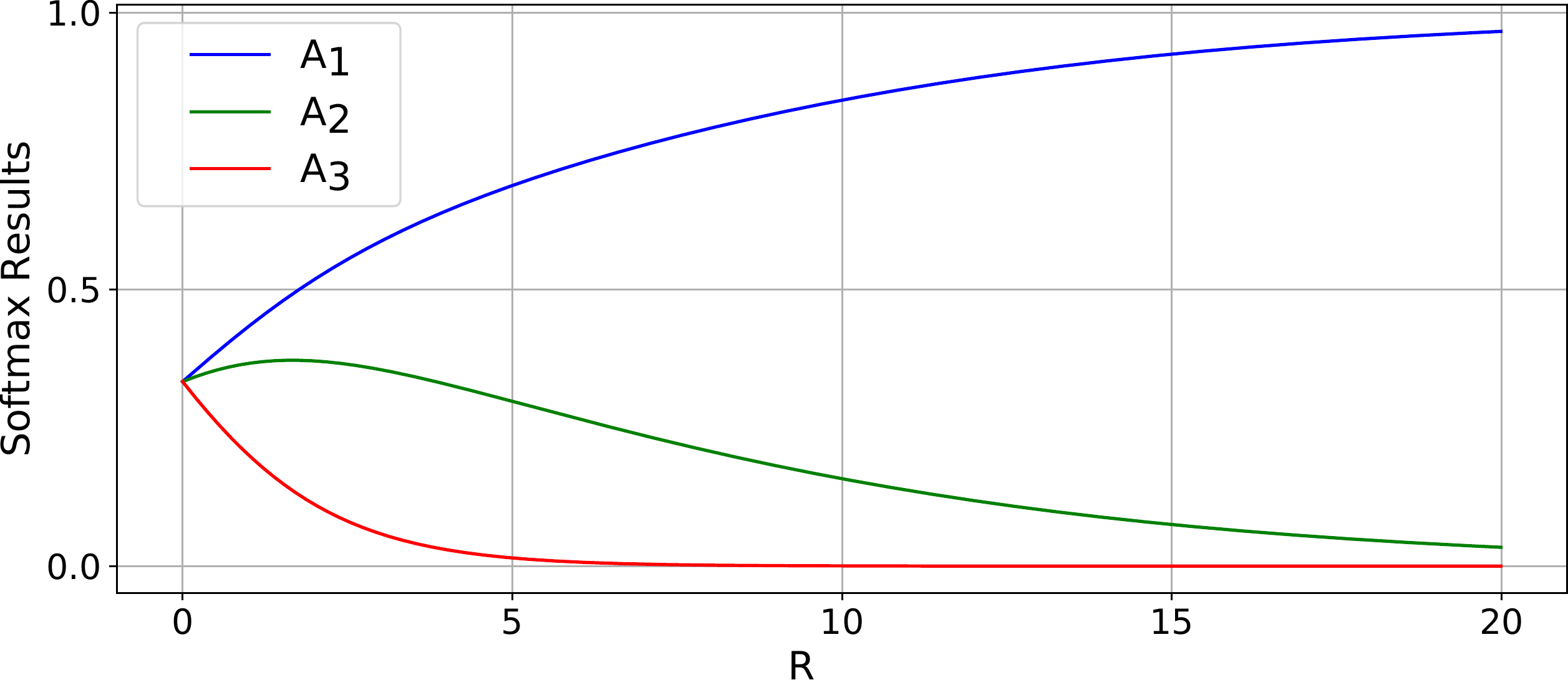}
        \hfill 
        \caption[]{}    
    	\label{fig:sub5e}
    \end{subfigure}
    \begin{subfigure}[b]{0.325\textwidth}
        \centering 
        \includegraphics[width=0.7\textwidth]{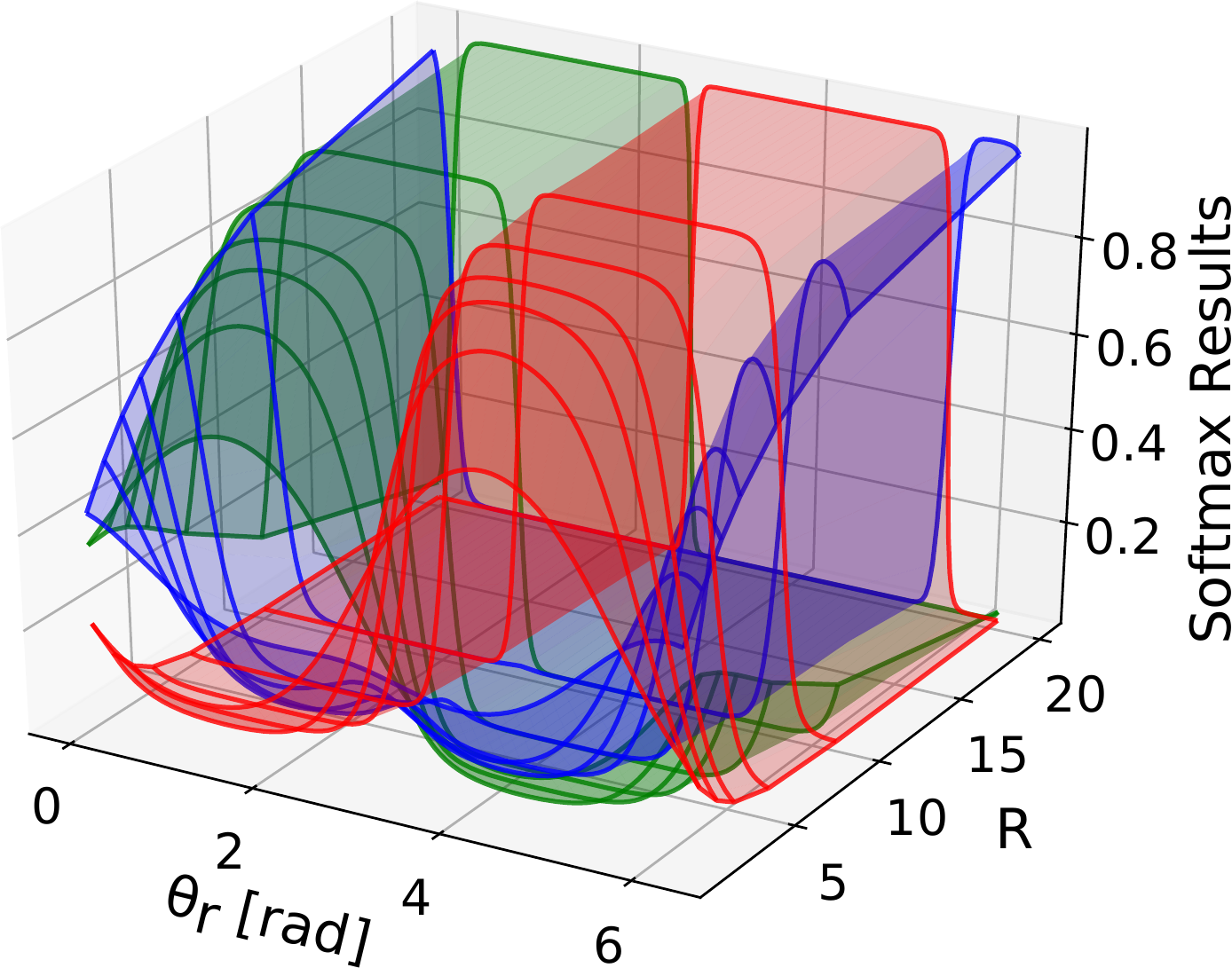}
        \hfill 
        \caption[]{}    
        \label{fig:sub5f}
    \end{subfigure}
    \caption[]
    {\small The diagrams of the output's neurons for an input sample of class $A_1$ as a function of: (a) $\theta_r$, (b) $R$ and (c) as a multi-variable function, as well as the corresponding diagrams of the Softmax outputs over: (a) $\theta_r$, (b) $R$ and (c) both of them.  Note that within a range centered around $\theta_r=0$, the feature vector's membership in the prevalent class is evident.} 
    \label{fig:5}
\end{figure*}

Thus, considering the feature vector $\bar{a}_e$ and the dominant class $i$, in the subspace of which $\bar{a}_e$ lies, we calculate the gradient of the softmax output of the $j$-th output neuron ($\nabla S_j$), as a function of two partial derivatives.
The first refers to the partial derivative over the feature vector's norm $R=\|\bar{a}_e\|$ and the second one over its angle $\theta_i$ with the weight $\bar{w}_i$  of the corresponding prevailing class (see Appendix~A):
\begin{equation}\label{eq:SofGrad}
\nabla S_j(dR,d\theta_i) = \frac{\partial S_j}{\partial R}dR+\frac{\partial S_j}{\partial \theta_i}d\theta_i,
\end{equation}
where:
\begin{equation}\label{eq:SofNorm}
\frac{\partial S_j}{\partial R} = \frac{S_j}{\sum_{k=0}^{N}{e^{z_k}}}\sum\limits_{k=0}^{N}{\left(\left[\cfrac{\partial z_j}{\partial R}-\cfrac{\partial z_k}{\partial R}\right] e^{z_k}\right)},
\end{equation}
\begin{equation}\label{eq:SofAng}
\frac{\partial S_j}{\partial \theta_i} = \frac{S_j}{\sum_{k=0}^{N}{e^{z_k}}}\sum\limits_{k=0}^{N}{\left(\left[\cfrac{\partial z_j}{\partial \theta_i}-\cfrac{\partial z_k}{\partial \theta_i}\right] e^{z_k}\right)}.
\end{equation}
Hence, the partial derivatives of each output neuron $z_j$ over $R$ and $\theta_i$ have to be calculated.

\subsubsection{Simple Case}
Aiming to calculate the term $\partial z_j/ \partial \theta_i$ in Eq.~{\ref{eq:SofAng}}, we will examine the simplistic scenario of classifying a feature vector $\bar{a}_e\in\mathbb{R}^3$  among three possible classes $A_1$, $A_2$ and $A_3$, with $A_1$ being the prevailing one, \textit{i.e.}, $i=1$ in Eq.~{\ref{eq:SofAng}}.
We will focus our derivations on the response and weights of class $A_2$, indicating that $j=2$ in Eq.~{\ref{eq:SofAng}}. Section~{\ref{subsub:gen}} presents our findings extended to a generalized model. 

Note that $z_2$ constitutes the output of the second neuron in the output layer expressed as a function of $\theta_2$ according to the common dot product rule:
\begin{equation}
z_2 = \bar{a}_e \cdot \bar{w}_2 = \|\bar{a}_e\| \|\bar{w}_2\| \cos\theta_2 = R \|\bar{w}_2\| \cos\theta_2.
\end{equation}
Then, in order for the term $\partial z_2 / \partial \theta_1$ to be computed, the above dot product needs to be written as a function of $\theta_1$ instead of $\theta_2$.
With the terms $\theta_1$ and $\theta_2$, we refer to the angles $\widehat{(\bar{a}_e,\bar{w}_1)}$ and $\widehat{(\bar{a}_e,\bar{w}_2)}$, respectively, where a slight change of $\theta_2$ affects $\theta_1$, as well.
We observe that variations of both $R$ and $\theta_1$ take place onto a unit plane $\hat{P}$ that is defined by $\bar{a}_e$ and $\bar{w}_1$, as illustrated in Fig.~\ref{fig:RP}.
Thus, we decompose $\bar{w}_2$ into two parts, one orthogonal ($\bar{w}_{2\bot}$) to $\hat{P}$ and one collinear ($\bar{w}_{2\parallel}$) with it.
Then, we can write:
\begin{equation}
z_2 = \bar{a}_e\cdot(\bar{w}_{2\parallel}+\bar{w}_{2\bot})=\bar{a}_e\cdot\bar{w}_{2\parallel}+\bar{a}_e\cdot\bar{w}_{2\bot}=\bar{a}_e\cdot\bar{w}_{2\parallel}
\end{equation}
since the dot product of the orthogonal component is always zero.
Considering the above, a change in $R$ and $\theta_1$ affects only the collinear part $\bar{w}_{2\parallel}$.
Thus, the dot product can be described by exploiting only terms being collinear with $\hat{P}$, as follows:
\begin{equation}\label{eq:OpDotSim}
z_2 = R \|\bar{w}_{2\parallel}\| \cos(\theta_1-\phi_2),
\end{equation}
where $\phi_2$ is the angle between the collinear with $\hat{P}$ vectors $\bar{w}_{2\parallel}$ and $\bar{w}_{1}$.
As a consequence, the norm of the projected weight onto the plane of variations defines, along with the norm $R$ of the feature vector, the amplitude of the dot product, \textit{i.e.}, its maximum possible value.
Correspondingly, the angle $\widehat{(\bar{w}_2,\bar{w}_1)}$ sets the phase difference  ($\phi_2$), in which the maximum value of the dot product is met.
Note that for the dominant class $A_1$, it is $\|\bar{w}_{1\parallel}\|=\|\bar{w}_{1}\|$, $\phi_1=0$ and thus:
\begin{equation}
z_1 = R  \|\bar{w}_1\|  \cos\theta_1,
\end{equation}
while 
\begin{equation}
z_3 = R \|\bar{w}_{3\parallel}\| \cos{(\theta_1-\phi_3)}.
\end{equation}

We observe that each output changes linearly with $R$ and sinusoidally with $\theta_1$.
Thus, we can conceive our example's neurons' outputs $z_1$, $z_2$ and $z_3$, as three sinusoidal waves of:
\begin{itemize}
\item common frequency,
\item different and linearly dependent on $R$ amplitude and
\item different and independent from $\bar{a}_e$ phase shift.
\end{itemize}
The above properties for our simple case are depicted in Fig.~\ref{fig:sub5a} and Fig.~\ref{fig:sub5b} with the horizontal axis representing the feature vector's angle of rotation ($\theta_r$) onto the plane of variations and the value of $R$, respectively.
Fig.~\ref{fig:sub5c} displays the diagram of the neurons' output as function of $\theta_r$ and $R$.

By further exploiting Eq.~\ref{eq:1}, we can calculate the classification outputs $y_1, y_2,$ and $y_3$ for each value of $\theta_r$ and $R$ from the above distributions, producing the corresponding graphs of Fig.~\ref{fig:sub5d},~\ref{fig:sub5e} and~\ref{fig:sub5f}.
Using depicted responses in Fig.~\ref{fig:sub5d} for each value of $\theta_r$, we can extract the prevailing class, which is changed under rotation.
Consequently, Fig.~\ref{fig:sub5e} verifies that the prevailing class can not be affected by modifications of $R$.
On the contrary, variations in rotation can not reach high softmax results without suitable modification of $R$.
According to Fig.~\ref{fig:sub5f}, the bigger the $R$, the closer the maximum softmax value to 100\%.
In addition, while $R$ increases, the graph over $\theta_r$ of the produced feature vector approximates the rectangular function, thus increasing the range of its orientations that secure high softmax results.
The above property will be particularly shown in our experiments.

Finally, from Eq.~\ref{eq:OpDotSim}, it follows that:
\begin{equation}\label{eq:DerDotR}
\frac{\partial z_2}{\partial R} = \|\bar{w}_{2\parallel}\| \cos{(\theta_1-\phi_2)},
\end{equation}
\begin{equation}
\frac{\partial z_2}{\partial \theta_1} = -R \|\bar{w}_{2\parallel}\| \sin{(\theta_1-\phi_2)}.
\end{equation}
Combining the above with Eq.~\ref{eq:SofGrad}, \ref{eq:SofNorm} and \ref{eq:SofAng}, we define the gradient of softmax over $R$ and $\theta_i$.

\subsubsection{Generalizing}\label{subsub:gen} 
\begin{figure}
    \centering
    \includegraphics[width=0.8\linewidth]{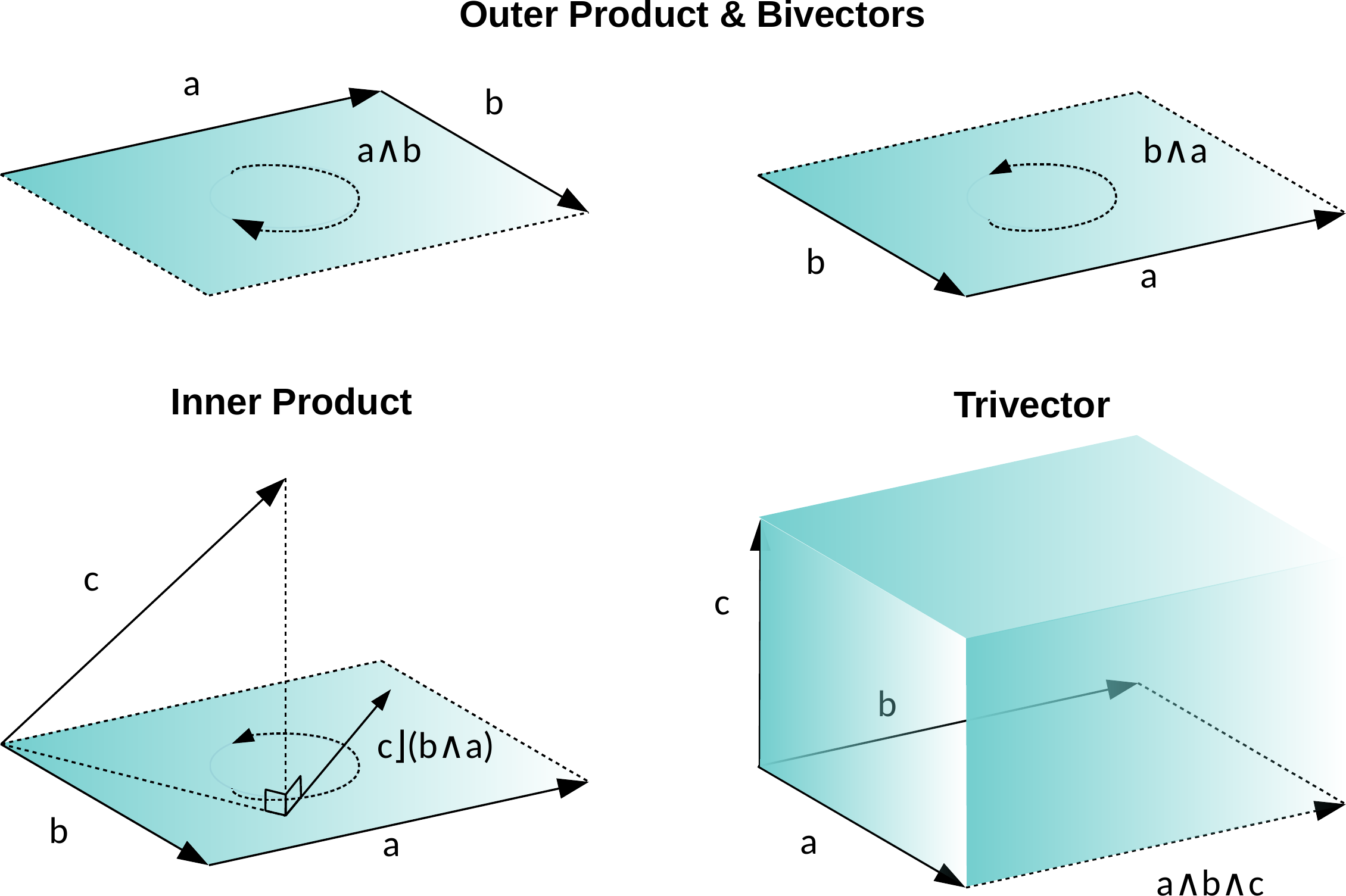}
    \caption{\small Clifford Algebra's basic components and operations.}
        \label{fig:Cl}
\end{figure}

In order to apply the above calculations in any generic space $\mathcal{F}$, some geometrical properties need to be defined.
To that end, we base our model on the Clifford algebra~\cite{clifford1882classification} and more specifically its power of geometrically manipulating objects in high dimensions\footnote{library: \textit{https://github.com/IoannisKansizoglou/Deep-Feature-Space}}.
Its basic components include vectors, bivectors and trivectors, while its basic geometrical product denoted with a simple juxtaposition between two compnents, say $v_1$ and $v_2$, emerges from the summation of other two products, \textit{viz.} the inner ($\rfloor$) and the outer ($\wedge$) products~\cite{franchini2010brief}:
\begin{equation}
{v}_1{v}_2 = {v}_1 \rfloor{v}_2 +{v}_1\wedge{v}_2.
\end{equation}
An illustration of the above components is shown in Fig.~\ref{fig:Cl}.
Here, we will keep in mind that the rotation of a feature vector $\bar{a}_e\in\mathcal{F}$ can be described as:
\begin{equation}
\bar{a}_e' = V\bar{a}_eV^{\dagger},
\end{equation}
where:
\begin{equation}
V = e^{-\hat{P}\frac{\theta_r}{2}} = cos{\frac{\theta_r}{2}} - sin{\frac{\theta_r}{2}}\hat{P},
\end{equation} 
is called rotor and rotates the vector $\bar{a}_e$ on the unit plane ($\hat{P}$) by the angle ($\theta_r$).
With $V^{\dagger}$ we refer to the reverse of $V$:
\begin{equation}
V^{\dagger} = e^{+\hat{P}\frac{\theta_r}{2}} = cos{\frac{\theta_r}{2}} + sin{\frac{\theta_r}{2}}\hat{P}.
\end{equation}
Consequently, the projection of any weight vector $\bar{w}_{j\parallel}$ onto the plane of $\bar{w}_i$ and $\bar{a}_e$ is calculated as follows.
Initially, through the wedge product $\bar{a}_e\wedge\bar{w}_i$, we calculate the unit bivector:
\begin{equation}
\hat{P} = \frac{\bar{a}_e\wedge\bar{w}_i}{\|\bar{a}_e\wedge\bar{w}_i\|},
\end{equation}
referring to the plane of interest.
Then, we apply:
\begin{equation}\label{eq:ClifRot}
\bar{w}_{j\parallel} = V(\bar{w}_j\rfloor \hat{P})V^{\dagger}.
\end{equation}
In Eq.~\ref{eq:ClifRot}, the inner product is exploited to calculate the complement (within the subspace of $\hat{P}$) of the orthogonal projection of $\bar{w}_j$ onto $\hat{P}$~\cite{suter2003geometric}.
Then, a counter-clockwise rotation of the produced vector by $\frac{\pi}{2}$ over $\hat{P}$ is performed through the rotor:
\begin{equation}
\begin{gathered}
V = \cos\frac{\pi}{4}-\sin\frac{\pi}{4}\hat{P}=\frac{\sqrt{2}}{2}(1-\hat{P}), \\
V^{\dagger} = \frac{\sqrt{2}}{2}(1+\hat{P}).
\end{gathered}
\end{equation}
Once $\|\bar{w}_{j\parallel}\|$ is calculated, we have:
\begin{equation}\label{eq:OpDot}
z_j =R \cdot \|\bar{w}_{j\parallel}\|\cdot \cos{(\theta_i-\phi_j)},
\end{equation}
and
\begin{equation}\label{eq:DerDot}
\begin{gathered}
\frac{\partial z_j}{\partial R} = \|\bar{w}_{j\parallel}\| \cos{(\theta_i-\phi_j)},\\
\frac{\partial z_j}{\partial \theta_i} = -R \|\bar{w}_{j\parallel}\| \sin{(\theta_i-\phi_j)}.
\end{gathered}
\end{equation}
For $j=i$, $\|\bar{w}_{j\parallel}\|=\|\bar{w}_{j}\|$ and $\phi_j=0$.

\subsection{Findings Overview} \label{sub:FinOv}

At this stage, we take a moment to discuss the findings of Sections~{\ref{sub:SpDiv}} and~{\ref{sub:SenAnal}} and highlight their importance to the consequent study.
To begin with, the geometrical interpretation, provided in Section~{\ref{sub:SpDiv}}, claims that the locus of a class is defined exclusively by the orientation of the vectors, while it constitutes a convex set of the space.
Hence, the feature vectors of a given class are anticipated to concentrate inside the corresponding locus, by angularly approximating themselves.
The sensitivity analysis conducted in Section~{\ref{sub:SenAnal}} further clarifies the higher impact of the feature vectors' orientation.
To this end, the resulting mathematical formulation enables us to investigate the gradients of softmax over the feature vectors' orientation and norm, providing an initiative intuition regarding the gradients' magnitude in Fig.~{\ref{fig:5}}.
In Section~{\ref{sub:NorOr}}, this formulation will be exploited to empirically prove the orientation's supremacy in practical scenarios.

\subsubsection{General Rules of Feature Space}\label{sub:RulFS}

Summing up, we write below certain properties that describe the studied feature space.
\begin{itemize}
\item \textit{R.I:} Each class occupies a convex subspace of the feature space defined and angularly delimited by its differential vectors (Eq.~\ref{eq:24}). 
\item \textit{R.II:} The prevalent class of each feature vector is exclusively defined by its orientation (Fig.~\ref{fig:5}).
\item \textit{R.III:} The softmax output of the prevalent class for a feature vector is proportional to its norm.
\end{itemize}
The proof of \textit{R.III} is given in Appendix~B.

\subsubsection{VC-dimension in $\mathcal{F}$}

Subsequently, we investigate a possible relation between the studied feature space $\mathcal{F}$ and the Vapnik-Chernovenkis (VC) dimension~{\cite{blumer1989learnability}}, which can provide a formal way to define the capacity of $\mathcal{F}$.
In our study, we seek the maximum number $d\in\mathbb{N}^{*}$ such that there is a possible layout of $d$ feature vectors in $\mathcal{F}$, for which the hypothesis $\mathcal{H}$, {\text{i.e.}}, the set of separation hyperplanes defined in Section~{\ref{sub:SpDiv}}, can classify the above vectors for any possible label assignment among them.
In such a case, we say that $\mathcal{H}$ shatters $d$ points in $\mathcal{F}$.
Knowing the properties of $\mathcal{F}$, we firstly proceed with several important observations.
According to the angular property, we know that the norm of the feature vectors does not affect the classification problem.
Thus, in order to define the VC-dimension of $\mathcal{F}$, we select feature vectors that lie upon a hypersphere manifold.
Furthermore, it is easy to show that in case that a binary classier is not able to shatter a specific layout of points in $\mathcal{F}$, then neither sets of three or more classifiers can.
To prove that, let us simply assume a layout of $d$ points in $\mathcal{F}$ that is shattered by a hypothesis $\mathcal{H}_3$ of three separation hyperplanes, \textit{viz.}, $\bar{w}_{12}$, $\bar{w}_{13}$ and $\bar{w}_{23}$.
Then, we can consider a labelling option of the layout, where only the labels of the first two classes are assigned and none of the third one.
Since $\mathcal{H}_3$ shatters the points and according to Eq.~{\ref{eq:22}}, we know that $D_{12}$ includes all the points with the label $1$.
Hence, $\bar{w}_{12}$ also separates the two classes and given their unbounded selection, we can claim the validity of our initial argument.
Thus, the VC-dimension of a binary classifier in $\mathcal{F}$ sets the upper bound of any hypothesis in that feature space.
In this case, our system degenerates to the simple case of a perceptron, or more typically, of a linear classifier, with the only differentiation being the norm restriction due to the angular property.
However, the above restriction does not affect the proof of VC-dimension for linear classifies, leading to the conclusion that $d$=$N$+$1$, where $N$ denotes the dimensionality of $\mathcal{F}$.

\subsection{Metrics} \label{sub:metr}

Taking into consideration the supremacy of the feature vectors' orientation in the classification outcome, discussed in Section~{\ref{sub:FinOv}}, we draw useful conclusions about the form of overfitting in $\mathcal{F}$.
As a whole, overfitting is mainly realized via the divergence between the performance curves of the training and the evaluation data.
Keeping in mind that the evaluation set remains agnostic to the model's parameters, the separability of its feature vectors is expected to differ from the training ones.
The above fact tends to become more intense when the performance curves between the two sets diverge and a larger amount of testing data are misplaced in $\mathcal{F}$.
As stated in Section~{\ref{sub:FinOv}}, since the vectors' orientation constitutes the decisive factor, the above misplacement is anticipated to be captured by observing their angular distribution in $\mathcal{F}$.
Ergo, aiming to review the nature of overfitting in $\mathcal{F}$, we propose two metrics that describe the angular distribution of all $\bar{a}_e\in\mathcal{F}$, \textit{viz.}, \textit{centrality} ($\mathcal{C}$) and \textit{separability} ($\mathcal{S}$).
These metrics focus only on the angular distribution of $\bar{a}_e$, exploiting the cosine distance metric $d_c\in[0,1]$, where:
\begin{equation}\label{eq:CosSim}
d_c(\bar{a}_{e_1},\bar{a}_{e_2}) = 1 - \frac{\bar{a}_{e_1}\cdot{\bar{a}_{e_2}}}{\|\bar{a}_{e_1}\| \|\bar{a}_{e_2}\|}.
\end{equation}

\subsubsection{Centrality ($\mathcal{C}$)}
This metric assesses the quality regarding the angular distribution between the central vectors of the target classes.
Each central vector derives from the mean value of the normalized $\hat{a}_{e}$ belonging to the corresponding target class.
In specific, let us consider a set of $N_A$ classes  $A = \{ A^{(i)} \}_{i=1}^{N_A}$, such that $\forall i,A^{(i)}$ exists a set of $N^{(i)}$ feature vectors $a_{e}^{(i)}=\{ \bar{a}_{e_j}^{(i)} \}_{j=0}^{N^{(i)}}$.
The central vector $\bar{c}^{(i)}$ of $A^{(i)}$ is:
\begin{equation}
{\bar{c}}^{(i)} = \frac{1}{N^{(i)}}\sum_j^{N^{(i)}}{\hat{a}_{e_j}^{(i)}},
\end{equation}
where $\hat{a}_{e_j}^{(i)}$ denotes the $L_2$-normalized $\bar{a}_{e_j}^{(i)}$ and $j\in\mathbb{N}$ a simple index variable.
Then, the centrality $\mathcal{C}^{(i)}$ of the $i$-th class is $\mathcal{C}^{(i)} = \min_{k} \{ d_c(\bar{c}^{(i)},\bar{c}^{(k)})\}$, with $k=\{ 1,2,...,N_A \}$ and $k \neq i$.
Ergo, the centrality ratio between the test and train sets is:
\begin{equation}
\mathcal{C}_R = \frac{1}{N_A}\sum_{i=0}^{N_A}{\frac{{\mathcal{C}^{(i)}}_{test}}{{\mathcal{C}^{(i)}}_{train}}}.
\end{equation}

\subsubsection{Separability ($\mathcal{S}$)}
This metric evaluates the percentage of the target class's similarity over the rest of the classes in $\mathcal{F}$.
For the $i$-th class, we find its nearest class $A^{(j)}$, where $j = \argmin_{k} \{ d_c(\bar{c}^{(i)},\bar{c}^{(k)})\}, k\neq{i}$.
Then, for $k,l=\{ 0,1,...,N^{(i)} \} $, we calculate the mean cosine distance between the feature vectors of the same class:
\begin{equation}
\mathcal{I}^{(i)}_1 = \frac{1}{{N^{(i)}}^2}\sum_{k}^{N^{(i)}}{\sum_{l}^{N^{(i)}}{d_c(\bar{a}_{e_k}^{(i)},\bar{a}_{e_l}^{(i)})}}, \ \ k\neq l.
\end{equation}
Moreover, for $k=\{ 0,1,...,N^{(i)}\}$ and $l=\{ 0,1,...,N^{(j)}\}$, we calculate the mean cosine distance between the feature vectors of the $i$-th class with the ones of the $j$-th class:
\begin{equation}
\mathcal{I}^{(i)}_2 = \frac{1}{{N^{(i)}}{N^{(j)}}}\sum_{k}^{N^{(i)}}{\sum_{l}^{N^{(j)}}{d_c(\bar{a}_{e_k}^{(i)},\bar{a}_{e_l}^{(j)})}}.
\end{equation}
Then, the separability $\mathcal{S}^{(i)}$ of the $i$-th class is $\mathcal{S}^{(i)} = \mathcal{I}^{(i)}_1/\mathcal{I}^{(i)}_2$ and the separability ratio between the test and train sets is:
\begin{equation}
\mathcal{S}_R = \frac{1}{N_A}\sum_{i=0}^{N_A}{\frac{{\mathcal{S}^{(i)}}_{test}}{{\mathcal{S}^{(i)}}_{train}}}.
\end{equation}

The above findings and metrics are exploited to evaluate our salient argument, denoting that training a deep feature extractor with Softmax-based losses creates an angular discrepancy between the training and the evaluation set in the studied feature space.
Therefore, training a cascade learning algorithm with the same training set, used for the aforementioned extractor, further deteriorates the final performance on the evaluation set.
For this purpose, we empirically study the common case of fusing the outputs from two unimodal extractors in a cascade learning algorithm.
The fusion operation of two input feature vectors is simply denoted by their concatenation ($^\frown$).
In specific, considering two vectors $\bar{a}_e^V\in\mathbb{R}^{n}$ and $\bar{a}_e^A\in\mathbb{R}^{m}$ with $n, m\in\mathbb{N^{*}}$ their corresponding lengths, the resulting fused features ($\bar{a}^F_e \in\mathbb{R}^{n+m}$) conform to the equation $\bar{a}_e^F={\bar{a}_e^V}{^\frown} \bar{a}_e^A.$

\section{Experiments}

The properties in Section~{\ref{sec:Method}} allow us to present a novel geometrical interpretation with the purpose to denote the adverse effect of overfitting in feature learning tasks that exploit deep architectures as feature extractors.
Hence, the following empirical studies have been conducted to gradually conclude in the final real-world scenario of feature fusion.
More specifically, we begin by detailing the datasets used during experimentation, while several experiments are presented to demonstrate the dominance of feature vectors' orientation during training and the norm's contribution to the resulting distribution.
Consequently, we evaluate the ability of centrality and separability to capture overfitting in $\mathcal{F}$.
Finally, we proceed with the main argument of the paper, by firstly measuring overfitting in the unimodal feature extractors through the proposed metrics and then, empirically proving the validity of our statement through training a fusion algorithm  for a representative classification task of emotion recognition.
Our results prove that an alternative training strategy, which obeys the aforementioned guidelines, favors the overall system's performance in practical applications.

\subsection{Datasets}

\subsubsection{CIFAR}
Both CIFAR-10 and CIFAR-100 datasets~\cite{krizhevsky2009learning} include 50,000 training and 10,000 testing images of $32\times32$ pixels.
CIFAR-10 is composed of images from 10 and CIFAR-100 from 100 classes.

\subsubsection{RML}
The audio-visual RML dataset~\cite{wang2008recognizing} consists of 720 video samples, each one containing scripted scenarios that express a specific emotional context among six basic classes, \textit{viz.}, \textit{happiness}, \textit{sadness}, \textit{surprise}, \textit{anger}, \textit{fear} and \textit{disgust}, from 8 different subjects.
The recording speed of the $720\times480$ video frames is 30 fps, while the sampling rate of the audio mono channel is 22.05 kHz.
During our experiments, we adopt the Leave-One-Speaker-One (LOSO) strategy to evaluate the speaker-invariant capabilities of each approach~\cite{schuller2010interspeech}.

\subsubsection{BAUM-1s}
The audio-visual BAUM-1s dataset~\cite{zhalehpour2016baum} contains 1184 video samples with spontaneous unscripted scenarios from 31 different subjects.
Since the specific dataset includes more emotional states than RML, we keep for cohesion only the 521 video samples that expose one of the aforementioned basic emotions.
The size of the video frames is $854\times480$ recorded with a speed of 30 fps, while the sampling rate of the audio signal is set to 48 kHz.
In order to assess the speaker-invariant capabilities throughout our experiments, we follow the Leave-One-Speakers-Group-Out (LOSGO) scheme of 6 groups that include each speaker once~\cite{zhalehpour2016baum}.

\subsubsection{eNTERFACE'05}
The audio-visual eNTERFACE'05 dataset~\cite{martin2006enterface} is composed of samples with acted scenarios from 41 subjects.
Those recordings expose one of the six basic emotions along with \textit{neutral}, which is discarded.
The camera's recording speed is 25 fps and the sampling rate of the audio signal is 48 kHz.
The experiments on eNTERFACE'05 has been also conducted according to the LOSGO scheme.

\subsubsection{ShapeNet}
The empirical evaluation between the distribution of the input data and the division in $\mathcal{F}$ is conducted, by exploiting the ShapeNet dataset~{\cite{yi2016}}, which includes 31,963 per-point labelled 3D shape collections of $16$ distinct categories.
During our study, we exploit at each experiment the 3D shapes among one of the categories: \textit{earphone}, \textit{rocket}, \textit{guitar}, \textit{table}, \textit{skateboard}, \textit{airplane} and \textit{car}.
The 3D shapes are divided into  training, evaluation and testing sets, in accordance to~{\cite{yi2016}}.

\subsubsection{ImageNet32}
In order to further evaluate the efficacy of the proposed metrics in capturing the nature of overfitting in $\mathcal{F}$, the large-scale ImageNet32 dataset~{\cite{chrabaszcz2017downsampled}} is employed.
ImageNet32 constitutes a downsampled version of the common ImageNet dataset~{\cite{deng2009imagenet}}, where each image is resized to the resolution of $32\times 32$.
Note that the total number of training and testing samples remains 1,281,167 and 50,000, respectively, as well as the total number of classes, which equals to 1,000.

\subsubsection{Pre-processing}
All images from CIFAR-10 and CIFAR-100 datasets are normalized based on the channel means and standard deviations.
Regarding the three emotional datasets, we follow the strategy proposed in our previous work~\cite{kansizoglou2019active}.
In specific, we detect the timestamps $t_c$ during which the subject speaks in every video sample.
For every $t_c$, we extract the corresponding frame ($f_n$) along with an audio sample within a time window $T_n=[t_c-t_w,t_c+t_w]$ with $t_w=500$ms.
The next $t_c$ is searched after $100$ms.
Then, a face image is cropped from each frame with the Haar Feature-Based Cascade detector~\cite{viola2001rapid}, while constantly keeping a fixed distance between the subjects' eyes of 55 pixels and then resized to the final size of $224\times224\times3$.
During training, we apply random adjustments on the face images' saturation and brightness.
The audio samples are resampled to a 16 kHz format and converted to a log-mel spectrogram representation of 64 mel-spaced frequency bins and a range of 125 Hz-7,5 kHz.
This is achieved through the short-time Fourier transform with Hann windows with 25ms size and a step of 10ms.
Finally, by keeping the $96$ middle columns, we set the size of the representation at $64\times96\times1$~\cite{kansizoglou2019active}.

\begin{figure}
    \centering
    \includegraphics[width=0.9\linewidth]{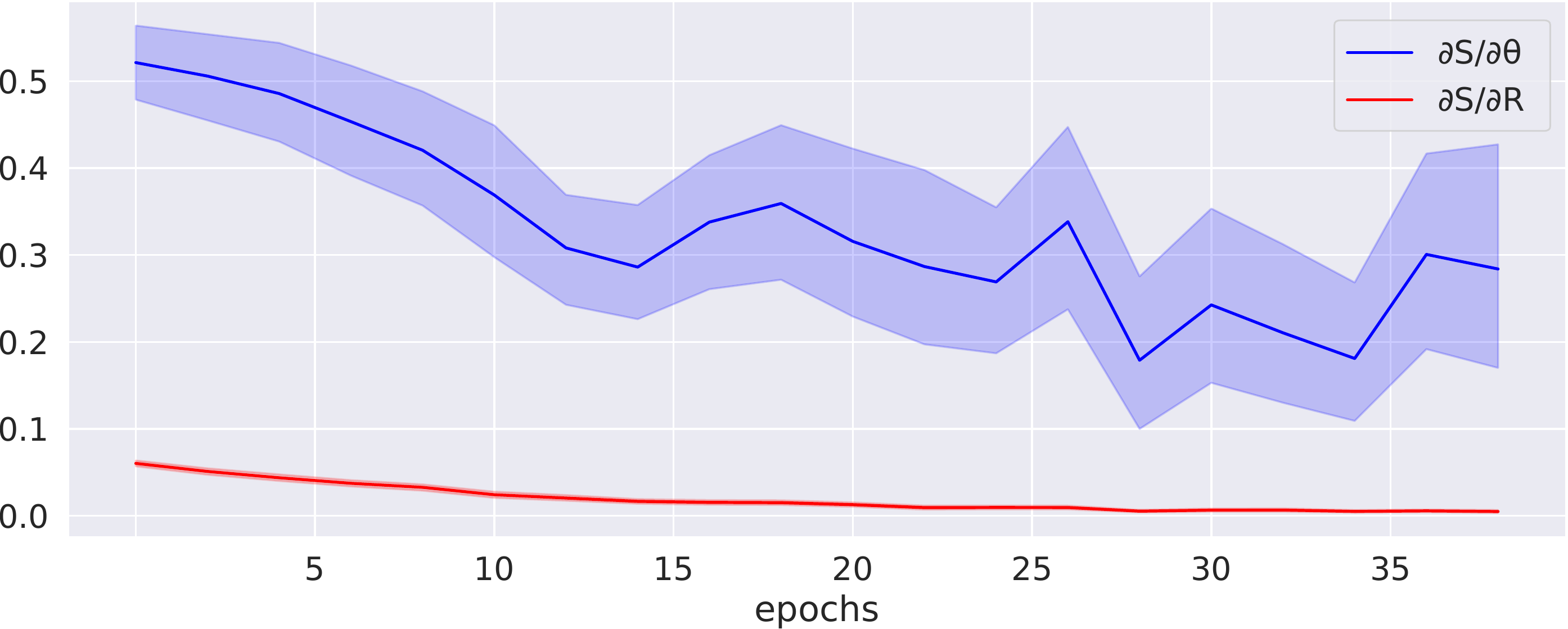}
    \caption{\small The mean and standard deviation of $\partial S_j / \partial R$ and $ \partial S_j / \partial \theta_i$ during training on CIFAR-10~{\cite{krizhevsky2009learning}}.}
        \label{fig:Grad}
\end{figure}

\begin{table}
\centering
\caption{Convolutional architecture.}
\label{table:SimpArchit}
\resizebox{0.8\linewidth}{!}{%
\renewcommand{\arraystretch}{1.}
\begin{tabular}{c|c|c|c}
 Layers & Kernel Size & Kernels & Output Size \\
 \hline
 Convolution & $[5\times5]$ & $20$ & $28\times28$ \\ 
 ReLU & - & - & $28\times28$ \\ 
 MaxPooling & $[2\times2]$ & - & $14\times14$ \\ 
 Convolution & $[5\times5]$ & $50$ & $10\times10$ \\ 
 ReLU & - & - & $10\times10$ \\ 
 MaxPooling & $[2\times2]$ & - & $5\times5$ \\ 
 Convolution & $[5\times5]$ & $500$ & $1\times1$ \\ 
 Fully Connected & - & $100$ & $100$ \\
 ReLU & - & - & $100$ \\
 Fully Connected & - & $10$ & $10$ \\
 ($L_2$-)Softmax & - & - & $10$ \\
\end{tabular}}
\end{table}

\subsection{Norm and Orientation in $\mathcal{F}$}
\label{sub:NorOr}

\begin{figure*}
    \centering
	\begin{subfigure}[b]{0.15\textwidth}   
        \centering 
        \includegraphics[width=0.85\textwidth]{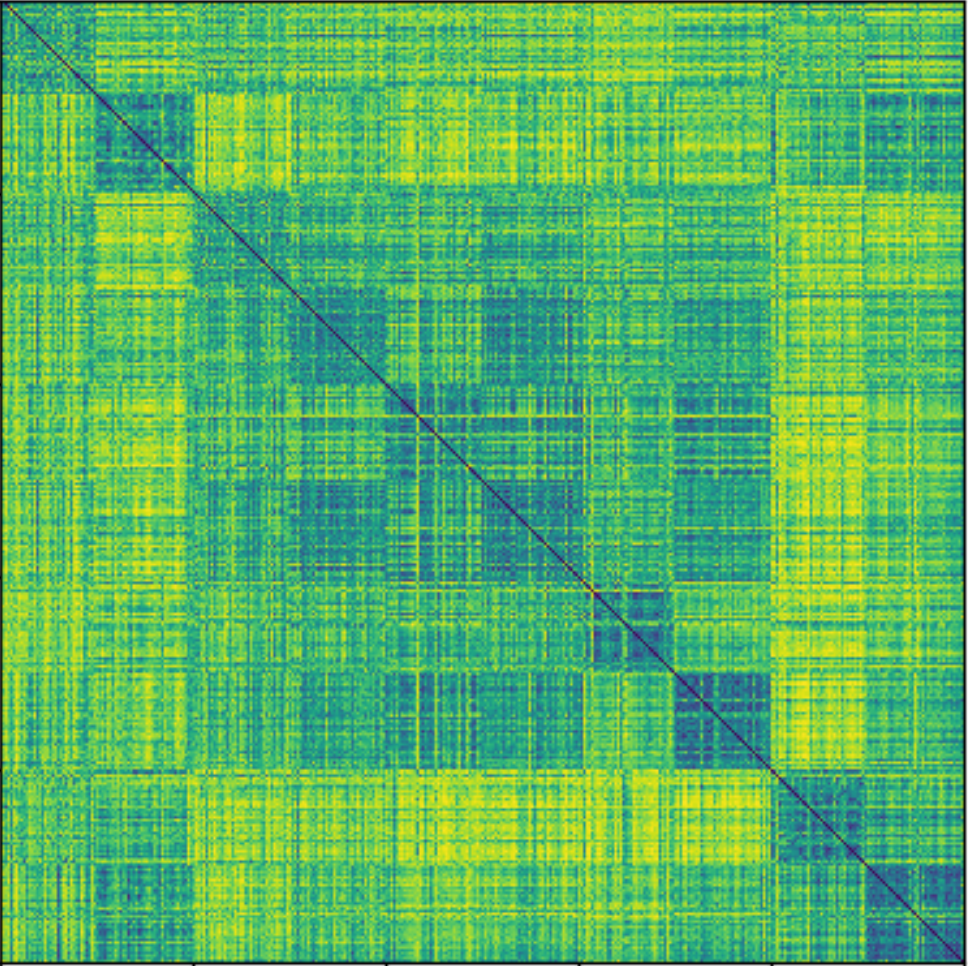}
        \caption[]{Softmax}    
    	\label{fig:sub6a}
    \end{subfigure}    
    \begin{subfigure}[b]{0.15\textwidth}
        \centering
        \includegraphics[width=0.85\textwidth]{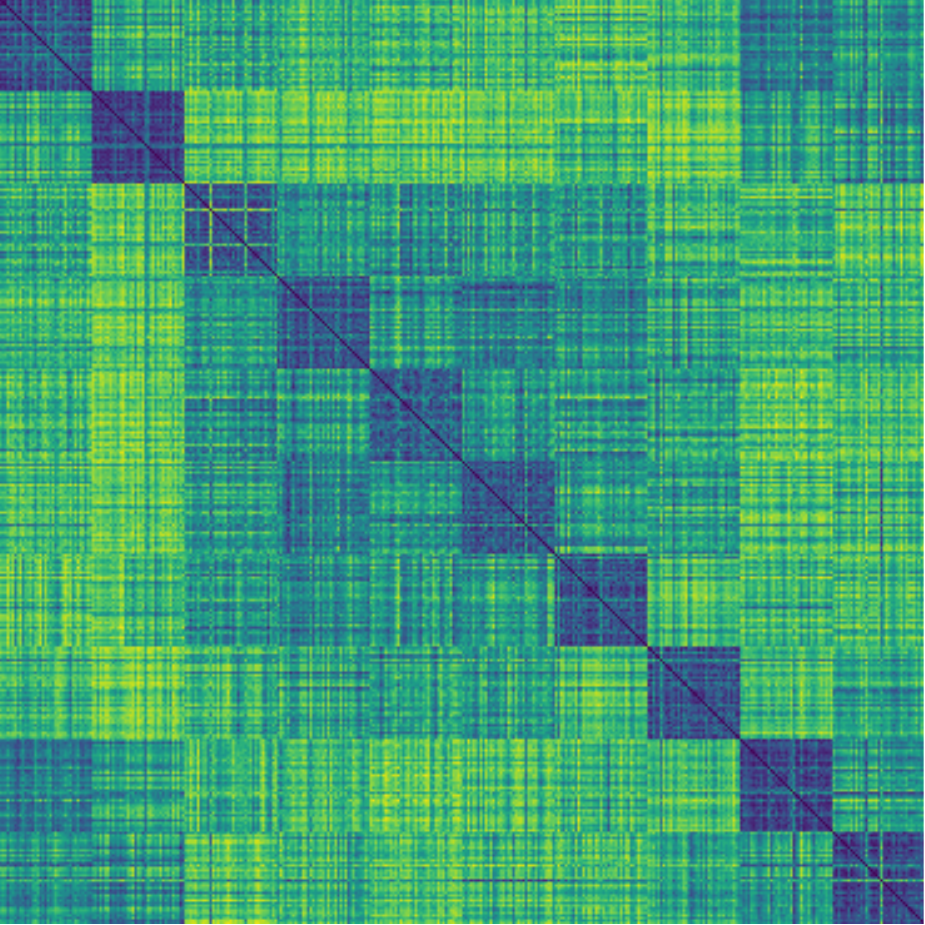}
        \caption[]{$L_2$-Soft.($s$=$20$)}    
        \label{fig:sub6b}
    \end{subfigure}
    \begin{subfigure}[b]{0.15\textwidth}
        \centering 
        \includegraphics[width=0.85\textwidth]{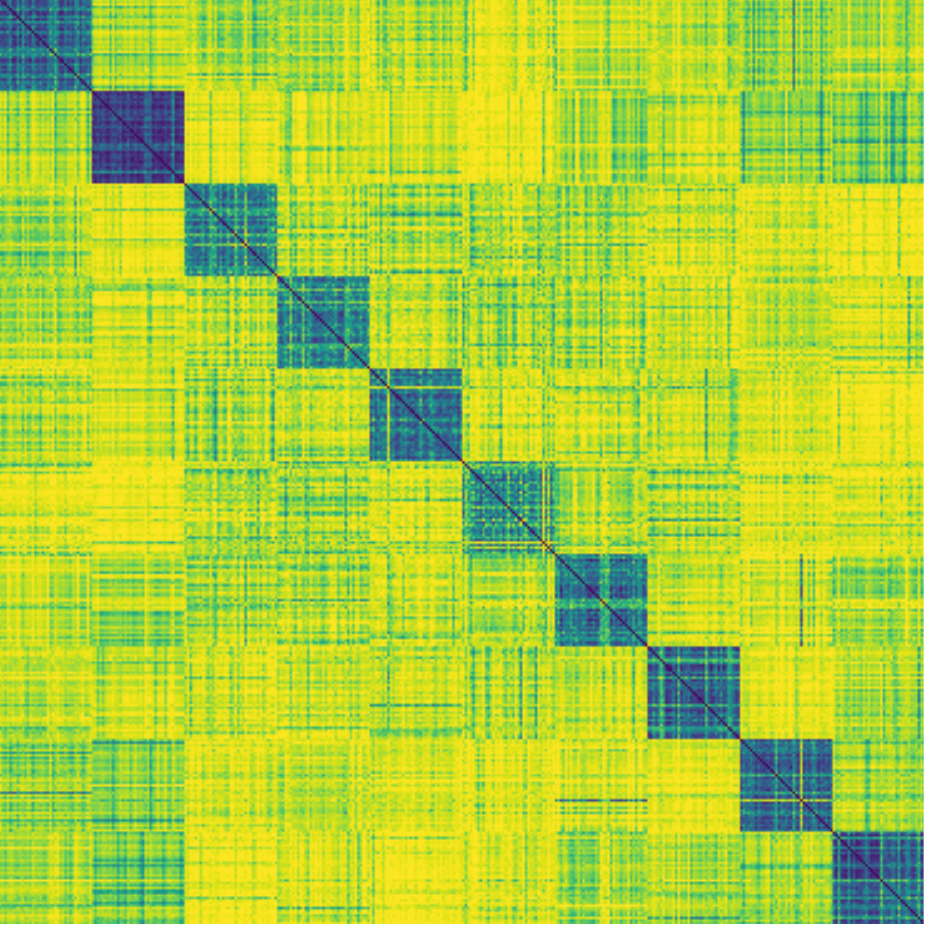}
        \caption[]{$L_2$-Soft.($s$=$1$)}    
        \label{fig:sub6c}
    \end{subfigure}
    \begin{subfigure}[b]{0.15\textwidth}   
        \centering 
        \includegraphics[width=0.85\textwidth]{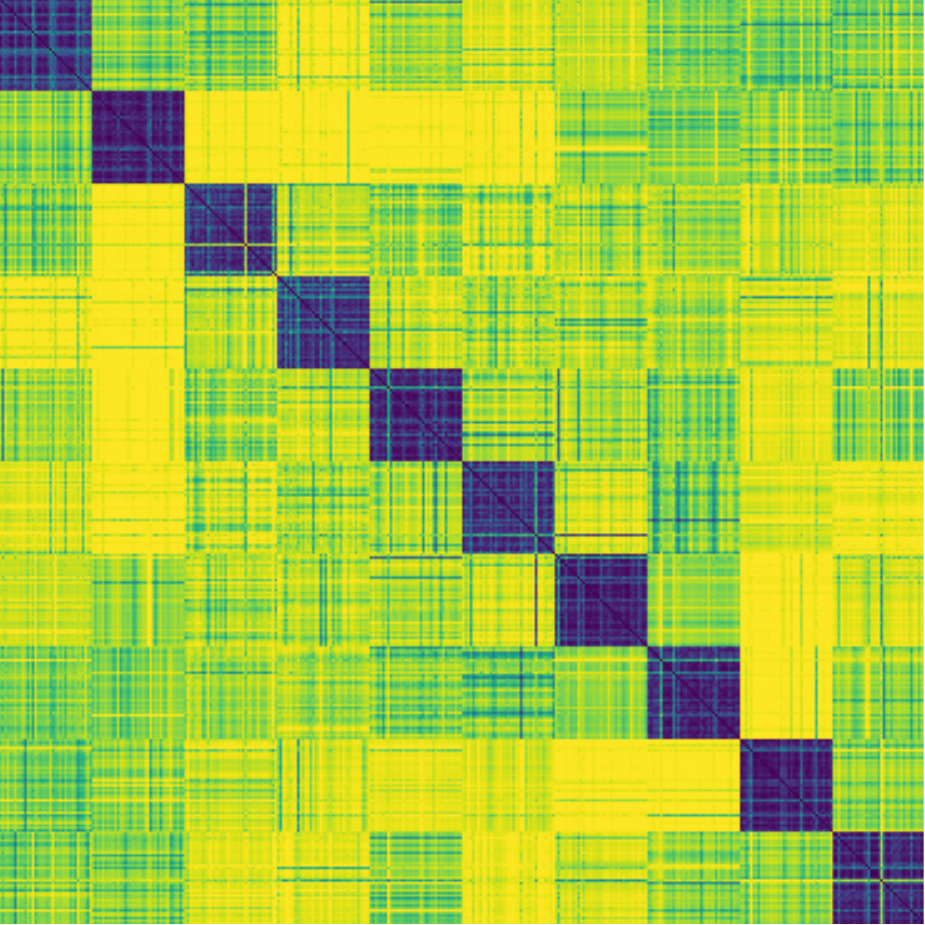}
        \caption[]{$L_2$-Soft.($s$=$0.1$)}    
        \label{fig:sub6d}
    \end{subfigure}
	\begin{subfigure}[b]{0.15\textwidth}   
        \centering 
        \includegraphics[width=0.85\textwidth]{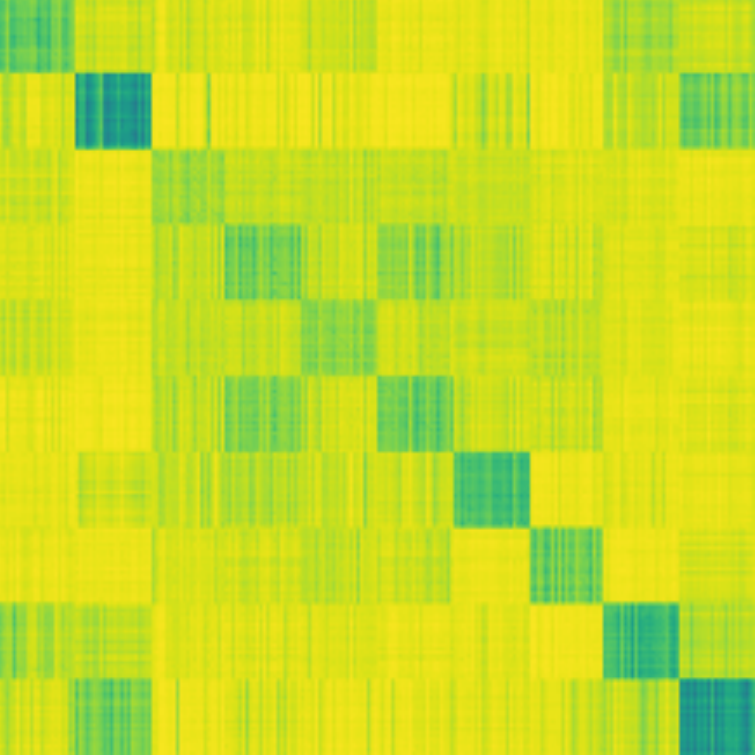}
        \caption[]{$L$-Softmax}
        \label{fig:sub6e}
    \end{subfigure}
    \begin{subfigure}[b]{0.15\textwidth}   
        \centering 
        \includegraphics[width=0.85\textwidth]{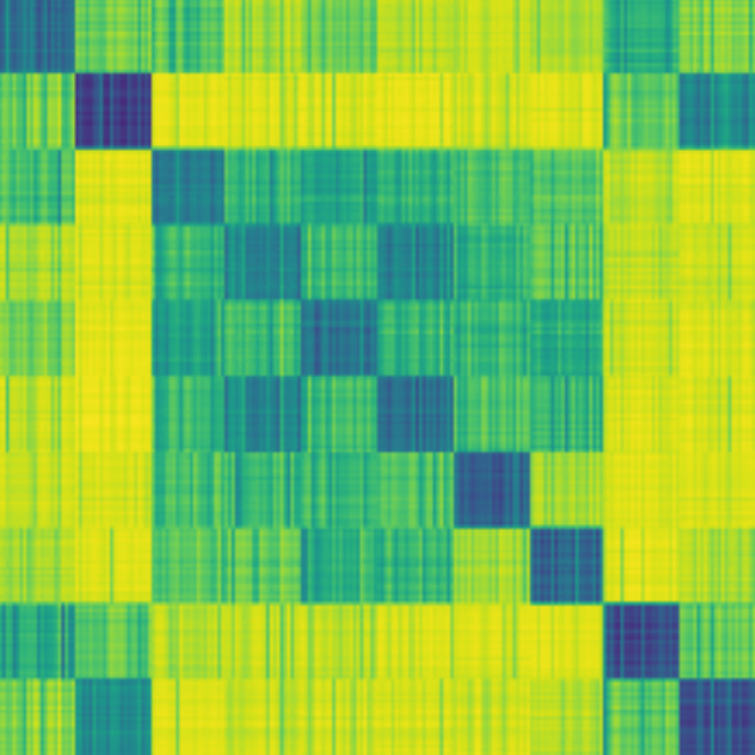}
        \caption[]{SphereFace}
        \label{fig:sub6f}
    \end{subfigure}
	\vskip\baselineskip
    \centering
	\begin{subfigure}[b]{0.15\textwidth}   
        \centering 
        \includegraphics[width=0.85\textwidth]{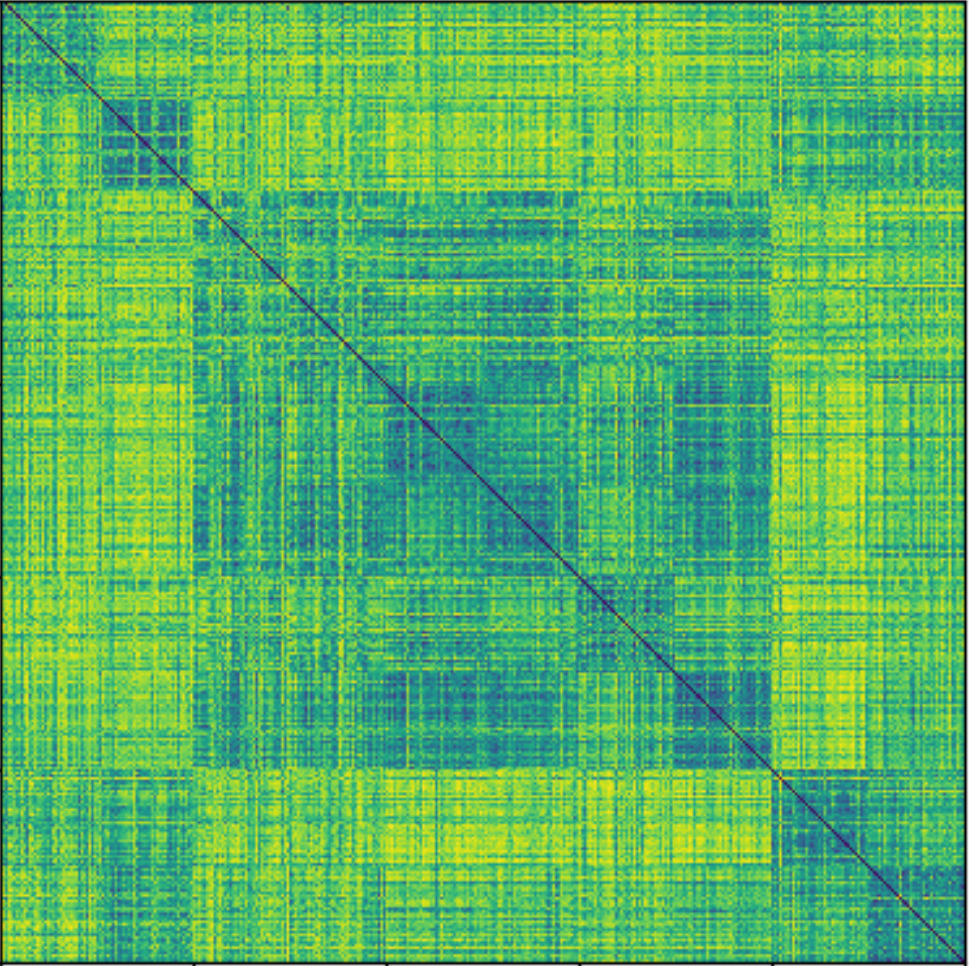}
        \caption[]{Softmax}    
    	\label{fig:sub6g}
    \end{subfigure}    
    \begin{subfigure}[b]{0.15\textwidth}
        \centering
        \includegraphics[width=0.85\textwidth]{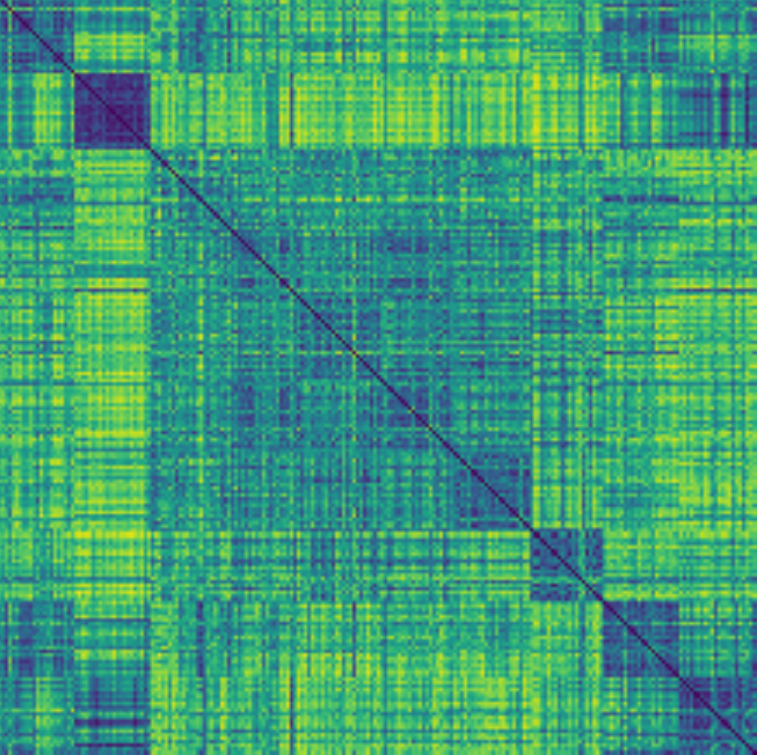}
        \caption[]{$L_2$-Soft.($s$=$20$)}    
        \label{fig:sub6h}
    \end{subfigure}
    \begin{subfigure}[b]{0.15\textwidth}
        \centering 
        \includegraphics[width=0.85\textwidth]{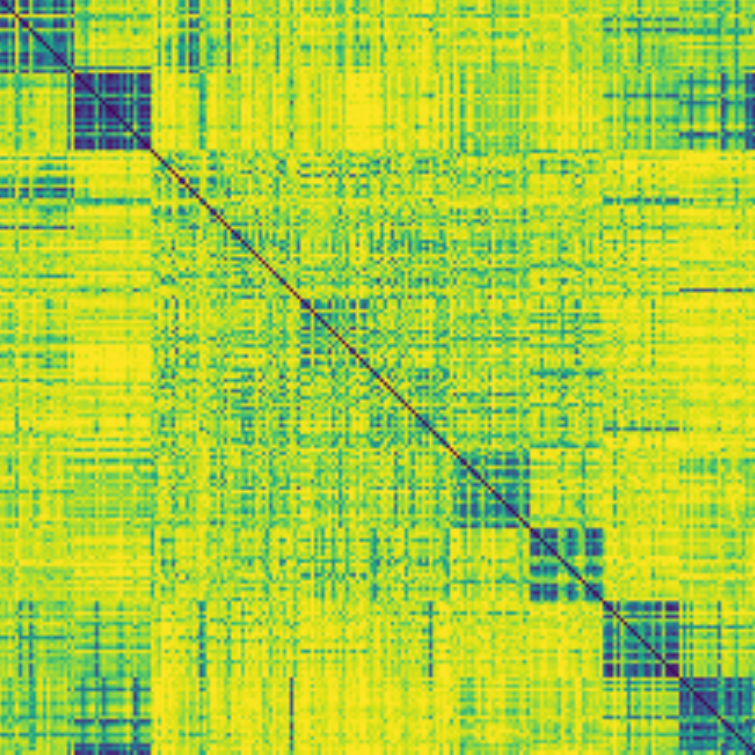}
        \caption[]{$L_2$-Soft.($s$=$1$)}    
        \label{fig:sub6i}
    \end{subfigure}
    \begin{subfigure}[b]{0.15\textwidth}   
        \centering 
        \includegraphics[width=0.85\textwidth]{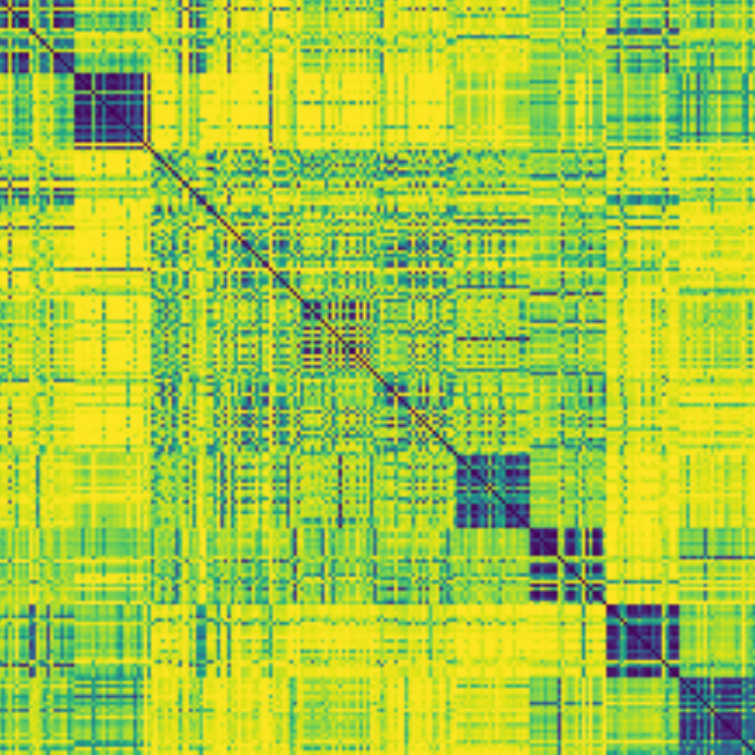}
        \caption[]{$L_2$-Soft.($s$=$0.1$)}    
        \label{fig:sub6j}
    \end{subfigure}
	\begin{subfigure}[b]{0.15\textwidth}   
        \centering 
        \includegraphics[width=0.85\textwidth]{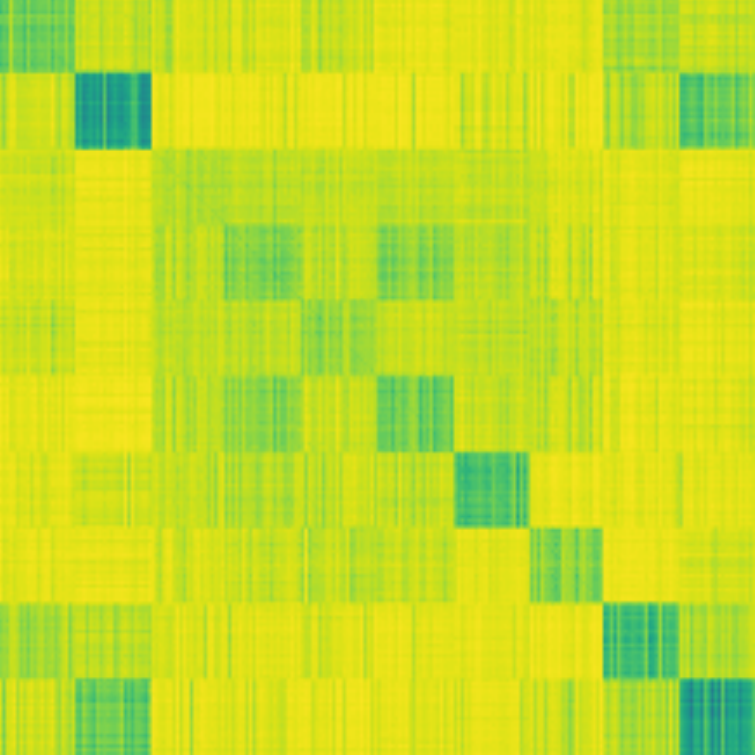}
        \caption[]{$L$-Softmax}
        \label{fig:sub6k}
    \end{subfigure}
    \begin{subfigure}[b]{0.15\textwidth}
        \centering
        \includegraphics[width=0.85\textwidth]{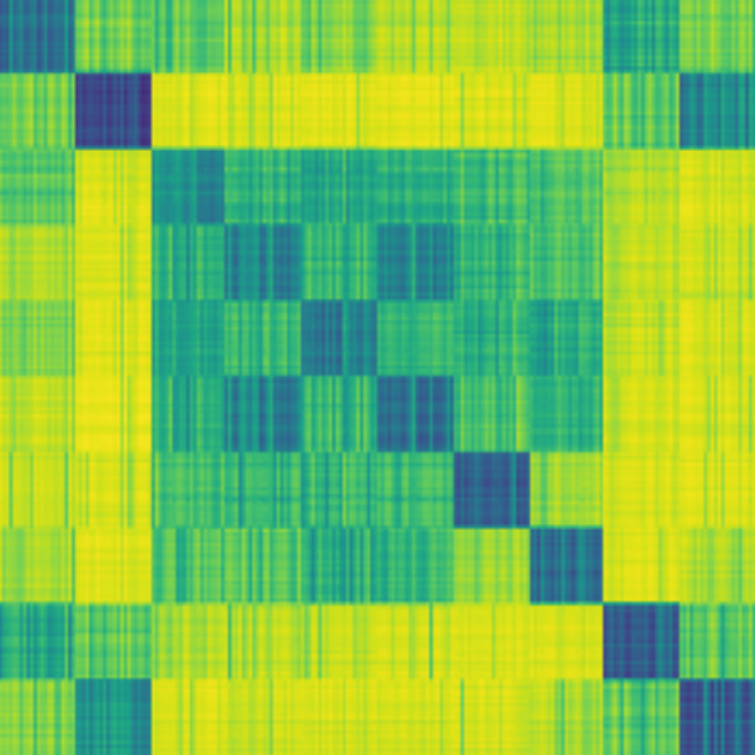}
        \caption[]{SphereFace}
        \label{fig:sub6l}
    \end{subfigure}
    \caption[]
    {\small Cosine distance ($d_c$) matrices between the feature vectors with $ReLU$ on both the training (first row) and testing set (second row) of CIFAR-10. Notice that the lower the scale factor, hence the radius of the feature vectors' hypersphere, the higher the angular similarity between the feature vectors of the same class, \textit{wrt.} the rest. The testing set's classification performance is not considerably affected.} 
    \label{fig:6}
\end{figure*}

We revisit Section~{\ref{sub:SenAnal}} from an experimental point of view in order to clarify the effects of the feature vectors' norm $R$ and orientation $\theta_i$.

\subsubsection{Setup}

During our experimentation, we exploited the simple convolutional architecture presented in Table~{\ref{table:SimpArchit}}, trained it on CIFAR and we studied the behavior of all $\bar{a}_e$.
Each training procedure lasted $40$ epochs with batch size $64$ and learning rate at $10^{-2}$. For each $\bar{a}_e$, we tracked the values of the partial derivatives $\partial S_j / \partial R$ and $ \partial S_j / \partial \theta_i$ based on Eq.~{\ref{eq:SofNorm}} and Eq.~{\ref{eq:SofAng}}, respectively.
Additionally, aiming to discern the effect of $R$, we employed, apart from Softmax, the $L_2$-Softmax~{\cite{ranjan2017l2}}, the \mbox{$L$-Softmax}~{\cite{liu2016large}} and the SphereFace~{\cite{liu2017sphereface}}.
In specific, by keeping the same experimental setup, we trained our CNN with $L_2$-Softmax for different values of $s$.
For stability purposes in the cases of $L$-Softmax and SphereFace, we trained each of them with the larger batch size of $256$, exploiting a decaying learning rate, as proposed in {\cite{liu2016large}} and {\cite{liu2017sphereface}}, respectively.
After training, we calculated the mean centrality and separability in $\mathcal{F}$. 
For visualization purposes, we also extracted the corresponding cosine distance matrices. 
To achieve that, we randomly chose 1000 feature vectors of each class from the training data and calculated their cosine distances $d_c$.
The same procedure was adopted to extract the cosine distance matrices for the testing set, as well.

\subsubsection{Results}

In Fig.~{\ref{fig:Grad}}, the mean and standard deviation of $\partial S_j / \partial R$ and $ \partial S_j / \partial \theta_i$ during the training procedure are displayed.
Note that the corresponding mean values of $ \partial S_j / \partial \theta_i$ are higher at least by one order of magnitude, stressing the orientation's dominance in the classification task.
Moving on to the resulting cosine distance matrices for the Softmax and $L_2$-Softmax of Fig.~{\ref{fig:6}}, one can ascertain the inversely proportional correlation between the norm and the differentiation of the feature vectors belonging to the same class, \textit{wrt.} the rest.
Taking into consideration the analysis in Section~{\ref{sub:SenAnal}} and {\cite{ranjan2017l2}}, the above fact is indeed highly anticipated. Fig. {\ref{fig:5}} showed that for large values of the feature vectors' norm, the Softmax output diagram approximates the rectangular function and broadens the range of the available orientations that succeed a classification accuracy of $100\%$.
As a result, the feature vectors are not motivated to further move away from their decision boundaries and approximate their class's center, thus remaining more distant from the vectors of the same class.

\begin{table}
\centering
\caption{Evaluation of feature space $\mathcal{F}$.}
\label{table:TrPerf}
\resizebox{\linewidth}{!}{%
\renewcommand{\arraystretch}{1.1}
\begin{tabular}{c|c|c|c|c|c|c}
 & Softmax & \multicolumn{3}{c|}{$L_2$-Softmax} & $L$-Softmax & SphereFace \\
 & & ($s$ = $20$) & ($s$ = $1$) & ($s$ = $0.1$) &  & \\
\hline{}
Centrality, $\mathcal{C}_R$& $.859$ & $.725$ & $.751$ & $.752$ & $.926$ &  $.949$ \\
Separability, $\mathcal{S}_R$& $.990$ & $.845$ & $.823$ & $.588$ & $.977$ & $.954$ \\
\hline{}Loss-Ratio, $\mathcal{L}_R$& $2.63$ & $10.85$ & $8.57$ & $19.52$ &$1.42$ & $1.24$ \\
\end{tabular}}
\end{table}

Another finding of Section~{\ref{sub:SenAnal}} constitutes the inability of $R$ to considerably affect the classification performance.
As shown in Table~{\ref{table:TrPerf}} and the testing distance matrices in Fig.~{\ref{fig:6}}, our simple CNN appears to overfit the training set since the separability ratio $\mathcal{S}_R$ is reduced.
We can state that, different values of $s$ cause higher angular separability between the training feature vectors, which is not given for the testing set.
Hence, specific care is required in angular-based losses, because they can significantly strengthen the gap between the distribution of the two sets.

\subsection{Distribution of Hyperplanes in $\mathcal{F}$}

\begin{table*}
\centering
\caption{Angles (in degrees) between the differential vectors of each class in $\mathcal{F}$.}
\label{table:HypDistr1}
\resizebox{\linewidth}{!}{%
\renewcommand{\arraystretch}{1.2}
\begin{tabular}{c|c|c|c|c|c|c|c|c|c|c|c|c|c|c|c|c|c|c|}
 & \multicolumn{3}{c|}{Earphone} & \multicolumn{3}{c|}{Rocket} & \multicolumn{3}{c|}{Guitar} & \multicolumn{3}{c|}{Table} & \multicolumn{3}{c|}{Skateboard}\\
 & $D_1$ & $D_2$ & $D_3$ & $D_1$ & $D_2$ & $D_3$ & $D_1$ & $D_2$ & $D_3$ & $D_1$ & $D_2$ & $D_3$ & $D_1$ & $D_2$ & $D_3$\\
\hline
\textit{Experiment 1} & $55.16$ & $55.49$ & $\textbf{69.36}$ & $51.10$ & $\textbf{66.76}$ & $62.14$ & $50.65$ & $\textbf{86.11}$ & $43.24$ & $52.87$ & $60.43$ & $\textbf{66.69}$ & $60.26$ & $35.21$ & $\textbf{84.53}$\\
\textit{Experiment 2} & $51.47$ & $56.37$ & $\textbf{72.16}$ & $49.56$ & $\textbf{70.36}$ & $60.08$ & $47.58$ & $\textbf{88.26}$ & $44.16$ & $50.85$ & $59.30$ & $\textbf{69.85}$ & $68.20$ & $26.83$ & $\textbf{84.97}$\\
\textit{Experiment 3} & $52.80$ & $59.67$ & $\textbf{67.53}$ & $47.52$ & $\textbf{74.78}$ & $57.70$ & $49.11$ & $\textbf{88.19}$ & $42.70$ & $45.34$ & $66.42$ & $\textbf{68.23}$ & $63.59$ & $29.04$ & $\textbf{87.38}$\\
\hline
$\mu$ & $53.14$ & $57.18$ & $\textbf{69.68}$ & $49.39$ & $\textbf{70.63}$ & $59.97$ & $49.11$ & $\textbf{87.52}$ & $43.37$ & $49.69$ & $62.05$ & $\textbf{68.26}$ & $64.01$ & $30.36$ & $\textbf{85.63}$\\
$\sigma$ & $1.87$ & $2.20$ & $2.33$ & $1.80$ & $4.02$ & $2.22$ & $1.54$ & $1.22$ & $0.74$ & $3.90$ & $3.83$ & $1.58$ & $3.99$ & $4.34$ & $1.53$\\
\end{tabular}}
\resizebox{\linewidth}{!}{%
\renewcommand{\arraystretch}{1.2}
\begin{tabular}{c|c|c|c|c|c|c|c|c|c|c|c|c|c|c|c|c|c|}
\hline
\multicolumn{13}{c|}{Airplane}\\
\hline
 & \multicolumn{3}{c|}{$D_1$} & \multicolumn{3}{c|}{$D_2$} & \multicolumn{3}{c|}{$D_3$} & \multicolumn{3}{c|}{$D_4$}\\
 & $\widehat{(\bar{w}_{12},\bar{w}_{13})}$ & $\widehat{(\bar{w}_{12},\bar{w}_{14})}$ & $\widehat{(\bar{w}_{13},\bar{w}_{14})}$ &  $\widehat{(\bar{w}_{21},\bar{w}_{23})}$ & $\widehat{(\bar{w}_{21},\bar{w}_{24})}$ & $\widehat{(\bar{w}_{23},\bar{w}_{24})}$ & $\widehat{(\bar{w}_{31},\bar{w}_{32})}$ & $\widehat{(\bar{w}_{31},\bar{w}_{34})}$ & $\widehat{(\bar{w}_{32},\bar{w}_{34})}$ & $\widehat{(\bar{w}_{41},\bar{w}_{42})}$ & $\widehat{(\bar{w}_{41},\bar{w}_{43})}$ & $\widehat{(\bar{w}_{42},\bar{w}_{43})}$\\
\hline
 \textit{Experiment 1} & $75.20$ & $61.33$ & $65.31$ & $51.90$ & $57.91$ & $55.96$ & $52.90$ & $56.63$ & $52.07$ & $60.76$ & $58.07$ & $71.97$\\
 $\mu\pm\sigma$ & \multicolumn{3}{c|}{$67.28\pm7.14$} & \multicolumn{3}{c|}{$55.26\pm3.06$} & \multicolumn{3}{c|}{$53.87\pm2.43$} & \multicolumn{3}{c|}{$63.60\pm7.37$}\\
 \hline
 \textit{Experiment 2} & $77.01$ & $66.60$ & $64.03$ & $56.18$ & $59.78$ & $55.94$ & $46.81$ & $55.25$ & $53.53$ & $53.61$ & $60.72$ & $70.53$\\
 $\mu\pm\sigma$ & \multicolumn{3}{c|}{$69.22\pm6.88$} & \multicolumn{3}{c|}{$57.30\pm2.15$} & \multicolumn{3}{c|}{$51.86\pm4.46$} & \multicolumn{3}{c|}{$61.62\pm8.49$}\\
 \hline
 \textit{Experiment 3} & $71.47$ & $64.73$ & $64.27$ & $59.06$ & $60.04$ & $58.97$ & $49.47$ & $54.79$ & $54.22$ & $55.22$ & $60.94$ & $66.81$\\
 $\mu\pm\sigma$ & \multicolumn{3}{c|}{$66.82\pm4.03$} & \multicolumn{3}{c|}{$59.36\pm0.59$} & \multicolumn{3}{c|}{$52.83\pm2.92$} & \multicolumn{3}{c|}{$60.99\pm5.79$}\\
 \hline
\multicolumn{13}{c|}{Car}\\
\hline
  & \multicolumn{3}{c|}{$D_1$} & \multicolumn{3}{c|}{$D_2$} & \multicolumn{3}{c|}{$D_3$} & \multicolumn{3}{c|}{$D_4$}\\
 & $\widehat{(\bar{w}_{12},\bar{w}_{13})}$ & $\widehat{(\bar{w}_{12},\bar{w}_{14})}$ & $\widehat{(\bar{w}_{13},\bar{w}_{14})}$ &  $\widehat{(\bar{w}_{21},\bar{w}_{23})}$ & $\widehat{(\bar{w}_{21},\bar{w}_{24})}$ & $\widehat{(\bar{w}_{23},\bar{w}_{24})}$ & $\widehat{(\bar{w}_{31},\bar{w}_{32})}$ & $\widehat{(\bar{w}_{31},\bar{w}_{34})}$ & $\widehat{(\bar{w}_{32},\bar{w}_{34})}$ & $\widehat{(\bar{w}_{41},\bar{w}_{42})}$ & $\widehat{(\bar{w}_{41},\bar{w}_{43})}$ & $\widehat{(\bar{w}_{42},\bar{w}_{43})}$\\
\hline
 \textit{Experiment 1} & $57.20$ & $63.18$ & $53.82$ & $75.44$ & $68.35$ & $59.49$ & $47.36$ & $59.83$ & $62.38$ & $48.47$ & $66.36$ & $58.13$\\
 $\mu\pm\sigma$ & \multicolumn{3}{c|}{$58.07\pm4.74$} & \multicolumn{3}{c|}{$\textbf{67.76}\pm7.99$} & \multicolumn{3}{c|}{$56.52\pm8.04$} & \multicolumn{3}{c|}{$57.65\pm8.95$}\\
 \hline
 \textit{Experiment 2} & $60.49$ & $60.10$ & $50.63$ & $73.79$ & $74.57$ & $56.55$ & $45.72$ & $65.33$ & $61.88$ & $45.33$ & $64.04$ & $61.57$\\
 $\mu\pm\sigma$ & \multicolumn{3}{c|}{$57.07\pm5.58$} & \multicolumn{3}{c|}{$\textbf{68.30}\pm10.19$} & \multicolumn{3}{c|}{$57.64\pm10.47$} & \multicolumn{3}{c|}{$56.98\pm10.17$}\\
 \hline
 \textit{Experiment 3} & $56.58$ & $61.08$ & $50.64$ & $75.87$ & $67.60$ & $57.01$ & $47.55$ & $58.42$ & $61.04$ & $51.32$ & $70.94$ & $61.95$\\
 $\mu\pm\sigma$ & \multicolumn{3}{c|}{$56.10\pm5.24$} & \multicolumn{3}{c|}{$\textbf{66.83}\pm9.45$} & \multicolumn{3}{c|}{$55.67\pm7.15$} & \multicolumn{3}{c|}{$61.40\pm9.83$}\\
\end{tabular}}
\end{table*}

\begin{table}
\centering
\caption{3D Point Cloud Statistics regarding the diversity of each class within the training dataset.}
\label{table:PtClSt}
\resizebox{\linewidth}{!}{%
\renewcommand{\arraystretch}{1.1}
\begin{tabular}{c|c|c|c|c|c|c|c|c|}
 & \multicolumn{2}{c|}{$A_1$} & \multicolumn{2}{c|}{$A_2$} & \multicolumn{2}{c|}{$A_3$} & \multicolumn{2}{c|}{$A_4$} \\
 & $DIV$ & $P_i$ & $DIV$ & $P_i$ & $DIV$ & $P_i$ & $DIV$ & $P_i$\\
 \hline
Earphone & $1.200$ & $1.00$ & $1.579$ & $1.00$ & $\textbf{4.164}$ & $\textbf{0.24}$ & $-$ & $-$\\
Rocket & $1.061$ & $1.00$ & $\textbf{3.515}$ & $\textbf{0.89}$ & $2.912$ & $1.00$ & $-$ & $-$\\
Guitar & $1.067$ & $1.00$ & $\textbf{1.137}$ & $1.00$ & $1.038$ & $1.00$ & $-$ & $-$\\
Table & $1.043$ & $1.00$ & $1.166$ & $1.00$ & $\textbf{3.068}$ & $\textbf{0.62}$ & $-$ & $-$\\
Skateboard & $1.869$ & $0.64$ & $1.081$ & $1.00$ & $\textbf{3.541}$ & $\textbf{0.80}$ & $-$ & $-$\\
Airplane & $1.025$ & $1.00$ & $1.023$ & $1.00$ & $1.202$ & $1.00$ & $\textbf{2.438}$ & $\textbf{0.78}$ \\
Car & $1.121$ & $1.00$ & $\textbf{1.409}$ & $\textbf{0.84}$ & $1.035$ & $1.00$ & $1.024$ & $1.00$ \\
\end{tabular}}
\end{table}

At this stage, it would be reasonable for the reader to wonder whether there is a possible correlation between the distribution of the input data of a DNN and the produced feature vectors in $\mathcal{F}$.
Keeping in mind that the second depends on the loci of the target classes in $\mathcal{F}$, the above correlation between the input data distribution and the resulting classes' loci is studied.

\subsubsection{Setup}

\begin{figure}
    \centering
    \includegraphics[width=0.8\linewidth]{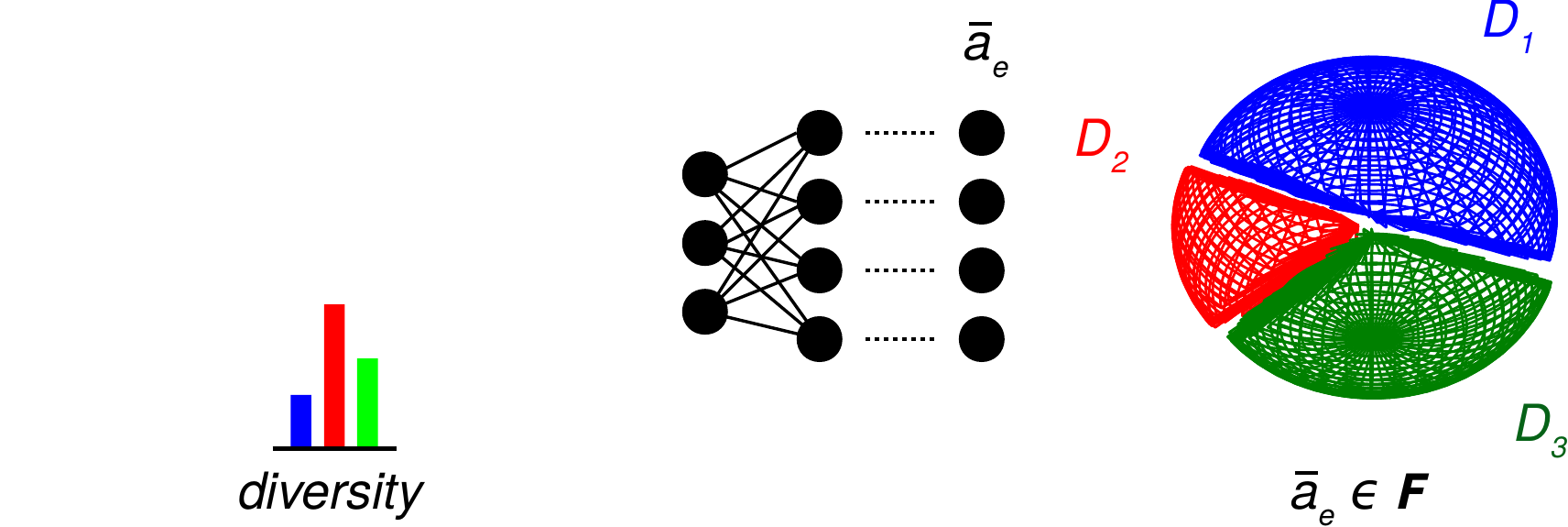}
    \caption{\small Experimental setup for studying the correlation between the input data's distribution and the division in $\mathcal{F}$. A class with larger diversity among the training dataset occupies a smaller subspace.}
        \label{fig:PoClExp}
\end{figure}

To achieve that, we investigate the challenge of instance segmentation of 3D point clouds since it is more straightforward to extract the necessary properties regarding the distribution of the input.
More specifically, we employ the well-established PointNet architecture~{\cite{qi2017pointnet}} and conduct our experiments on the ShapeNet dataset~{\cite{yi2016}}.
Within each experimentation, we exploit only one of the available categories of ShapeNet and train the PointNet with the purpose to classify each point of the cloud to the correct class of the specific category.
As an example, the category \textit{rocket}, also depicted in Fig.~{\ref{fig:PoClExp}}, includes three classes, \textit{viz.}, the body, the fin and the nose.
The input of the architecture constitutes the entire 3D point cloud of one instance of the category \textit{rocket}, while the network's output is the classification score of each point.
Since the output layer uses shared weights for each point~{\cite{qi2017pointnet}}, we secure that the same separation hyperplanes are simultaneously used to classify each point of the instance.
In other words, for $n\in\mathbb{N}^{*}$ number of points, we end up with $n$ feature vectors in $\mathcal{F}$ classified by the hyperplanes defined by the weights of the last layer.
PointNet is trained using the setup proposed in~{\cite{qi2017pointnet}}.

\subsubsection{Results}
After training, we measure the angles between the resulting differential vectors for each class as presented in Table~{\ref{table:HypDistr1}}.
In the case of three classes, there is exactly one angle to be calculated for each class since, based on Eqs.~{\ref{eq:22}} and {\ref{eq:23}}, the class is defined by two differential vectors.
In the case of four classes, we end up with three angles.
Note that each experiment is repeated three times, thus taking into consideration its stochastic nature and providing better apprehension.
We also present the mean $\mu$ and standard deviation $\sigma$ among the same experiments.
For the cases of \textit{airplane} and \textit{car}, we added the mean and the standard deviation among the angles of the same target class, as well.
In addition, Table~{\ref{table:PtClSt}} keeps a statistical measurement of the dataset's 3D shapes, regarding the diversity of each class among the dataset.
The class's diversity is computed through the divergence of its point cloud among different instances of the dataset and it can be interpreted as a measurement regarding the complexity that the network needs to learn.
In specific, let $\mathcal{P}^c_i$ and $\mathcal{P}^c_j$, $i,j=1,2,...,N_I$, $i\neq j$, be the point clouds of the $c$-th class from two of the total $N_I\in\mathbb{N}^{*}$ dataset's instances.
We compute the mean Euclidean distance between all points of $\mathcal{P}^c_i$ and $\mathcal{P}^c_j$, denoted as $\bar{d}_{\mathcal{E}}(\mathcal{P}^c_i,\mathcal{P}^c_j)$.
In addition, we compute the corresponding distance between the points of $\mathcal{P}^c_i$ itself ($\bar{d}_{\mathcal{E}}(\mathcal{P}^c_i,\mathcal{P}^c_i)$).
The diversity of the $c$-th class among the dataset arises from the equation below:
\begin{equation}
DIV^c = \frac{1}{N_I}\frac{\sum_{i=1}^{N_I}{\sum_{j=1,j\neq i}^{N_I}{\bar{d}_{\mathcal{E}}(\mathcal{P}^c_i,\mathcal{P}^c_j)}}}{\sum_{i=1}^{N_I}{\bar{d}_{\mathcal{E}}(\mathcal{P}^c_i,\mathcal{P}^c_i)}}.
\end{equation}
Yet, since it is quite common for a class not to be included in a point cloud instance, we also keep the percentage of instances ($P_i$) that include points from each class.
We argue that this metric also quantifies the diversifying nature of a class.
Due to the heavy computational cost required for the calculation of the above, our results in Table~{\ref{table:PtClSt}} have emerged from a randomly selected subset  containing $25\%$ of the total instances.

\begin{table*}
\centering
\caption{Centrality and Separability ratio between the testing and training set of $D_a$ for both visual and audio feature extractors. This experiment was conducted over multiple iterations, each time leaving out a speaker $s_i$ or speakers-group $gr_j$. }
\label{table:Unimodal}
\resizebox{\linewidth}{!}{%
\renewcommand{\arraystretch}{1.2}
\begin{tabular}{c|c|c|c|c|c|c|c|c|c|c|c|c|c|c|}
\textbf{Angular} & \multicolumn{8}{c|}{RML} & \multicolumn{6}{c|}{BAUM-1s}  \\
\textbf{Metrics} & $s_1$ & $s_2$ & $s_3$ & $s_4$ & $s_5$ & $s_6$ & $s_7$ & $s_8$ & $gr_1$ & $gr_2$ & $gr_3$ & $gr_4$ & $gr_5$ & $gr_6$\\
\hline{} & \multicolumn{8}{c|}{\textit{Visual Feature Extractor}} & \multicolumn{6}{c|}{\textit{Visual Feature Extractor}}  \\
\hline{}Centrality, $\mathcal{C}_R$& $.697$ & $.760$ & $.801$ & $.635$ & $.787$ & $.713$ & $.610$ & $.657$ & $.688$ & $.607$ & $.687$ & $.468$ & $.593$ & $.556$ \\
Separability, $\mathcal{S}_R$& $.687$ & $.628$ & $.632$ & $.826$ & $.653$ & $.768$ & $.768$ & $.778$ & $.489$ & $.553$ & $.519$ & $.536$ & $.420$ & $.488$ \\
\hline{}Loss-Ratio, $\mathcal{L}_R$& $22.085$ &$21.159$ &$8.431$ &$8.397$ &$11.516$ &$10.048$ &$18.999$ &$13.547$ &$12.859$ &$19.052$ &$15.056$ &$22.634$ &$20.673$ &$23.746$ \\
\hline{} & \multicolumn{8}{c|}{\textit{Audio Feature Extractor}} & \multicolumn{6}{c|}{\textit{Audio Feature Extractor}}  \\
\hline{}Centrality, $\mathcal{C}_R$& $.646$ & $.739$ & $.816$ & $.705$ & $.767$ & $.802$ & $.870$ & $.646$ & $.744$ & $.720$ & $.605$ & $.656$ & $.547$ & $.857$ \\
Separability, $\mathcal{S}_R$& $.653$ & $.692$ & $.857$ & $.833$ & $.975$ & $.976$ & $.930$ & $.884$ & $.742$ & $.730$ & $.882$ & $.814$ & $.670$ & $.684$ \\
\hline{}Loss-Ratio, $\mathcal{L}_R$& $13.168$ & $10.227$ & $5.003$ & $8.810$ & $7.460$ & $6.283$ & $4.942$ & $6.247$ & $11.294$ & $13.050$ & $13.358$ & $11.570$ & $17.016$ & $10.681$ \\
\end{tabular}}
\end{table*}

From Table~{\ref{table:HypDistr1}}, we ascertain that for each category there is consistency regarding the succession of the angles' values among the classes.
The above is highly important since it proves that the distribution of the classifiers in $\mathcal{F}$ is not fully stochastic, but rather follows several rules based on the data distribution.
By further comparing Tables~{\ref{table:HypDistr1}} and~{\ref{table:PtClSt}} for each category and each class, we notice a strong relation between the diversity of the class and its angles in $\mathcal{F}$.
Focusing on the most diversifying classes of each category we can exclude that their differentional vectors present the highest angles.
Keeping in mind Eq.~{\ref{eq:25}}, the larger the angles between the differential vectors, the smaller the size of a class's locus.
Hence, classes with high diversity occupy less space in $\mathcal{F}$.
At a first glance, this statement may appear a bit misplaced, but one should keep in mind that samples with higher divergence provoke higher loss values and, in turn, cause more intensive changes to the DNN's parameters.
Those changes can be visible in $\mathcal{F}$ as larger corrections of feature vector's norm and angle.
Eventually, more diverging samples are prone to movement and tend to concentrate more in the studied space.

\subsection{Multi-modal Fusion and Strategies}

\begin{figure}
    \centering
	\begin{subfigure}[b]{0.115\textwidth}   
        \centering 
        \includegraphics[width=0.8\textwidth]{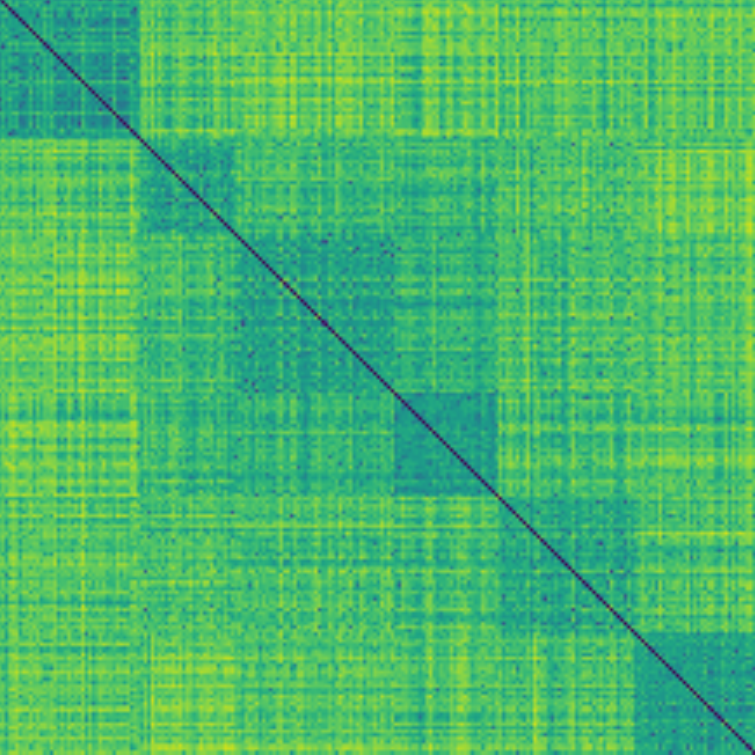}
        \caption[]{$76.25\%$}    
    	\label{fig:sub7a}
    \end{subfigure}    
    \begin{subfigure}[b]{0.115\textwidth}
        \centering
        \includegraphics[width=0.8\textwidth]{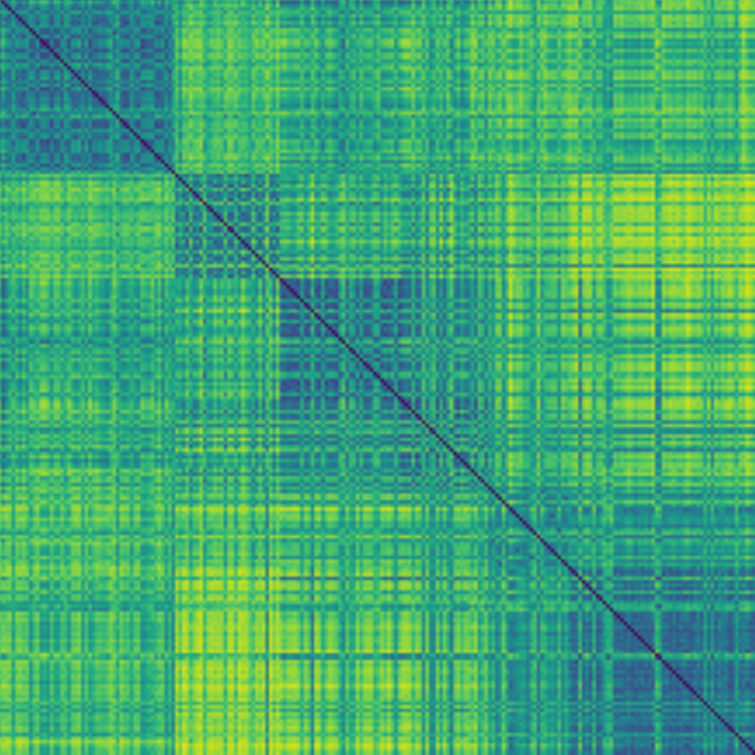}
        \caption[]{$68.87\%$}    
        \label{fig:sub7b}
    \end{subfigure}
    \centering
    \begin{subfigure}[b]{0.115\textwidth}
        \centering 
        \includegraphics[width=0.8\textwidth]{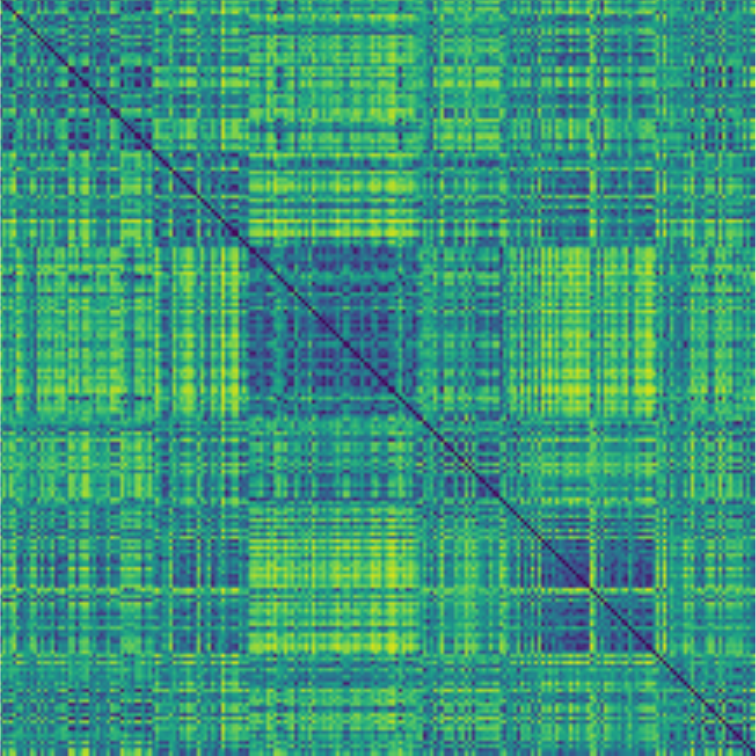}
        \caption[]{$69.60\%$}    
        \label{fig:sub7c}
    \end{subfigure}
    \begin{subfigure}[b]{0.115\textwidth}   
        \centering 
        \includegraphics[width=0.8\textwidth]{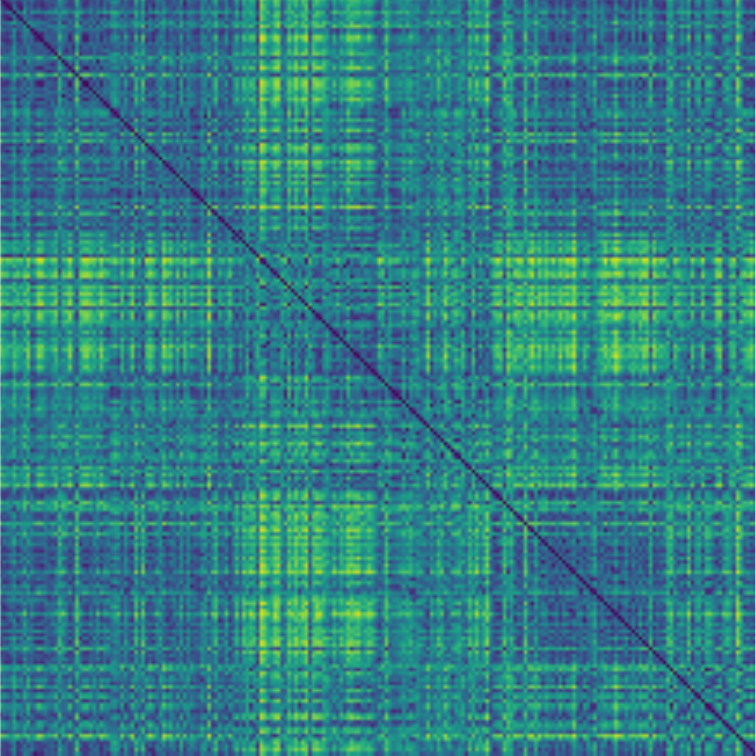}
        \caption[]{$44.66\%$}    
        \label{fig:sub7d}
    \end{subfigure}
    \caption[]
    {\small Cosine distance ($d_c$) matrices between the feature vectors and the corresponding performance on $D_a$ for: the (a) training and (b) testing set of the visual feature extractor, as well as  the (c) training and (d) testing set of the audio feature extractor.}  
    \label{fig:7}
\end{figure}

The current section deals with the shape of overfitting in $\mathcal{F}$ and its adverse effect on feature extraction under cascade training tasks, like modalities fusion.
Moreover, an alternative fusion strategy that can considerably benefit the overall performance is proposed.
The above are examined in the A-V emotion recognition challenge, by exploiting two deep unimodal feature extractors and a fusion model.

\subsubsection{Setup of Feature Extractors}

First, we investigate the relationship between overfitting and angular distribution both for audio and visual modalities on RML and BAUM-1s.
During training, we formed a uniform dataset $D_a$, by combining the three emotional databases and we evaluated the resulting distribution on each speaker ($s_i$, $i=1,2,...,8$) of RML and each group of speakers ($gr_j$, $j=1,2,...,6$) of BAUM-1s.
This course of experiments was performed under a leave-one-out evaluation scheme, each time considering a different speaker or group as a testing set.
We exploited the MobileNetV2~\cite{sandler2018mobilenetv2} pre-trained on ImageNet~\cite{deng2009imagenet} to fine-tune our face images.
The training audio samples were also used to fine-tune the pre-trained VGGish~\cite{hershey2017cnn} architecture.
We utilized the Adam optimizer~\cite{kingma2014adam} with batch size $32$ and an initial learning rate at $10^{-5}$ that exponentially decays by $90\%$ for every 5,000 training steps.
We trained each architecture for $10$ epochs and investigated the angular distribution of the feature vectors.

\begin{figure}
    \centering
    \begin{subfigure}{0.5\textwidth}
        \centering
        \includegraphics[width=.9\linewidth]{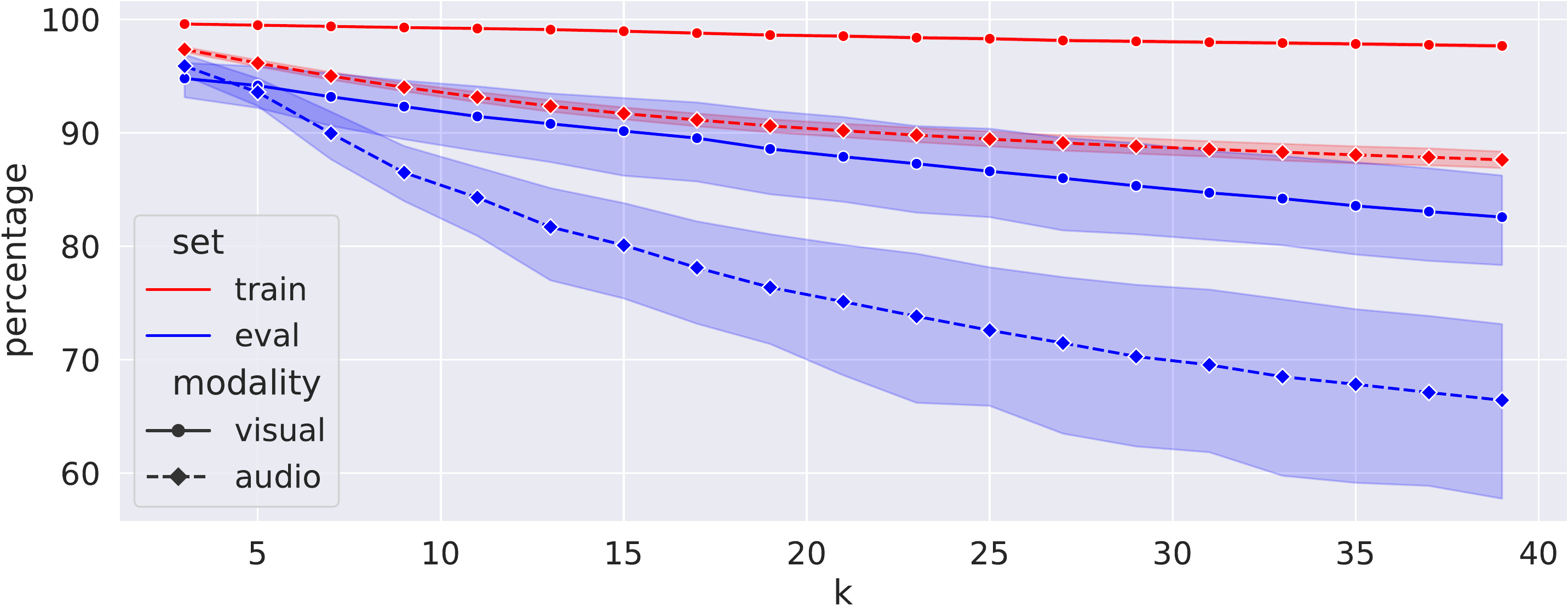}
        \caption{RML}
        \label{fig:subknna}
    \end{subfigure}%
	\vskip\baselineskip
    \centering
    \begin{subfigure}{0.5\textwidth}
        \centering
        \includegraphics[width=.9\linewidth]{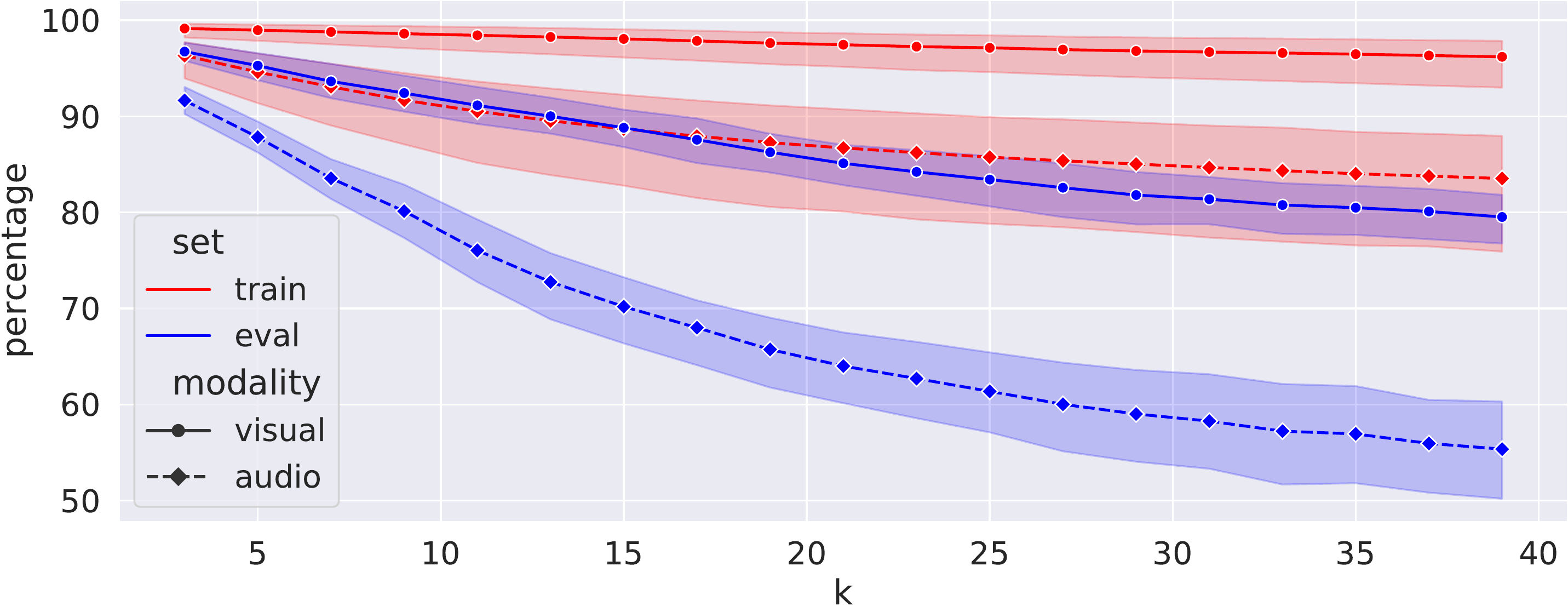}
        \caption{BAUM-1s}
        \label{fig:subknnb}
    \end{subfigure}
    \caption{\small Percentage of the accurately classified feature vectors extracted by the unimodal feature extractors on (a) RML~{\cite{wang2008recognizing}} and (b) BAUM-1s~{\cite{zhalehpour2016baum}}, using $k$-NN with $d_c$ and $k\in [3,39]$. The larger the value of $k$, the bigger the gap between the succeeded percentage on the training (red) and the corresponding testing (blue) vectors.}
    \label{fig:knn}
\end{figure}

\subsubsection{Feature Extractors' Evaluation} 
\label{subsub:FetExt}

Table~\ref{table:Unimodal} presents the resulting values of the $\mathcal{C}_R$ and $\mathcal{S}_R$ for both unimodal extractors.
We can observe the lower values on BAUM-1s, which constitutes a spontaneous, hence more difficult dataset.
Generally, Table~\ref{table:Unimodal} contains low values regarding both centrality and separability measures, indicating a significant gap between the distribution of training and testing sets.
The above property can be further interpreted through a qualitative representation illustrated in Fig.~\ref{fig:7}.
According to Section~\ref{sub:NorOr}, we applied Eq.~\ref{eq:CosSim} as our distance metric for the feature vectors.
Comparing speakers $s_3$, $s_4$, $s_5$, $s_6$ and $s_7$ against the rest of the speakers or speaker-groups, we identify the most favorable results, which also correspond to the lower values of $\mathcal{L}_R$.

In addition, we employ $k$-NN, aiming to further demonstrate the divergence of the feature vectors' angular distribution between the training and the testing set.
Hence, we apply $k$-NN to the feature vectors for each one of the trained unimodal feature extractors in Table~{\ref{table:Unimodal}}, using the cosine distance metric.
In specific, given a speaker and one of the two sets, we measure $\forall \bar{a}_e$ of the set the percentage of the positive neighbors, \textit{i.e.}, the number of those that correspond to the same class, over the number of the nearest neighbors $k$.
In case that this percentage is higher than $50\%$, the current feature vector is considered to be correctly classified.
We repeat the experiment for any odd number of nearest neighbors $k$ within $[3, 39]$, keeping the percentage of the accurately classified feature vectors.
In Fig.~{\ref{fig:knn}}, the mean and the standard deviation among the speakers, both in RML and BAUM-1s, is illustrated for each value of $k\in [3,39]$.
For each unimodal extractor on both datasets there is a gap between the percentage succeeded by $k$-NN on the training and the corresponding testing feature vectors, which becomes even wider as we increase the value of $k$.
The above indicates the divergence between the angular distribution of the two sets, settling an empirical illustration of the separability problem in cascade tasks.

As shown, the distribution of feature vectors in $\mathcal{F}$ is affected during training, leading to better central distribution and angular separability of the training vectors as compared against the testing ones.
Hence, in cases of further training, like fusion, where such a distribution constitutes the input space of the cascade architecture, the existing discriminable training features hamper the correct classification of the more ambiguous features of the testing set.
With the above in mind, an unintuitive but crucial question arises.
\textit{Would it be beneficial to divide the available training data into two sets, one for training the unimodal extractors and one for their fusion?}
By doing so, the training of the fusion model would also include more ambiguous samples since they would be novel to the unimodal network.
Such an input sample would closely resemble the distribution of the actual testing set and the rest of this subsection is dedicated to the assessment of this approach.

\subsubsection{Setup of Fusion Strategies}

We made use of the above architectures for our unimodal feature descriptors, \textit{viz.}, the MobileNetV2 and the VGGish for the visual and the audio channels, respectively.
For fusion, we tested the performance of both a Support Vector Machine (SVM) and a DNN architecture.
The latter contains two Fully-Connected (FC) layers of $512$ neurons with dropout rate $0.3$ and three batch normalization layers before, in-between and after the FC layers.
$D_a$ was shuffled and split in half, forming the sets: $D_1$ and $D_2$.
Then, we evaluated the performance of the total architecture considering the strategies below:
\begin{itemize}

\item $S_{1-1}$: We exploited the face images and speech samples of $D_1$ to fine-tune the visual and audio channels.
Then, the fusion model was trained from the A-V data of $D_1$.

\item $S_{1-2}$: The face images and speech samples of $D_1$ were similarly exploited to fine-tune the visual and audio channels, respectively.
Then, the fusion model was trained with the A-V data of the unused set $D_2$.

\item $S_{a-a}$: We fine-tuned both channels with the unimodal data of the total set $D_a$. Consequently, we trained their fusion with the same A-V data.
Note that this training scheme also describes the experimental protocol for producing the results in Table \ref{table:Unimodal}.

\end{itemize}

Each training procedure has been conducted for 10 epochs, exploiting the Adam optimizer with mini batch 32 and learning rate starting at $10^{-5}$ and exponentially decaying by $90\%$ every 5,000 steps.
Similarly, we assessed each strategy on RML and BAUM-1s, following the LOSO and LOSGO schemes, respectively.
We also measured the mean performance among all the aforementioned speakers and speakers-groups.

\subsubsection{Results of Fusion Models}

\begin{figure*}
    \centering
    \begin{subfigure}[b]{0.325\textwidth}
        \centering
        \includegraphics[width=0.9\textwidth]{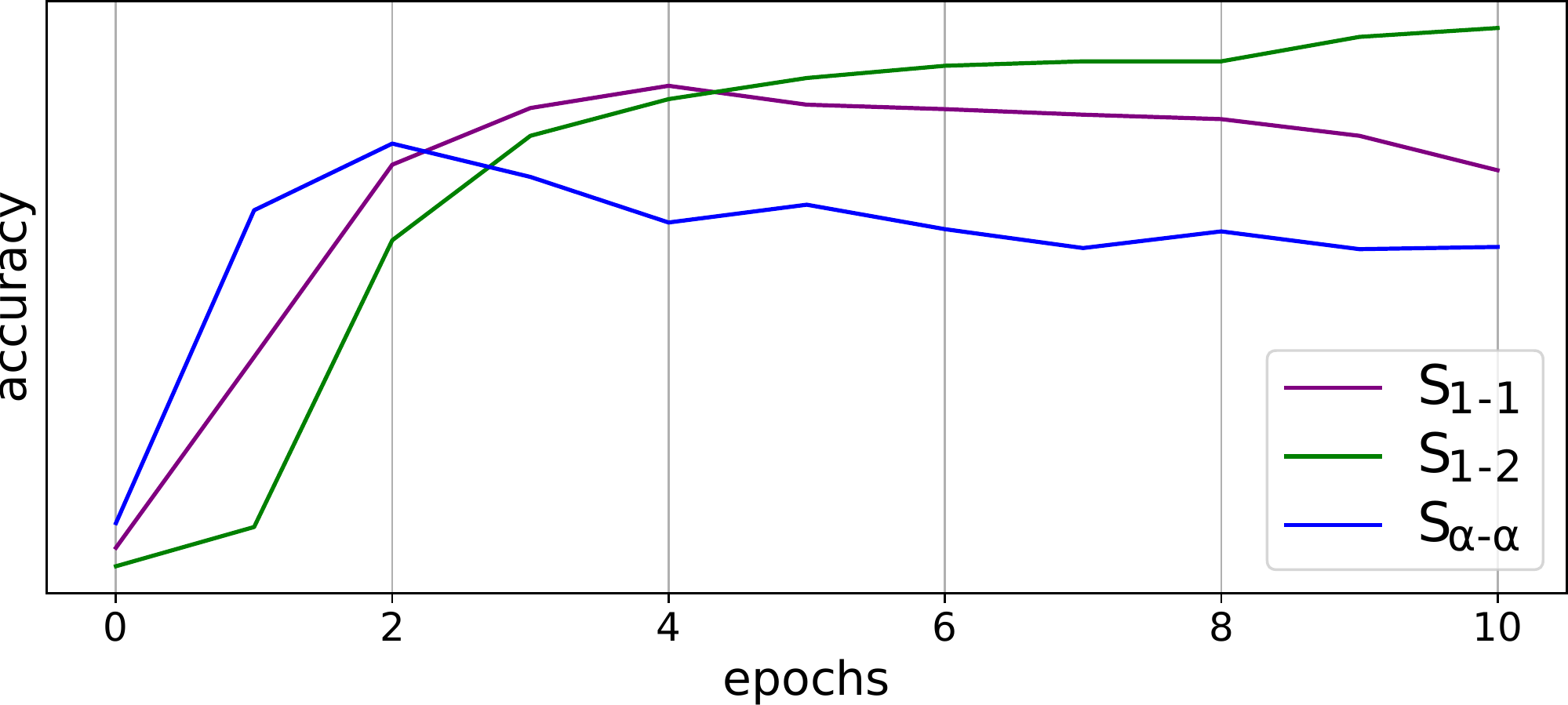}
        \caption[]{$s_4$}    
        \label{fig:sub8a}
    \end{subfigure}
    \begin{subfigure}[b]{0.325\textwidth}
        \centering 
        \includegraphics[width=0.9\textwidth]{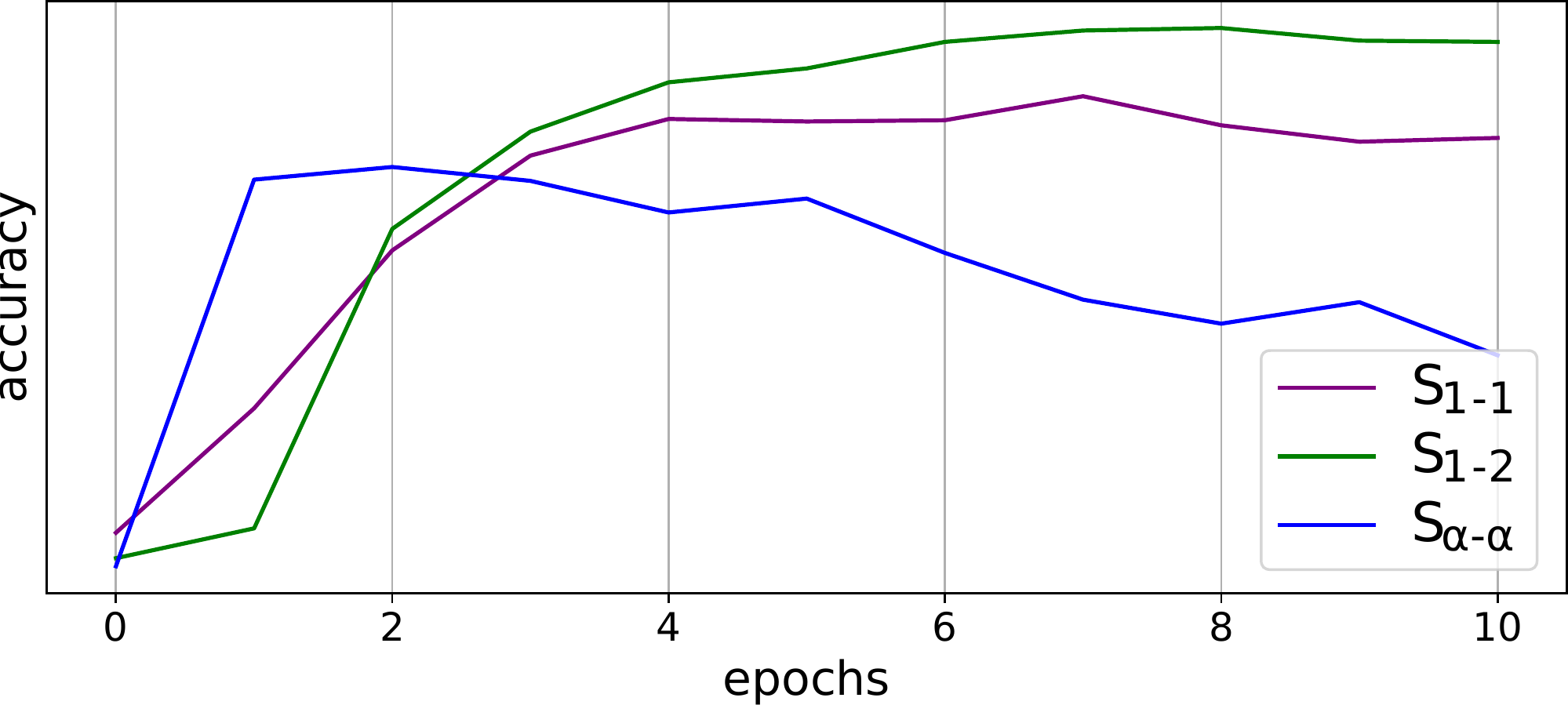}
        \caption[]{$gr_1$}    
        \label{fig:sub8b}
    \end{subfigure}
    \begin{subfigure}[b]{0.325\textwidth}
        \centering 
        \includegraphics[width=0.9\textwidth]{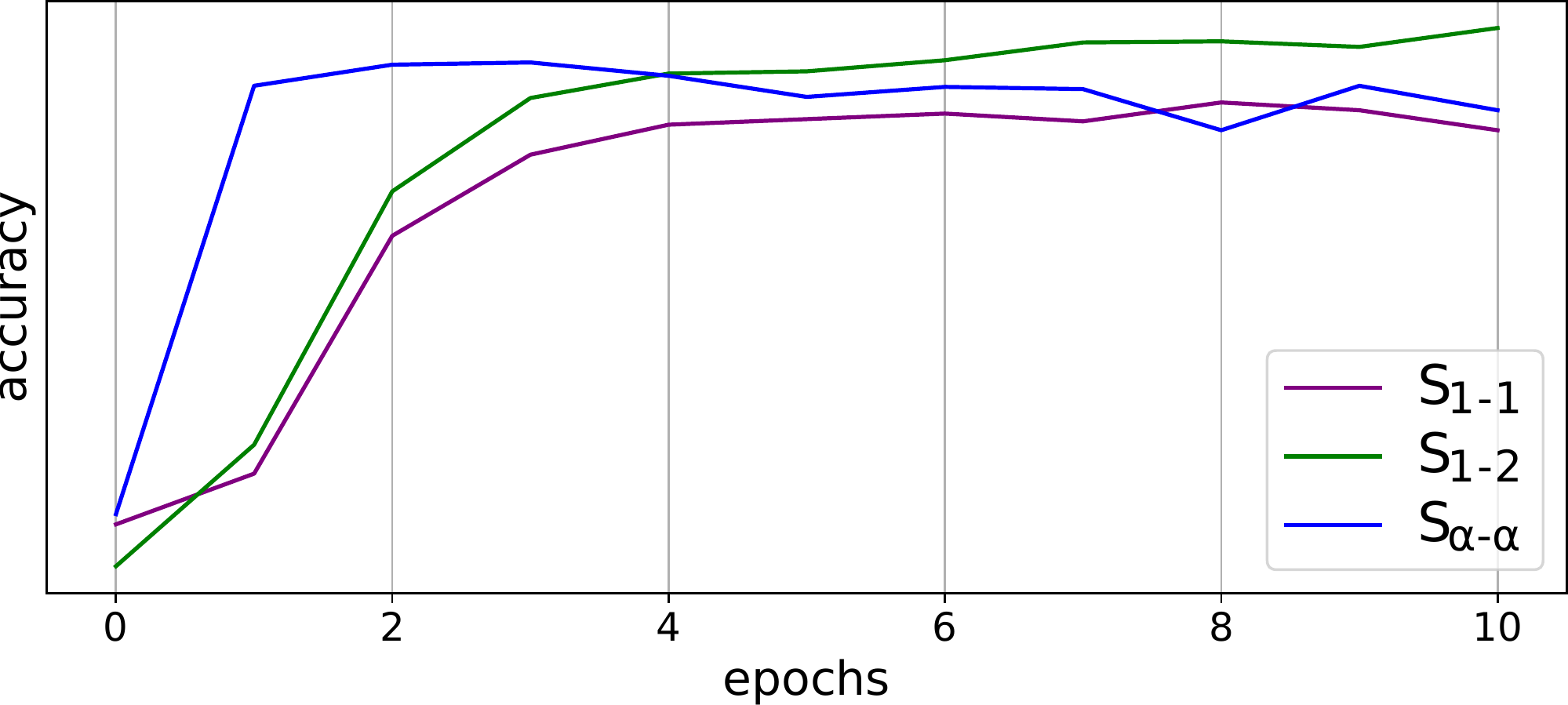}
        \caption[]{$s_1$}    
        \label{fig:sub8c}
    \end{subfigure}
    \vskip\baselineskip
    \centering
    \begin{subfigure}[b]{0.325\textwidth}   
        \centering 
        \includegraphics[width=0.9\textwidth]{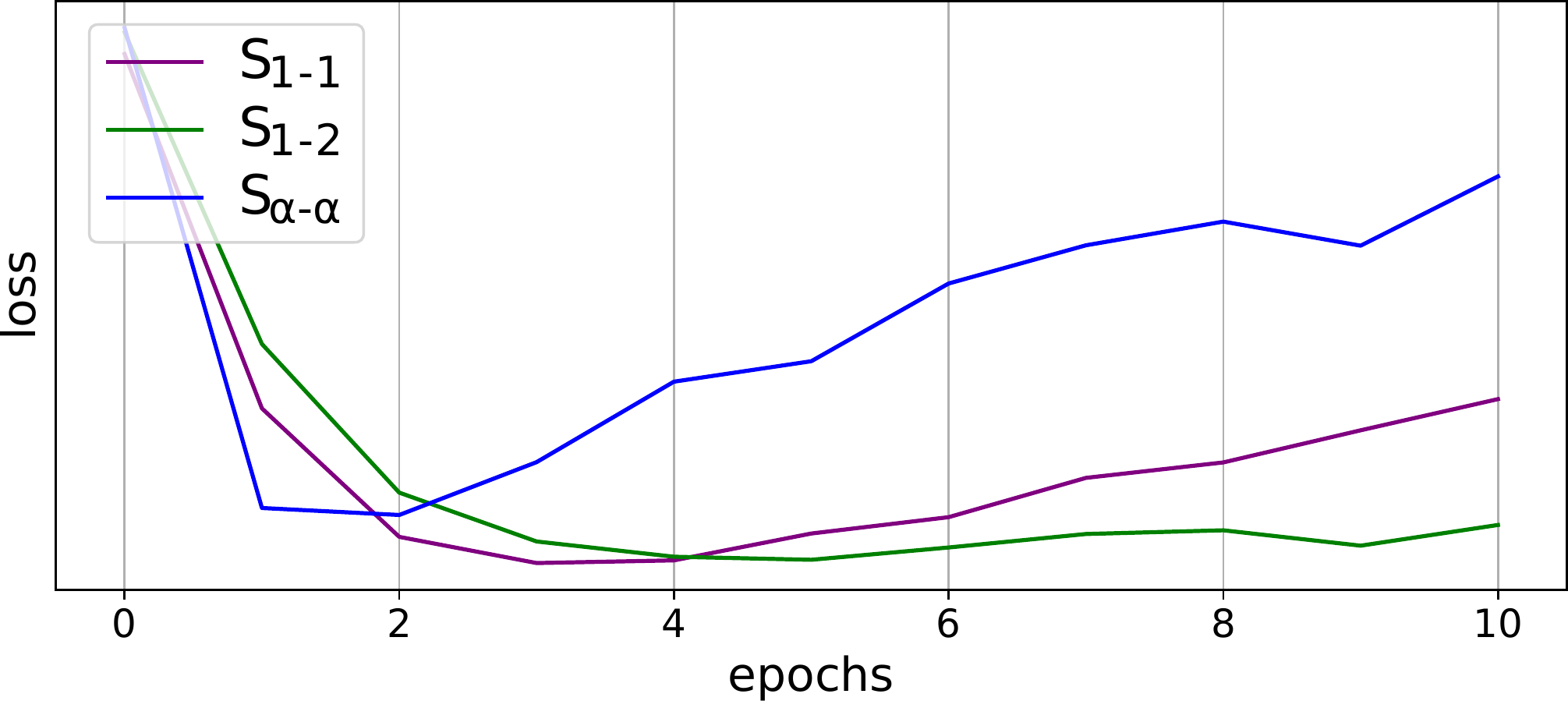}
        \caption[]{$s_4$}    
        \label{fig:sub8d}
    \end{subfigure}
    \begin{subfigure}[b]{0.325\textwidth}   
        \centering 
        \includegraphics[width=0.9\textwidth]{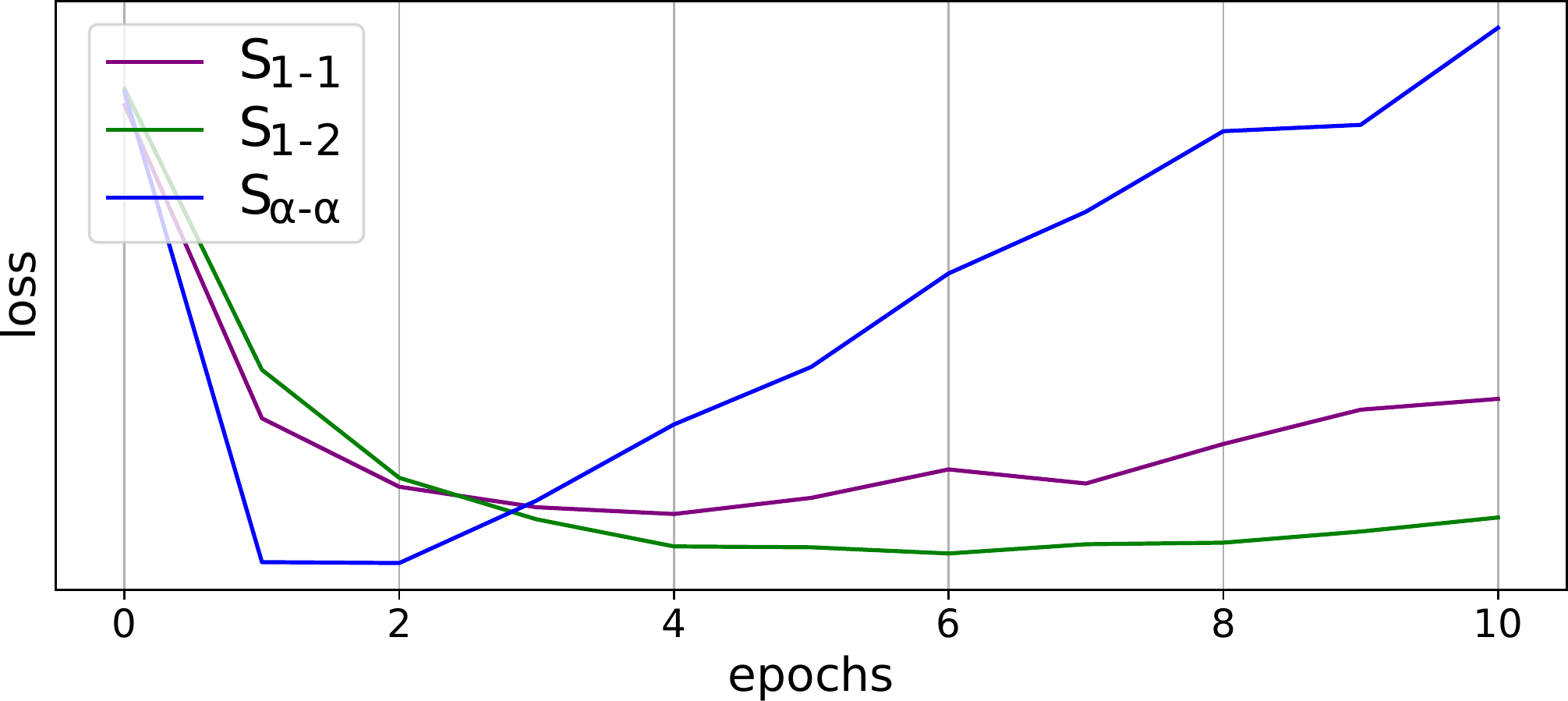}
        \caption[]{$gr_1$}    
    	\label{fig:sub8e}
    \end{subfigure}
    \begin{subfigure}[b]{0.325\textwidth}
        \centering 
        \includegraphics[width=0.9\textwidth]{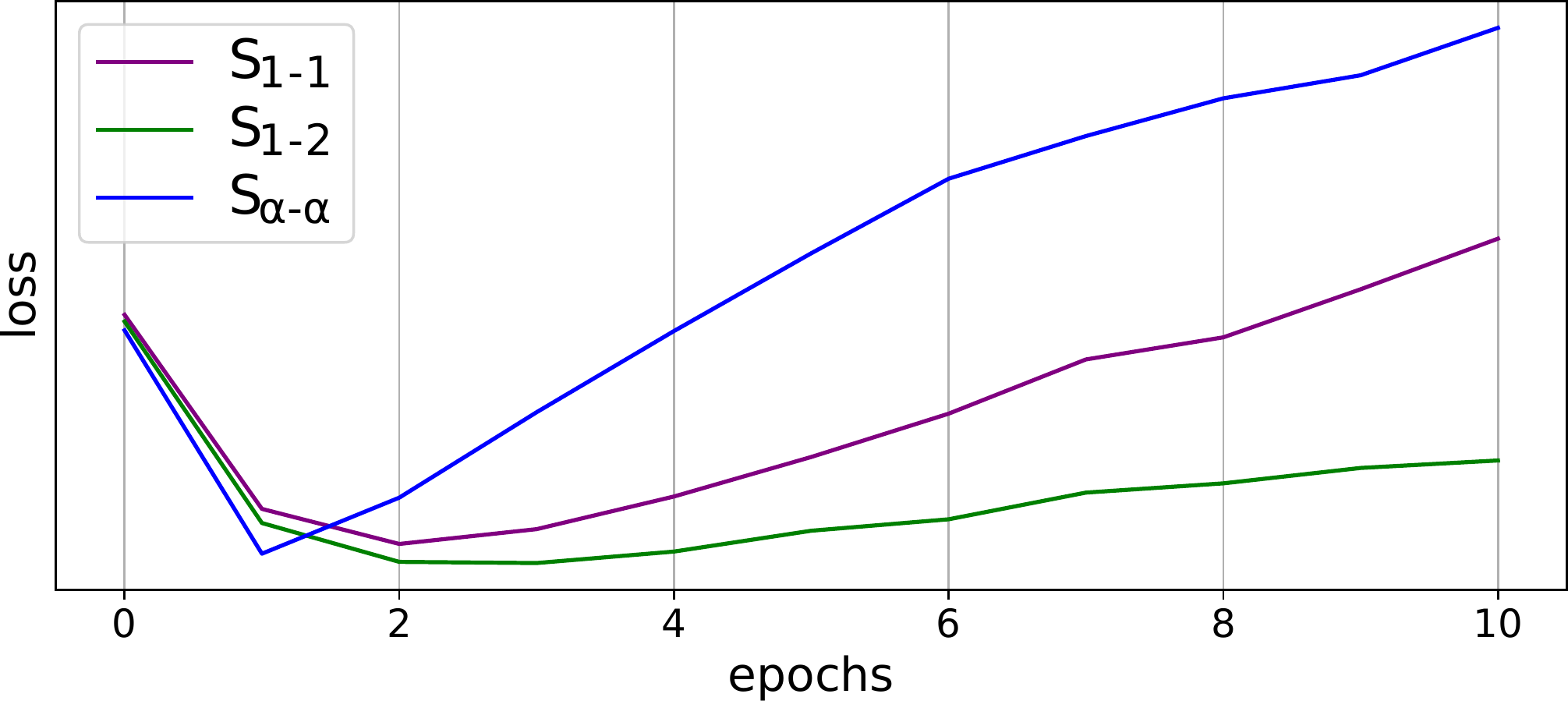}
        \caption[]{$s_1$}    
        \label{fig:sub8f}
    \end{subfigure}
    \caption[]
    {\small Accuracy (first row) and loss (second row) curves of three representative evaluation examples during training the DNN fusion model under different strategies. When the same training data are used (see $S_{1-1}$, $S_{a-a}$), overfitting appears in the early stages of training.} 
    \label{fig:8}
\end{figure*}

\begin{table*}
\centering
\caption{Classification performance (\%) of fusion models under different training strategies.}
\label{table:TrFus1}
\resizebox{.98\linewidth}{!}{%
\renewcommand{\arraystretch}{1.2}
\begin{tabular}{c|c|c|c|c|c|c|c|c|c|c|c|c|c|c|c|c|}
\multirow{2}{*}{\textbf{Strategy}} & \multicolumn{8}{c|}{RML} & \multicolumn{6}{c|}{BAUM-1s} & \multirow{2}{*}{\textbf{mean}}  \\
 & $s_1$ & $s_2$ & $s_3$ & $s_4$ & $s_5$ & $s_6$ & $s_7$ & $s_8$ & $gr_1$ & $gr_2$ & $gr_3$ & $gr_4$ & $gr_5$ & $gr_6$ &  \\
\hline{}&\multicolumn{14}{c|}{\textit{DNN Fusion Model}}& \\
\hline{}$S_{1-1}$ & $51.15$ & $60.59$ & $69.88$ & $68.61$ & $61.43$ & $69.87$ & $49.15$ & $66.86$ & $55.82$ & $34.43$ & $57.94$ & $46.27$ & $49.10$ & $32.34$ & $58.53$ \\
$S_{1-2}$ & $\textbf{54.36}$ & $\textbf{64.12}$ & $70.07$ & $\textbf{69.58}$ & $\textbf{63.05}$ & $\textbf{73.82}$ & $54.21$ & $\textbf{68.78}$ & $\textbf{56.26}$ & $\textbf{36.02}$ & $\textbf{58.38}$ & $\textbf{46.91}$ & $\textbf{53.77}$ & $\textbf{41.58}$ & $\textbf{61.46}$ \\
$S_{a-a}$ & $45.41$ & $54.41$ & $\textbf{73.05}$ & $67.72$ & $60.78$ & $71.96$ & $\textbf{54.62}$ & $52.50$ & $41.81$ & $34.33$ & $53.31$ & $44.59$ & $52.70$ & $29.16$ & $55.64$ \\
\hline{}&\multicolumn{14}{c|}{\textit{SVM Fusion Model}}& \\
\hline{}$S_{1-1}$ & $53.44$ & $65.59$ & $71.36$ & $75.72$ & $64.26$ & $75.83$ & $55.81$ & $65.53$ & $40.12$ & $36.22$ & $59.30$ & $48.12$ & $50.47$ & $36.65$ & $60.25$ \\
$S_{1-2}$ & $\textbf{55.96}$ & $\textbf{67.65}$ & $72.22$ & $\textbf{79.52}$ & $\textbf{67.42}$ & $\textbf{80.17}$ & $62.05$ & $\textbf{73.79}$ & $\textbf{52.07}$ & $\textbf{36.72}$ & $\textbf{59.71}$ & $\textbf{51.12}$ & $\textbf{50.68}$ & $\textbf{39.32}$ & $\textbf{64.05}$ \\
$S_{a-a}$ & $48.39$ & $62.35$ & $\textbf{75.93}$ & $79.07$ & $66.86$ & $79.40$ & $\textbf{63.20}$ & $62.52$ & $35.61$ & $36.52$ & $55.66$ & $46.58$ & $47.59$ & $32.55$ & $59.62$ \\
\hline{}&\multicolumn{14}{c|}{\textit{Pre-Trained DNN Fusion Model}}& \\
\hline{}$S'_{1-2}$ & $\textbf{51.83}$ & $\textbf{75.29}$ & $\textbf{82.65}$ & $84.24$ & $\textbf{80.23}$ & $\textbf{75.76}$ & $\textbf{76.72}$ & $\textbf{75.29}$ & $\textbf{61.17}$ & $\textbf{40.70}$ & $\textbf{50.15}$ & $\textbf{47.96}$ & $\textbf{69.09}$ & $42.71$ & $\textbf{68.98}$ \\
$S'_{a-a}$ & $43.12$ & $72.06$ & $80.93$ & $\textbf{86.63}$ & $74.39$ & $75.06$ & $73.10$ & $72.79$ & $56.17$ & $38.71$ & $44.19$ & $40.62$ & $63.55$ & $\textbf{43.43}$ & $65.18$ 
\end{tabular}}
\end{table*}

The results in Table~\ref{table:TrFus1} verify the adverse effect of the overfitted unimodal descriptors on the performance of the fusion model.
Firstly, comparing strategy $S_{1-1}$ against $S_{1-2}$, we can ensure that the second one outperforms the first at any case for both fusion models.
This fact is highly anticipated since the final multi-modal architecture of $S_{1-2}$, in contrast to $S_{1-1}$, has been trained on the whole dataset, including both $D_1$ and $D_2$.
Therefore, we can not conclude that training the fusion model on unfamiliar data, has benefited the final performance.
For this purpose, lets focus on $S_{1-2}$ and $S_{a-a}$ that have been trained on the same amount of data.
We observe that except for two cases, including $s_3$ and $s_7$, the results related to any other speaker have been improved through the proposed cascade strategy.
Paying attention to the mean performance of each strategy on the whole pool of speakers and speakers-groups, we can highlight this advantageous property even more.
The mean performance appears to have been improved by $5.82\%$ on the DNN and $4.43\%$ on the SVM fusion model. 

Noticing the better performance of SVM, one could state either that DNN seems not to be the most appropriate fusion model for this multi-modal case, or that there could be a better set of hyperparameters, such as learning rate, batch size, architecture, \textit{etc.} to achieve better performance and annul the above benefit.
However, as shown in Fig.~\ref{fig:8} and particularly in the loss curves, when the feature extractors and the fusion model are both trained with the same data, overfitting appears quite earlier, which limits the room for improvement of the fusion model.
The accuracy curves of $S_{a-a}$ display a quick peak during the first epochs, according to the similarity of the feature vectors between the evaluation speaker and the training set, but do not exhibit any further improvement.
On the contrary, the accuracy curves of the DNN under the $S_{1-2}$ strategy imply a smoother behavior.
Hence, the reconsideration of the feature extractors' training choice, such as the proposed $S_{1-2}$ scheme, denotes the best improvement candidate.

Paying closer attention to each individual speaker's performance, the obtain results for $s_3$ and $s_7$ constitute an exception to the rest of our experimental results in Table~{\ref{table:TrFus1}}.
This behavior can be interpreted by re-examining Table~{\ref{table:Unimodal}} (which follows strategy $S_{a-a}$), in order to find out that the audio feature extractor of both speakers exhibits the highest $\mathcal{C}_R$ rates, while also keeping increased $\mathcal{S}_R$ values, indicating similar training and testing distributions.
In a similar manner, the visual feature extractor of $s_6$ also achieves high centrality and separability values in Table~{\ref{table:Unimodal}}, explaining the fact that the alternative fusion strategy has not significantly improved the performance of $S_{a-a}$.
A more careful observation shows that the above also applies for the rest of the speakers mentioned in Section~{\ref{subsub:FetExt}}.
On the contrary, $s_1$, $s_2$ and $s_8$ have been particularly harmed by $S_{a-a}$, which are the main cases that highlight the advantageous properties of $S_{1-2}$.
The visual extractor of $gr_6$, presenting the second best value of $\mathcal{C}_R$ in Table~{\ref{table:Unimodal}}, reveals the adverse effect of $S_{a-a}$ on the fusion model, due to the low produced $\mathcal{S}_R$ values.

Finally, aiming to farther evaluate the effects of our analysis, we applied certain configurations on $S_{1-2}$ and $S_{a-a}$ and conducted the same experiments both on RML and BAUM-1s.
In specific, instead of exploiting the above datasets to train our unimodal extractors, we made use of two unimodal dataset, \textit{viz.}, the AffectNet~\cite{mollahosseini2017affectnet} and IEMOCAP~\cite{busso2008iemocap} datasets for the visual and audio channels, respectively.
Then, the modified $S'_{1-2}$ strategy included the training of the fusion model on our evaluation datasets: RML and BAUM-1s, always sustaining the LOSO and LOSGO schemes.
On the other hand, within $S'_{a-a}$, the unimodal extractors were firstly further fine-tuned on the evaluation datasets and then, the fusion model was also trained on the same data.
The produced results are presented in the last two rows of Table~\ref{table:TrFus1}.
Note that, similarly to our previous results, apart from two cases, the further training of the unimodal channels with data, which are also exploited afterwards for training the fusion model, deteriorated the performance of the overall model.

\subsection{Overfitting in $\mathcal{F}$}

\begin{figure*}
    \centering
    \begin{subfigure}{0.45\textwidth}
        \centering
        \includegraphics[width=.85\linewidth]{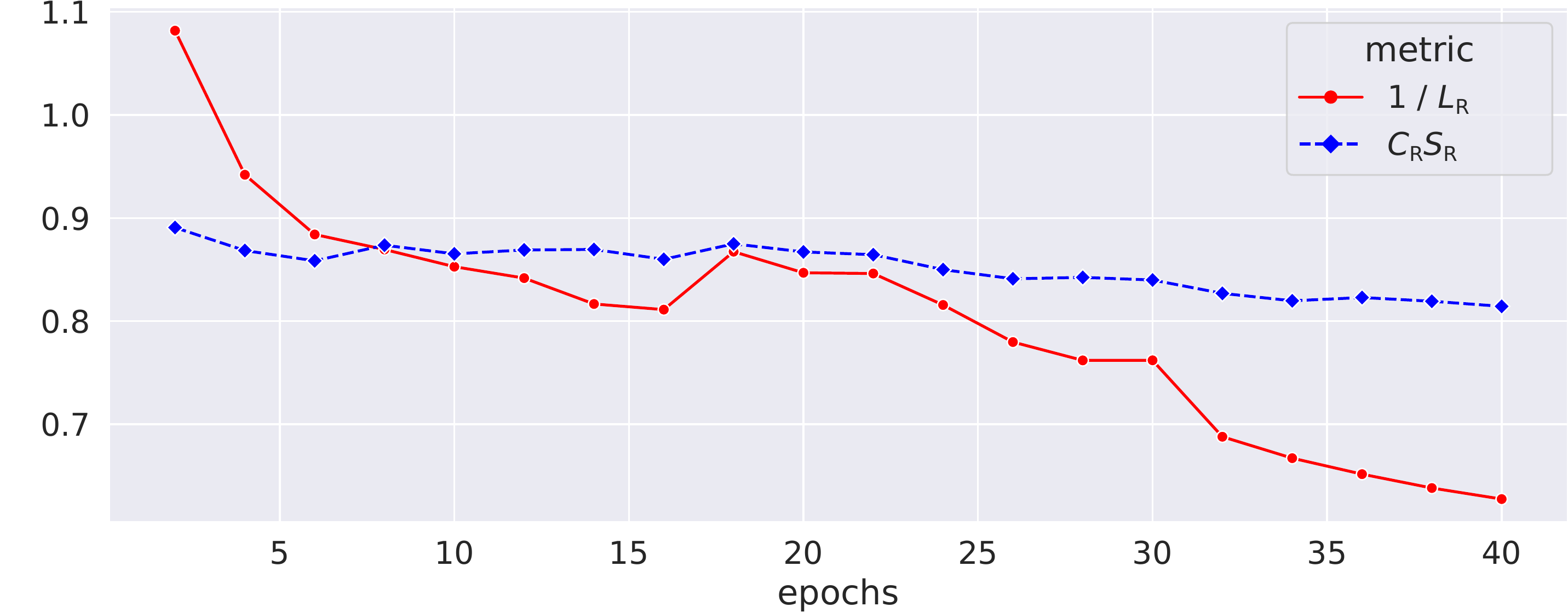}
        \caption{\small Obtained values}
        \label{fig:subimage32a}
    \end{subfigure}%
    \begin{subfigure}{0.45\textwidth}
        \centering
        \includegraphics[width=.85\linewidth]{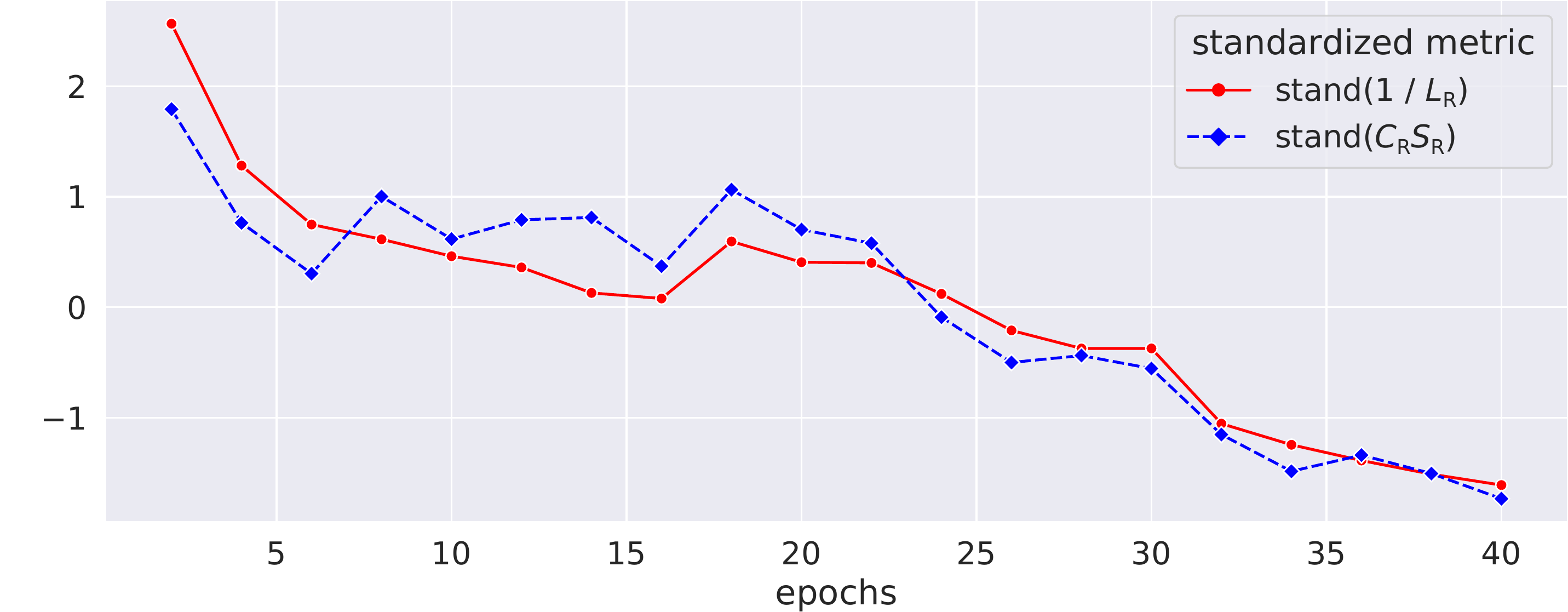}
        \caption{\small Standardized values}
        \label{fig:subimage32b}
    \end{subfigure}
    \caption{\small Qualitative comparison of the proposed $\mathcal{C}_R\mathcal{S}_R$ metric (\textit{blue}) against the inverse of the common loss-ratio $\mathcal{L}_R$ one (\textit{red}) on the large-scale ImageNet32 dataset, using the ResNet-18 architecture. In (a) the obtained values of the two metrics during the training procedure are depicted, while in (b) the standardized values of the two metrics are displayed, proving the high correlation, thus the ability of centrality and separability metrics to capture the divergence between the training and the testing sets in $\mathcal{F}$.}
    \label{fig:image32}
\end{figure*}

At this stage, we present a statistical assessment of $\mathcal{C}_R$ and $\mathcal{S}_R$ regarding their efficiency in capturing overfitting in $\mathcal{F}$.
Hence, we compare their correlation with the loss-ratio ($\mathcal{L}_R$) that is widely exploited to evaluate the quality of each training procedure of DNNs.
Notice that $\mathcal{L}_R$ can not be adopted in cases of feature extractors since the loss is indeterminable.
We exploit the Pearson Correlation Coefficient ($\rho$) between $\mathcal{L}_R$ and the product of the proposed statistics $\mathcal{C}_R\mathcal{S}_R$, with the latter capturing the behavior of both metrics in one value.
Thereafter, we observe in our CIFAR experimentation (Table~{\ref{table:TrPerf}}) a high correlation between them with $\rho=-0.9786$.
The negative sign, indicating the negative correlation, is highly anticipated since the higher the loss ratio of the testing over the training samples, the lower their regarding separability and centrality.
Regarding Table~{\ref{table:Unimodal}}, the correlation coefficient $\rho$ of the visual extractor equals $-0.8397$ on RML and $-0.8415$ on BAUM-1s, while for the audio extractor the corresponding values are $-0.8176$ and $-0.9585$, respectively.
The above essentially indicate that the $\mathcal{C}_R\mathcal{S}_R$ product effectively resembles the behavior of $\mathcal{L}_R$, proving their capability of capturing overfitting even on the output space of a feature extractor.

A final assessment regarding the capacity of $\mathcal{C}_R\mathcal{S}_R$ is conducted, by training the well-established ResNet-18 architecture~{\cite{he2016deep}} on the ImageNet32 dataset~{\cite{chrabaszcz2017downsampled}}.
For every two epochs, the feature vectors of both the training and the testing sets were kept to measure their corresponding $\mathcal{C}_R$ and $\mathcal{S}_R$ values.
The training procedure lasted for $40$ epochs, a batch size of $256$ and an initial learning rate of $0.1$, with step-wise decaying by an order of magnitude when the procedure reaches the $50\%$ and $75\%$ of the its total duration.
In Fig.~{\ref{fig:subimage32a}}, the obtained values of $\mathcal{C}_R\mathcal{S}_R$ are displayed, as well as the inverse of the common loss-ratio $\mathcal{L}_R$ in order to visually assess the positive correlation between the two metrics.
Subsequently, we apply a Z-score normalization (standardization) to both metrics, producing two highly correlated signals, as shown in Fig.~{\ref{fig:subimage32b}}.
Proceeding to the calculation of pearson correlation coefficient between the obtained $\mathcal{C}_R\mathcal{S}_R$ and $\mathcal{L}_R$ values, we find it to be equal to $-0.9558$. This result further denotes the validity of our findings and, thus, the ability of $\mathcal{C}_R\mathcal{S}_R$ to capture the divergence between training and testing sets in $\mathcal{F}$.

\section{Conclusion}
With the paper at hand, a cohesive study has been conducted, describing the basic properties of the feature space $\mathcal{F}$ in a DNN architecture under the classification task.
More specifically, the angular classification property of the Softmax function has been investigated, showing its decisive role in the final result and the feature vectors' orientation.
Our analysis allowed for the representation of the nature of overfitting in $\mathcal{F}$ by using two statistical metrics that focus: (a) on the distribution of the per-class central vectors and (b) the separability between the feature vectors of the bordering classes.
We proved that they can accurately describe the level of overfitting in $\mathcal{F}$, which is mainly considered as a gap between the per-class distributions of the training and testing vectors.
Meanwhile, certain of the adverse effects on cascade and fusion applications of DNNs are denoted. It has been shown that in cases of low centrality and separability values in the unimodal feature extractors, an alternative training strategy should be considered.
This would include the division of the initial available training data into two clusters: one for the unimodal and one for the fusion training procedure.
The above strategy has been proved particularly beneficial in most of our cases, achieving on average of about $5\%$ more accurate results among our two evaluation datasets.
As part of our future work, we shall consider the above property so as to investigate a suitable configuration on the Softmax function that enforces both the centrality and separability values to stay quite higher.
Such a function could potentially enhance the performance of the feature extractors and simultaneously approximate the performance achieved by the proposed training strategy.

\appendices
\section{Partial Derivatives of Softmax}
\label{proof1}
Below, we calculate the partial derivatives for the $j$-th output of the softmax function ($S_j$) over the feature vector's $\bar{a}_e$ norm ($R$), when $i$ is the prevailing class.

\begin{equation}
\label{proof}
\begin{gathered}
\frac{\partial S_j}{\partial R} = \cfrac{\partial\left[\cfrac{e^{z_j}}{\sum_{k=0}^{N}{e^{z_k}}}\right]}{\partial R} = \\
= \cfrac{\cfrac{\partial e^{z_j}}{\partial R}\sum\limits_{k=0}^{N}{e^{z_k}}-e^{z_j}\cfrac{\partial\left[\sum\limits_{k=0}^{N}{e^{z_k}}\right]}{\partial R}}{\left(\sum\limits_{k=0}^{N}{e^{z_k}}\right)^2}=\\
=\cfrac{ e^{z_j}\cfrac{\partial z_j}{\partial R}\sum\limits_{k=0}^{n}{e^{z_k}}-e^{z_j}\sum\limits_{k=0}^{N}{\left(\cfrac{\partial z_k}{\partial R} e^{z_k}\right)}}{\left(\sum\limits_{k=0}^{N}{e^{z_k}}\right)^2} =\\
= S_j  \cfrac{\sum\limits_{k=0}^{N}{\left(\left[\cfrac{\partial z_j}{\partial R}-\cfrac{\partial z_k}{\partial R}\right] e^{z_k}\right)}}{\sum\limits_{k=0}^{N}{e^{z_k}}}.
\end{gathered}
\end{equation}

Similarly, we can calculate the partial derivative over the feature vector's angle ($\theta_i$):
\begin{equation}
\begin{gathered}
\frac{\partial S_j}{\partial \theta_i} = S_j  \cfrac{\sum\limits_{k=0}^{N}{\left(\left[\cfrac{\partial z_j}{\partial \theta_i}-\cfrac{\partial z_k}{\partial \theta_i}\right] e^{z_k}\right)}}{\sum\limits_{k=0}^{N}{e^{z_k}}}.
\end{gathered}
\end{equation}

\section{Proof of R.III}
\label{proof2}

Considering Eq.~\ref{proof}, the partial derivative of the $i$-th softmax output of a feature vector $\bar{a}_e$ over its norm $R$ is given by:
\begin{equation}\label{eq:NormDer}
\frac{\partial S_i}{\partial R} = S_i  \cfrac{\sum\limits_{k=0}^{N}{\left(\left[\cfrac{\partial z_i}{\partial R}-\cfrac{\partial z_k}{\partial R}\right] e^{z_k}\right)}}{\sum\limits_{k=0}^{N}{e^{z_k}}}.
\end{equation}
We know that $S_i>0$ and $\sum\limits_{k=0}^{N}{e^{z_k}}>0$. Moreover, provided that $i$ constitutes the dominant class, we can write for any class $k\neq i$:
\begin{equation}
\begin{gathered}
z_i>z_k, \\
R \|\bar{w}_{i}\|  \cos(\theta_i)>R  \|\bar{w}_{k\parallel}\|  \cos(\theta_i-\phi_k),
\end{gathered}
\end{equation}
according to Eq.~\ref{eq:OpDot}. Eliminating the positive value $R$, we have:
\begin{equation}
\|\bar{w}_{i}\| \cos(\theta_i)>\|\bar{w}_{k\parallel}\| \cos(\theta_i-\phi_k),
\end{equation}
which from Eq.~\ref{eq:DerDot} results in:
\begin{equation}
\frac{\partial z_i}{\partial R}>\frac{\partial z_k}{\partial R}, \forall k\neq i.
\end{equation}
Hence, the numerator of the fraction in Eq.~\ref{eq:NormDer} is also positive, meaning that:
\begin{equation}
\frac{\partial S_i}{\partial R} > 0.
\end{equation}

\ifCLASSOPTIONcaptionsoff
  \newpage
\fi

\bibliographystyle{IEEEtran}
\bibliography{IEEEabrv,root}

\begin{IEEEbiography}[{\includegraphics[width=1in,height=1.25in,clip,keepaspectratio]{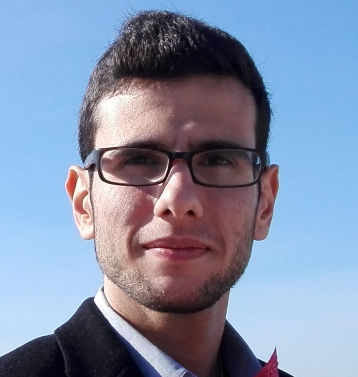}}]{Ioannis Kansizoglou}
received the diploma degree from the Department of Electrical and Computer Engineering, Aristotle University of Thessaloniki, Greece, in 2017. Currently, he is working toward the PhD degree in the Laboratory of Robotics and Automation, Department of Production and Management Engineering, Democritus University of Thrace, Greece, working on emotion analysis and its application in robotics.
\end{IEEEbiography}

\begin{IEEEbiography}[{\includegraphics[width=1in,height=1.25in,clip,keepaspectratio]{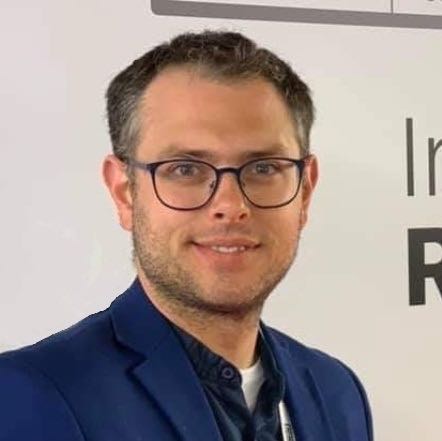}}]{Loukas Bampis}
received the diploma degree in Electrical and Computer Engineering and PhD degree in machine vision from the Democritus University of Thrace (DUTh), Greece, in 2013 and 2019, respectively. He is currently a postdoctoral fellow in the Laboratory of Robotics and Automation, Department of Production and Management Engineering, DUTh. His research interests include real-time localization and place recognition techniques.
\end{IEEEbiography}

\begin{IEEEbiography}[{\includegraphics[width=1in,height=1.25in,clip,keepaspectratio]{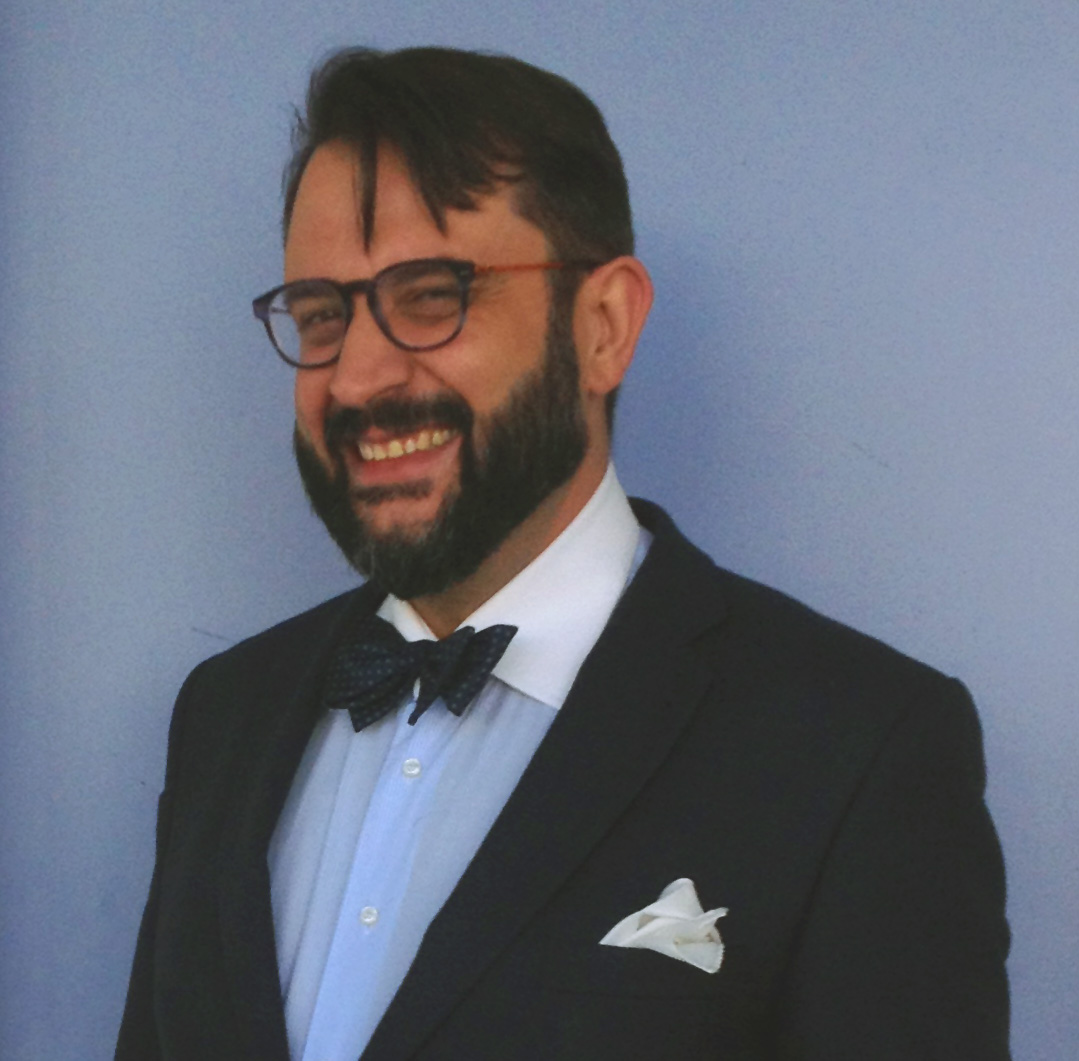}}]{Antonios Gasteratos}
(FIET, SMIEEE) received the MEng and PhD degrees from the Department of Electrical and Computer Engineering, Democritus University of Thrace (DUTh), Greece. He is the director of the Laboratory of Robotics and Automation, DUTh. He has served as a reviewer for numerous scientific journals and international conferences. He is a subject editor at Electronics Letters. His research interests include robotics.
\end{IEEEbiography}

\vfill

\end{document}